%% file: main_arxiv.tex
\documentclass[11pt,letterpaper]{article}

\usepackage[margin=1in]{geometry}
\usepackage[round,authoryear]{natbib}
\input{macros_arxiv}

\title{Generative Modeling with Continuous Flows: Sample Complexity of Flow Matching}

\author[1]{Mudit Gaur}
\author[2]{Prashant Trivedi}
\author[3]{Shuchin Aeron}
\author[4]{Amrit Singh Bedi}
\author[4]{George K. Atia}
\author[5]{Vaneet Aggarwal}

\affil[1]{Department of Electrical and Computer Engineering, Ohio State Univeristy}
\affil[2]{Department of Computer Science and Information Systems, BITS Pilani}
\affil[3]{Department of Electrical and Computer Engineering, Tufts University}
\affil[4]{ Department of Computer Science, University of Central Florida}
\affil[5]{ Edwardson School of Industrial Engineering, Purdue University}

\date{}

\begin{document}
\maketitle
\begin{abstract}
Flow matching has emerged as a promising alternative to diffusion-based generative modeling, offering a
continuous-time ordinary differential equation formulation with fast sampling and a simple regression-based
training objective. Despite its empirical success, some theoretical analyses idealize the learning procedure by
assuming that the velocity field is obtained by exact empirical risk minimization. This assumption does not
reflect the practical finite-dataset setting, where the empirical flow-matching loss is only approximately
minimized by a finite number of stochastic-gradient updates. We develop an empirical-risk-minimizer-free
analysis for flow matching that explicitly accounts for both finite-sample statistical error and finite-time
empirical optimization error. Our analysis considers a finite dataset of \(N\) flow-matching samples and a learned parameter
\(\theta_T\) obtained after \(T\) stochastic gradient updates on the empirical objective. Under regularity assumptions on the interpolation path, sub-exponential source and target distributions, approximation by a
bounded-parameter smooth neural velocity class, empirical-loss optimization, and uniform flow stability, we
prove a guarantee that holds with high probability over the finite training dataset and in expectation over the
stochastic-gradient randomness. The resulting bound separates approximation error, statistical error from
the finite dataset, tail error from unbounded source and target distributions, and empirical optimization error
from approximate minimization of the finite-sample objective. For a prescribed bounded-parameter smooth neural velocity class, the statistical term is controlled through
a metric-entropy/Rademacher uniform-convergence bound for the induced truncated squared-loss class. The
empirical optimization term is controlled by an expected finite-time stochastic-gradient guarantee under an
empirical Polyak-\L{}ojasiewicz condition and a finite-dataset stochastic-gradient oracle. The sub-exponential endpoint tails are handled through an explicit endpoint-magnitude truncation argument. Under the stated approximation, tail, flow-stability, smooth-network-complexity, and empirical optimization
conditions, the bound yields Wasserstein-2 error of order \(O(\epsilon)\) with dataset size
\(\widetilde O(\epsilon^{-4})\) and optimization budget \(\widetilde O(\epsilon^{-2})\), up to the
corresponding problem-dependent and logarithmic factors. Our result provides an ERM-free finite-dataset
analysis of flow matching that explicitly separates finite-sample statistical error from finite-time
stochastic-gradient optimization error on the empirical flow-matching objective.
\end{abstract}

\input{Sections/intro}

\input{Sections/related_works}
\input{Sections/preliminaries}

\input{Sections/approach}
\input{Sections/theory-clean}
\input{Sections/conclusion}

\newpage
\bibliography{references}

\newpage
\appendix
\onecolumn
\input{Sections/lemma_2}

\input{Sections/lemma_3}
\input{Sections/theoretical_result}
\input{Sections/supp_lemmas}

\input{simulation_results}
\end{document}

%% file: macros_arxiv.tex
\usepackage[T1]{fontenc}
\usepackage{times}
\usepackage{microtype}
\usepackage{amsmath,amssymb,amsthm,mathtools,bm,mathrsfs}
\usepackage{aliascnt}
\usepackage{booktabs,array,tabularx,multirow,longtable}
\usepackage{enumitem}
\usepackage{graphicx}
\usepackage{placeins}
\usepackage{xcolor}
\usepackage{url}
\usepackage{authblk}
\usepackage{algorithm}
\usepackage{algorithmic}
\usepackage{hyperref}
\usepackage{subcaption}
\usepackage[nameinlink,noabbrev,capitalise]{cleveref}
\hypersetup{colorlinks=true,linkcolor=blue!45!black,citecolor=blue!45!black,urlcolor=blue!45!black}
\newtheorem{theorem}{Theorem}[section]
\newaliascnt{proposition}{theorem}

\aliascntresetthe{proposition}
\newaliascnt{lemma}{theorem}
\newtheorem{lemma}[lemma]{Lemma}
\aliascntresetthe{lemma}
\newaliascnt{corollary}{theorem}

\aliascntresetthe{corollary}
\newaliascnt{assumption}{theorem}
\newtheorem{assumption}[assumption]{Assumption}
\aliascntresetthe{assumption}
\newaliascnt{condition}{theorem}

\aliascntresetthe{condition}
\theoremstyle{definition}
\newaliascnt{definition}{theorem}

\aliascntresetthe{definition}
\theoremstyle{remark}
\newaliascnt{remark}{theorem}
\newtheorem{remark}[remark]{Remark}
\aliascntresetthe{remark}
\crefname{theorem}{Theorem}{Theorems}
\Crefname{theorem}{Theorem}{Theorems}
\crefname{proposition}{Proposition}{Propositions}
\Crefname{proposition}{Proposition}{Propositions}
\crefname{lemma}{Lemma}{Lemmas}
\Crefname{lemma}{Lemma}{Lemmas}
\crefname{corollary}{Corollary}{Corollaries}
\Crefname{corollary}{Corollary}{Corollaries}
\crefname{assumption}{Assumption}{Assumptions}
\Crefname{assumption}{Assumption}{Assumptions}
\crefname{condition}{Condition}{Conditions}
\Crefname{condition}{Condition}{Conditions}
\crefname{definition}{Definition}{Definitions}
\Crefname{definition}{Definition}{Definitions}
\crefname{remark}{Remark}{Remarks}
\Crefname{remark}{Remark}{Remarks}

\setlist[itemize]{leftmargin=1.35em,itemsep=.15em,topsep=.25em}
\setlist[enumerate]{leftmargin=1.55em,itemsep=.15em,topsep=.25em}

\usepackage{pifont}

\crefname{algorithm}{Algorithm}{Algorithms}
\Crefname{algorithm}{Algorithm}{Algorithms}

%% file: Sections/intro.tex
\section{Introduction}

Generative models have become a central component of modern machine learning, with applications in image synthesis, scientific modeling, computational biology, and reinforcement learning. Diffusion and score-based generative models have achieved strong empirical performance by transforming a simple source distribution into a target data distribution through a learned reverse-time stochastic process \citep{song2021score}. Their success has motivated a substantial theoretical literature on score estimation, denoising objectives, numerical discretization, optimization error, and statistical guarantees for accurate generation. At the same time, diffusion-based sampling can require many numerical integration steps, and its training objective is tied to estimating score functions along a noisy probability path.

Flow matching provides an alternative framework for continuous-time generative modeling \citep{lipman2023flow}. Rather than learning a score function, flow matching learns a time-dependent velocity field whose associated ordinary differential equation transports samples from a source distribution to a target distribution. The learned model is a deployable velocity field of the form
\[
u_\theta(x,t),
\]
which depends only on the current state and time. This viewpoint leads to a regression-based training objective and can enable efficient deterministic sampling through ODE simulation. Several variants have been proposed, including rectified flow, conditional flow matching, optimal-transport flow matching, and stochastic interpolants
\citep{lipman2023flow,liu2023flow,tong2023improving,albergo2023stochastic}.
These works demonstrate that both the interpolation path and the coupling between source and target samples can substantially affect the geometry, regularity, and computational efficiency of the resulting generative flow.

Despite this empirical and methodological progress, the theoretical understanding of how flow-matching velocity fields are learned from finite data remains incomplete. In the diffusion setting, recent works study sample complexity and convergence guarantees for score estimation and score-based generation
\citep{chen2023restoration,guptaimproved,gaur2025sample,guan2025mirror},
as well as discrete diffusion models
\citep{srikanth2026discretestatediffusionmodels}.
These results address questions such as how accurately a learned vector field must approximate its population target and how many samples or optimization steps are sufficient for the generated distribution to approach the desired target distribution. In the flow-matching setting, existing work has established stability and error-propagation guarantees for learned flows \citep{benton2024error}, while recent analyses study statistical, approximation, and sample-complexity properties of flow matching and rectified flows
\citep{zhou2025an,sahoo2026orderoptimalsamplecomplexityrectified}.
These results provide important complementary perspectives on velocity estimation and terminal distributional error.

This paper focuses on a distinction between statistical and optimization error that arises in practical finite-dataset training. A flow-matching model is commonly trained using a fixed stored dataset and a finite number of stochastic-gradient updates. Consequently, the empirical flow-matching objective is generally only approximately minimized. This setting differs both from an idealized exact empirical-risk-minimization formulation and from a population-oracle or fresh-sample optimization model. Two distinct effects must therefore be controlled. The first is the statistical error incurred by replacing the population objective with a finite empirical objective. The second is the empirical optimization error incurred by performing only finitely many stochastic-gradient updates on that fixed objective.

We develop an empirical-risk-minimizer-free analysis for this finite-dataset setting. Given a training dataset of $N$ flow-matching samples, we study the parameter $\theta_T$ obtained after $T$ stochastic-gradient updates on the empirical objective
\(\widehat L_N.\)
Our analysis explicitly separates the discrepancy between the population loss $L$ and the empirical loss $\widehat L_N$ from the error caused by only approximately minimizing $\widehat L_N$. An empirical risk minimizer is introduced only as a theoretical comparator and is not assumed to be available to the training algorithm.

Our formulation is based on a coupling between a source distribution ($\pi_0$) and a target distribution ($\pi_1$), together with a linear, or rectified, interpolation path. The regression target is the associated pathwise velocity, while the deployable target is its conditional mean given the current state and time. This marginal velocity field depends only on the current state and time and can therefore be used during generation. We explicitly assume that the marginal velocity field is path-consistent with the prescribed interpolation path, so that the corresponding reference ODE reproduces the path marginals and reaches ($\pi_1$) at terminal time. A uniform flow-stability argument then converts integrated velocity-field error along the reference path into Wasserstein-2 error for the generated terminal distribution.

A further distinction from compact-support formulations is that we allow sub-exponential source and target distributions. This is important because standard source distributions used in generative modeling, including Gaussian noise, are unbounded. We handle unbounded data through endpoint-magnitude truncation. On the event that both coupled endpoints have bounded norm, the linear interpolation state, pathwise velocity, and induced squared-loss class are uniformly controlled. The contribution of the complementary event is then bounded using the assumed endpoint-tail behavior. This construction avoids truncation arguments involving singular factors near the terminal time and is naturally compatible with linear and rectified interpolation paths.

The central question addressed in this work is:

\begin{center}
\emph{How many training samples and stochastic-gradient updates are sufficient for a neural velocity field trained by flow matching to generate a distribution close to the target distribution?}
\end{center}

Our proof proceeds in two stages. First, uniform flow stability converts excess population flow-matching risk into terminal Wasserstein-2 error. Second, the excess population risk is decomposed into approximation, finite-dataset statistical, and empirical optimization contributions. Under unbounded endpoint distributions, the statistical term further separates into a truncated uniform-convergence contribution and an endpoint-tail contribution. For a prescribed bounded-parameter smooth neural velocity class, the truncated statistical contribution is controlled through a metric-entropy and Rademacher uniform convergence argument for the induced scalar squared-loss class. The empirical optimization contribution is controlled through an expected finite time stochastic gradient guarantee under empirical smoothness, an empirical proximal Polyak \L{}ojasiewicz condition, and a finite dataset stochastic gradient oracle.

The resulting theorem holds with high probability over the draw of the finite training dataset and in conditional expectation over the stochastic-gradient randomness. Under the stated approximation, endpoint-tail, flow-stability, neural-class-complexity, and empirical-optimization conditions, the bound yields expected Wasserstein-2 error of order
\(
O(\epsilon)
\)
with dataset size
\(
\widetilde O\left(\epsilon^{-4}\right)
\)
and optimization budget
\(
\widetilde O\left(\epsilon^{-2}\right),
\)
up to the displayed problem-dependent and logarithmic factors. These rates should be interpreted for a prescribed bounded-parameter smooth neural velocity class. In particular, dimension and architecture dependence may enter through the neural-class complexity, approximation error, endpoint-tail parameters, flow-stability constant, and empirical optimization constants. We do not claim that these quantities are dimension free or uniformly controlled over arbitrary architectures.

We complement the theoretical analysis with synthetic experiments designed to illustrate the mechanisms represented in the bound. The experiments examine how velocity-estimation and terminal distributional errors vary with the dataset and optimization budgets, compare fixed finite-dataset training with a one-pass fresh-sample procedure, and study the effect of different couplings for a linear interpolation path. These experiments are not intended to verify the formal asymptotic rates or to compete with large-scale generative-model benchmarks. Rather, they provide controlled diagnostic evidence for the approximation--statistical--optimization structure underlying the analysis.

The main contributions of this work are summarized as follows:

\begin{itemize}
\item \textbf{Finite-dataset ERM-free analysis for flow matching.}
We analyze a velocity field trained on a fixed finite empirical flow-matching objective without assuming that the training algorithm has access to an exact empirical risk minimizer. The learned parameter is the output of a finite number of stochastic-gradient updates.

\item \textbf{Separation of finite-sample statistical and empirical optimization error.}
We distinguish the error incurred by replacing the population objective with a finite empirical objective from the error incurred by only approximately minimizing that empirical objective. This makes explicit the separate roles of the dataset size \(N\) and the optimization budget \(T\).

\item \textbf{Sub-exponential endpoints and explicit tail control.}
We replace compact-support assumptions with sub-exponential source and target distributions. Endpoint-magnitude truncation yields a bounded squared-loss class on a high-probability event, while the contribution of its complement is controlled through an explicit tail term. This permits standard unbounded source distributions such as Gaussian noise.

\item \textbf{Metric-entropy control for a smooth neural squared-loss class.}
For a prescribed bounded-parameter smooth neural velocity class, we derive a uniform-convergence bound for the induced truncated scalar squared-loss class using parameter-space covering numbers and Rademacher complexity. This yields a squared-loss-level statistical contribution of order
\(
    \widetilde O\left(N^{-1/2}\right),
\)
up to the displayed network-complexity, truncation-envelope, and logarithmic factors.

\item \textbf{Path-based formulation with a deployable marginal velocity field.}
We formulate the learning problem using interpolation paths induced by couplings between the source and target distributions while ensuring that the learned velocity field depends only on the current state and time. The coupling is treated as part of the prescribed path and can affect the associated velocity-regression problem and its learning difficulty.

\item \textbf{Finite-time empirical optimization guarantee.}
Under empirical smoothness, an empirical proximal-PL condition, bounded conditional gradient variance, and the stated uniform empirical regularity conditions, we control the expected empirical suboptimality after \(T\) finite-dataset stochastic-gradient updates. Combined with the statistical and flow-stability arguments, this produces a finite-sample and finite-time terminal distributional guarantee.

\end{itemize}

%% file: Sections/related_works.tex
\subsection{Related Work}

Works such as \citet{chen2023restoration} proved iteration complexity bounds for diffusion models. More recent work, such as \citet{guan2025mirror}, provides convergence analysis for generative modeling in convex domains. Several recent works have analyzed the sample complexity of score-based diffusion models. \citet{guptaimproved} derive a sample complexity bound assuming access to ERM, and \citet{gaur2025sample} relax this assumption, not requiring access to ERM.

\citet{zhou2025an} performed a sample complexity analysis for flow matching models wherein the Wasserstein distance between the true and estimated distributions is bounded by decomposing the error in calculating the velocity function into errors incurred due to the limited approximation power of the neural network function class, as well as the finite sample size. As stated earlier, this work assumes access to the ERM of the velocity estimation loss. Additionally, the sample complexity result obtained in \citet{zhou2025an} is exponential in the data dimension. Our result avoids this exponential dependence by assuming a constant error due to the limited approximation power of the class of neural networks used to approximate the velocity function. We do this since the focus of our work is to analyze the statistical and optimization errors incurred in flow matching. Similar assumptions have been used in previous works on diffusion modeling, such as \citet{guptaimproved} and \citet{guan2025mirror}.

Recently, \citet{benton2024error} presented error bounds for the generative process using flow matching, assuming that the true velocity field lies within an $\epsilon^{2}$ neighborhood of the learned velocity field. Their analysis provides error bounds under the assumption of upper bounded $L_2$ approximation error and certain regularity conditions on the data distribution. However, their results do not address the finite time sample complexity, a crucial aspect for understanding the number of training samples required for estimating the velocity field. This work is inspired by existing theoretical analyses  such as \citep{pmlr-v202-gaur23a},  \citep{gaur2023global},\citep{gaur2024closinggapachievingglobal} as well as  diffusion models \citep{gaur2025improvedsamplecomplexitydiffusion,srikanth2026discrete}. We adopt a similar high-level decomposition as a conceptual starting point. 

%% file: Sections/preliminaries.tex
\section{Preliminaries and Problem Formulation}

We introduce the flow-matching setup used throughout the paper. The goal is to learn a continuous-time
generative model that transports samples from a source distribution to a target distribution by solving an
ordinary differential equation. We first define the source and target distributions, specify probability paths
through couplings, introduce the associated marginal velocity field, and then state the finite-dataset learning
objective analyzed in this work.

\paragraph{Source and target distributions.}
Let \(\pi_0\) denote a source distribution on \(\mathbb{R}^d\), such as a standard Gaussian distribution,
and let \(\pi_1\) denote the target data distribution. The goal of generative modeling is to construct a
sampling procedure that transforms an initial sample from \(\pi_0\) into a sample whose distribution is close
to \(\pi_1\).

Given a time-dependent velocity field
\[
    u:
    \mathbb{R}^d
    \times
    [0,1]
    \to
    \mathbb{R}^d,
\]
the associated ODE is
\[
    \frac{d}{dt}Y_t
    =
    u(Y_t,t),
    \qquad
    Y_0
    \sim
    \pi_0.
\]
If the terminal law of \(Y_1\) is close to \(\pi_1\), then this ODE defines a generative model.

\paragraph{Probability paths and couplings.}
Flow matching specifies a reference probability path connecting \(\pi_0\) and \(\pi_1\). Let
\[
    (X_0,X_1)
    \sim
    \Gamma,
    \qquad
    X_0
    \sim
    \pi_0,
    \qquad
    X_1
    \sim
    \pi_1,
\]
where
\(
    \Gamma
    \in
    \Pi(\pi_0,\pi_1)
\)
is a coupling between the source and target distributions. Let
\[
    I_t:
    \mathbb{R}^d
    \times
    \mathbb{R}^d
    \to
    \mathbb{R}^d,
    \qquad
    t
    \in
    [0,1],
\]
be an interpolation map satisfying the endpoint conditions
\[
    I_0(x_0,x_1)
    =
    x_0,
    \qquad
    I_1(x_0,x_1)
    =
    x_1.
\]
The associated path is
\[
    X_t
    :=
    I_t(X_0,X_1),
    \qquad
    \rho_t
    :=
    \operatorname{Law}(X_t).
\]
Thus,
\[
    \rho_0
    =
    \pi_0,
    \qquad
    \rho_1
    =
    \pi_1.
\]
When the interpolation map is differentiable in time, define the pathwise velocity by
\[
    \dot X_t
    :=
    \frac{d}{dt}
    I_t(X_0,X_1).
\]
This quantity describes the velocity of a particle moving along the interpolation path associated with the
coupled endpoint pair \((X_0,X_1)\).

\paragraph{Marginal velocity field.}
The pathwise velocity \(\dot X_t\) depends on the coupled endpoints \((X_0,X_1)\). During generation,
however, these endpoints are not observed. A generative ODE can only depend on the current state and time.
The relevant deployable target velocity field is therefore the conditional mean
\[
    u^\star(x,t)
    :=
    \mathbb{E}
    \left[
        \dot X_t
        \,\middle|\,
        X_t=x,\,
        t
    \right].
\]
The field \(u^\star\) depends only on \((x,t)\) and is the ideal deployable target that the learned velocity
field seeks to approximate. The fact that the ODE driven by \(u^\star\) reproduces the prescribed path marginals is not inferred solely
from the conditional-mean definition above. Instead, the required path-consistency property is imposed
explicitly in Assumption~\ref{assump:flow_stability}.

\paragraph{The linear interpolation path.}
The main theorem in this work is stated for the linear, or rectified, interpolation path
\[
    X_t
    =
    (1-t)X_0+tX_1,
    \qquad
    t
    \in
    [0,1].
\]
For this path,
\[
    \dot X_t
    =
    X_1-X_0.
\]
The corresponding marginal velocity field is
\[
    u^\star(x,t)
    =
    \mathbb{E}
    \left[
        X_1-X_0
        \,\middle|\,
        X_t=x,t\
    \right].
\]
This formulation includes the standard rectified-flow setup and allows different choices of the coupling
\(\Gamma\), such as independent couplings or optimal-transport-inspired couplings. The coupling is part of
the probability-path specification: changing \(\Gamma\) changes the regression target and can affect the
difficulty of learning the marginal velocity field, even when the interpolation map itself remains linear.

\paragraph{Learned velocity field.}
We approximate the marginal velocity field \(u^\star\) using a neural network
\[
    u_\theta:
    \mathbb{R}^d
    \times
    [0,1]
    \to
    \mathbb{R}^d,
    \qquad
    \theta
    \in
    \Theta.
\]
The learned velocity field is deployable because it takes only the current state \(x\) and time \(t\) as
inputs. For a fixed parameter \(\theta\), samples are generated by solving
\[
    \frac{d}{dt}Y_t^\theta
    =
    u_\theta(Y_t^\theta,t),
    \qquad
    Y_0^\theta
    \sim
    \pi_0.
\]
We denote the terminal law of this learned flow by
\[
    \widehat\pi^\theta_1
    :=
    \operatorname{Law}(Y_1^\theta).
\]
For the final learned parameter \(\theta_T\), the generated terminal distribution is \(\widehat\pi^{\theta_T}_1.\)
The objective is to make this distribution close to the target distribution \(\pi_1\).

\paragraph{Population flow-matching objective.}
For the linear interpolation path, a population flow-matching sample is
\[
    Z
    =
    (X_0,X_1,t),
    \qquad
    (X_0,X_1)
    \sim
    \Gamma,
    \qquad
    t
    \sim
    \operatorname{Uniform}[0,1].
\]
The corresponding state and pathwise velocity are
\[
    X_t
    =
    (1-t)X_0+tX_1,
    \qquad
    \dot X_t
    =
    X_1-X_0.
\]
For a parameter \(\theta\in\Theta\), define the single-sample flow-matching loss
\[
    \ell_\theta(Z)
    :=
    \left\|
        u_\theta
        \left(
            (1-t)X_0+tX_1,
            t
        \right)
        -
        (X_1-X_0)
    \right\|_2^2.
\]
The population flow-matching loss is
\[
    L(\theta)
    :=
    \mathbb{E}
    \left[
        \ell_\theta(Z)
    \right].
\]
Equivalently,
\[
    L(\theta)
    =
    \int_0^1
    \mathbb{E}_{(X_0,X_1)\sim\Gamma}
    \left[
        \left\|
            u_\theta
            \left(
                (1-t)X_0+tX_1,
                t
            \right)
            -
            (X_1-X_0)
        \right\|_2^2
    \right]
    dt.
\]

Let
\[
    L^\star
    :=
    \inf_{u}
    \mathbb{E}
    \left[
        \left\|
            u(X_t,t)-\dot X_t
        \right\|_2^2
    \right],
\]
where the infimum is over all measurable deployable velocity fields
\[
    u:
    \mathbb{R}^d
    \times
    [0,1]
    \to
    \mathbb{R}^d.
\]
The minimizer is the marginal velocity field \(u^\star\). By the \(L^2\) projection identity,
\[
    L(\theta)-L^\star
    =
    \mathbb{E}
    \left[
        \left\|
            u_\theta(X_t,t)-u^\star(X_t,t)
        \right\|_2^2
    \right].
\]
Thus, controlling the excess population flow-matching loss is equivalent to controlling the integrated
squared error between the learned deployable velocity field and the target marginal velocity field along the
reference path.

\paragraph{Empirical flow-matching objective.}
In practice, the population objective is replaced by a finite empirical objective. Given an i.i.d.\ training
dataset
\[
    \mathcal{D}_N
    :=
    \{Z_i\}_{i=1}^N,
    \qquad
    Z_i
    =
    (X_{0,i},X_{1,i},t_i),
\]
where
\[
    (X_{0,i},X_{1,i})
    \sim
    \Gamma,
    \qquad
    t_i
    \sim
    \operatorname{Uniform}[0,1],
\]
the empirical flow-matching loss is
\[
    \widehat L_N(\theta)
    :=
    \frac{1}{N}
    \sum_{i=1}^N
    \ell_\theta(Z_i).
\]
Equivalently,
\[
    \widehat L_N(\theta)
    =
    \frac{1}{N}
    \sum_{i=1}^N
    \left\|
        u_\theta
        \left(
            (1-t_i)X_{0,i}+t_iX_{1,i},
            t_i
        \right)
        -
        (X_{1,i}-X_{0,i})
    \right\|_2^2.
\]

Let
\[
    \widehat\theta_N
    \in
    \arg\min_{\theta\in\Theta}
    \widehat L_N(\theta)
\]
denote an empirical risk minimizer. This parameter is used only as a theoretical comparator; the algorithm
is not assumed to have access to it. Instead, the learned parameter is obtained by running a finite number
\(T\) of stochastic-gradient updates on the empirical objective \(\widehat L_N\). We denote the resulting
parameter by
\(\theta_T.\) Thus, the analysis distinguishes two resources:
\[
    N
    =
    \text{number of training samples},
    \qquad
    T
    =
    \text{number of stochastic-gradient updates}.
\]
The dataset size \(N\) controls the statistical error between \(L\) and \(\widehat L_N\), while the
optimization budget \(T\) controls how closely \(\theta_T\) approaches the empirical minimizer
\(\widehat\theta_N\).

\paragraph{Problem statement.}
This work studies the following question:

\begin{center}
\emph{How many training samples and stochastic-gradient updates are sufficient for a neural velocity field
trained by flow matching to generate a distribution close to the target distribution?}
\end{center}

More precisely, if \(\theta_T\) denotes the parameter obtained after \(T\) stochastic-gradient updates on the
finite empirical objective \(\widehat L_N\), we seek to bound
\(
    W_2
    \left(
        \widehat\pi^{\theta_T}_1,
        \pi_1
    \right).
\)
The remainder of the paper develops a finite-sample and finite-time analysis of this quantity without
assuming access to the exact empirical risk minimizer \(\widehat\theta_N\).

%% file: Sections/approach.tex
\section{Our Approach}
\label{sec:approach}

We now describe the assumptions and proof strategy used to obtain an ERM-free
finite-dataset sample-complexity guarantee for flow matching. The goal is to control
the terminal distributional error of the learned flow in terms of the finite-sample and
finite-time errors incurred while learning the velocity field. The main ingredients are a
sub-exponential data condition, an approximation condition for the neural velocity
class, a statistical uniform-convergence bound for the empirical flow-matching
objective, an empirical optimization condition for projected finite-dataset
stochastic-gradient training, and a path-consistent uniform flow-stability condition
converting velocity-field error into Wasserstein error.

\subsection{Assumptions}

We first collect the assumptions used in the main result.

\begin{assumption}[Sub-exponential source and target distributions]
\label{assump:subexp}

There exist constants \(
    K_X,
    c_0,
    c_1
    >
    0
\) such that, for every \(r\geq 0\),
\[
    \mathbb{P}
    \left(
        \|X_0\|_2>r
    \right)
    \le
    c_0
    \exp
    \left(
        -c_1r/K_X
    \right),
    \qquad
    \mathbb{P}
    \left(
        \|X_1\|_2>r
    \right)
    \le
    c_0
    \exp
    \left(
        -c_1r/K_X
    \right).
\]
Here
\[
    (X_0,X_1)
    \sim
    \Gamma,
    \qquad
    \Gamma
    \in
    \Pi(\pi_0,\pi_1),
\]
is the coupling used to define the interpolation path.
\end{assumption}

Assumption~\ref{assump:subexp} replaces compact support of the source and target
distributions. This is important because standard source distributions used in
generative modeling, such as Gaussian noise, are unbounded. Sub-exponential and
related tail assumptions are standard in high-dimensional probability and statistical
learning theory \citep{vershynin2018high,wainwright2019high}, and analogous
tail-control conditions are common in recent analyses of diffusion and generative
models \citep{guptaimproved,gaur2025sample}. In our proof,
this assumption is used only to control the contribution of the endpoint-truncation
complement.

\begin{assumption}[Approximation error]
\label{assump:approx}

Let \(L^\star\) denote the minimum population flow-matching risk over all
measurable deployable velocity fields. Let \(
    \theta^\star_\Theta
    \in
    \arg\min_{\theta\in\Theta}
    L(\theta)
\) be a population minimizer within the prescribed neural velocity class. There exists
\(
    \epsilon_{\mathrm{approx}}
    \ge
    0
\)
such that
\[
    L(\theta^\star_\Theta)
    -
    L^\star
    \le
    \epsilon_{\mathrm{approx}}.
\]
\end{assumption}

Assumption~\ref{assump:approx} isolates the approximation power of the prescribed
neural velocity class. We keep \(\epsilon_{\mathrm{approx}}\) explicit throughout
the theorem rather than optimizing over architectures. This separates approximation
effects from finite-dataset statistical error and finite-time optimization error.

\begin{assumption}[Empirical proximal-PL condition]
\label{assump:empirical_pl}

For the finite dataset
\(
    \mathcal{D}_N
    =
    \{Z_i\}_{i=1}^N,
\)
let 
\[
    \widehat\theta_N
    \in
    \arg\min_{\theta\in\Theta}
    \widehat L_N(\theta)
\] 

denote an empirical risk minimizer over the compact convex set \(\Theta\). For
\(\theta\in\Theta\), define the constrained gradient-mapping quantity
\[
    \mathcal{M}_N(\theta)
    :=
    -2\alpha_N
    \min_{\theta'\in\Theta}
    \left[
        \left\langle
            \nabla\widehat L_N(\theta),
            \theta'-\theta
        \right\rangle
        +
        \frac{\alpha_N}{2}
        \|\theta'-\theta\|_2^2
    \right],
\]
where \(\alpha_N\) is the empirical smoothness constant from
Assumption~\ref{assump:empirical_sgd}. The empirical flow-matching loss
\(\widehat L_N\) satisfies the proximal Polyak--\L{}ojasiewicz condition on
\(\Theta\): there exists \(\mu_N>0\) such that, for every \(\theta\in\Theta\),
\[
    \frac{1}{2}
    \mathcal{M}_N(\theta)
    \ge
    \mu_N
    \left(
        \widehat L_N(\theta)
        -
        \widehat L_N(\widehat\theta_N)
    \right).
\]
\end{assumption}

The proximal Polyak--\L{}ojasiewicz condition is the constrained analogue of the
usual gradient-domination condition used in the analysis of projected and proximal
first-order methods \citep{karimi2016linear}. When \(\Theta=\mathbb{R}^P\), or
whenever the constraint is inactive in the relevant optimization region, we have
\[
    \mathcal{M}_N(\theta)
    =
    \|
        \nabla\widehat L_N(\theta)
    \|_2^2,
\]
and Assumption~\ref{assump:empirical_pl} reduces to the ordinary
Polyak--\L{}ojasiewicz inequality
\[
    \frac{1}{2}
    \|
        \nabla\widehat L_N(\theta)
    \|_2^2
    \ge
    \mu_N
    \left(
        \widehat L_N(\theta)
        -
        \widehat L_N(\widehat\theta_N)
    \right).
\]
The empirical minimizer \(\widehat\theta_N\) is a theoretical comparator; the
algorithm is not assumed to have access to it.

\begin{assumption}[Empirical smoothness, projected SGD, and interior operating regime]
\label{assump:empirical_sgd}

For the finite dataset
\[
    \mathcal{D}_N
    =
    \{Z_i\}_{i=1}^N,
\]
the empirical loss
\[
    \widehat L_N(\theta)
    =
    \frac{1}{N}
    \sum_{i=1}^N
    \ell_\theta(Z_i)
\]
is differentiable in \(\theta\) and is \(\alpha_N\)-smooth on the compact convex
parameter region \(\Theta\). That is, for every
\(\theta,\theta'\in\Theta\),
\[
    \left\|
        \nabla\widehat L_N(\theta)
        -
        \nabla\widehat L_N(\theta')
    \right\|_2
    \le
    \alpha_N
    \left\|
        \theta-\theta'
    \right\|_2.
\]

For the smooth-activation network class described in
Section~\ref{subsec:neural_vel_class}, this differentiability requirement is
compatible with the architecture. In particular, for every fixed finite dataset
\(\mathcal{D}_N\), bounded parameters and bounded activation derivatives imply
that a finite empirical smoothness constant \(\alpha_N\) exists on \(\Theta\).
The value of \(\alpha_N\) may depend on the fixed dataset, the architecture, the
parameter bound, the activation constants, and the endpoint magnitudes in the
dataset.

Conditioned on \(\mathcal{D}_N\), the algorithm is initialized at
\(
    \theta_0
    \in
    \Theta
\)
and performs projected finite-dataset stochastic-gradient updates
\[
    \theta_{k+1}
    =
    \Pi_\Theta
    \left(
        \theta_k
        -
        \eta_k g_k
    \right),
    \qquad
    g_k
    =
    \nabla_\theta
    \ell_{\theta_k}(Z_{I_k}),
\]
where \(
    I_k
    \sim
    \operatorname{Uniform}
    \left\{
        1,\ldots,N
    \right\}
\) is sampled independently of the past optimization history and
\(\Pi_\Theta\) denotes Euclidean projection onto \(\Theta\).

Let \(
    \{\mathcal{F}_k\}_{k\geq 0}
\) denote the filtration generated by the optimization process up to iteration \(k\),
conditioned on \(\mathcal{D}_N\). The stochastic-gradient oracle is conditionally
unbiased for the empirical objective:
\[
    \mathbb{E}
    \left[
        g_k
        \,\middle|\,
        \mathcal{F}_k,\mathcal{D}_N
    \right]
    =
    \nabla\widehat L_N(\theta_k).
\]
Moreover, there exists
\(
    \sigma_N^2
    >
    0
\) such that
\[
    \mathbb{E}
    \left[
        \left\|
            g_k
            -
            \nabla\widehat L_N(\theta_k)
        \right\|_2^2
        \,\middle|\,
        \mathcal{F}_k,\mathcal{D}_N
    \right]
    \le
    \sigma_N^2.
\]

Finally, to obtain the clean \(O(T^{-1})\) optimization bound under the bounded
conditional-variance assumption above, we work in an interior operating regime:
for every
\(
    k
    =
    0,\ldots,T-1,
\) the projected update is inactive both for the realized stochastic step and for the
corresponding mean-gradient step, namely
\[
    \theta_k
    -
    \eta_k g_k
    \in
    \Theta,
    \qquad
    \theta_k
    -
    \eta_k
    \nabla\widehat L_N(\theta_k)
    \in
    \Theta
\]
almost surely conditional on \(\mathcal{D}_N\). Equivalently, the projection
acts as the identity along the optimization trajectory and the corresponding
mean-gradient trajectory.
\end{assumption}

\begin{lemma}[Automatic containment]
\label{lemma:automatic_containment}

Since
\(
    \theta_0
    \in
    \Theta
\)
and \(\Pi_\Theta\) maps \(\mathbb{R}^P\) into \(\Theta\), the projected recursion
in Assumption~\ref{assump:empirical_sgd} satisfies
\(
    \theta_k
    \in
    \Theta,
    \qquad
k = 0,1,\ldots,T.
\) In particular,
\(
    \theta_T
    \in
    \Theta.
\)
\end{lemma}

Smoothness and bounded conditional gradient variance are standard assumptions in
stochastic-gradient analysis
\citep{ghadimi2013stochastic,ajalloeian2020convergence,koloskova2022sharper}.
In the present finite-dataset setting, the stochastic gradient is unbiased for the
empirical gradient
\(
    \nabla\widehat L_N(\theta_k),
\)
rather than for the population gradient
\(
    \nabla L(\theta_k).
\) This distinction is central to the analysis because the statistical error of the
empirical objective and the optimization error from approximately minimizing it
are controlled separately.

The projection guarantees that the iterates remain in the compact parameter region
on which the empirical smoothness and network regularity properties are imposed.
The separate interior-operating-regime clause is used only to retain the clean
bounded-variance \(O(T^{-1})\) optimization recursion without an additional
boundary-active error term. If projection is active at non-negligible frequency,
one would instead require an interpolation-type noise condition or a separate
boundary analysis.

\begin{assumption}[Uniform-in-\(N\) empirical-constant regularity]
\label{assump:uniform_empirical_constants}

There exist deterministic constants
\[
    \mu_{\min}
    >
    0,
    \qquad
    \alpha_{\max}
    <
    \infty,
    \qquad
    \sigma_{\max}^2
    <
    \infty,
    \qquad
    \overline e_0
    <
    \infty,
\]
all independent of \(N\), and for each \(N\) an event
\(
    \mathcal{G}_N
\)
over the draw of \(\mathcal{D}_N\) such that
\(
    \mathbb{P}
    \left(
        \mathcal{G}_N
    \right)
    \ge
    1-\delta_{\mathrm{opt}}.
\)
On \(\mathcal{G}_N\), the empirical proximal-PL constant \(\mu_N\), the
empirical smoothness constant \(\alpha_N\), the conditional gradient-variance
bound \(\sigma_N^2\), and the initialization gap
\[
    e_0
    :=
    \widehat L_N(\theta_0)
    -
    \widehat L_N(\widehat\theta_N)
\]
satisfy simultaneously
\(
    \mu_{\min}
    \le
    \mu_N
    \le
    \alpha_N
    \le
    \alpha_{\max},
    \sigma_N^2
    \le
    \sigma_{\max}^2,
    \qquad
    e_0
    \le
    \overline e_0.
\)
\end{assumption}

Assumption~\ref{assump:uniform_empirical_constants} makes explicit the
\(N\)-uniformity required for the displayed
\(
    T
    =
    \widetilde O
    \left(
        \epsilon^{-2}
    \right)
\)
optimization scaling. It is a regularity condition on the realized empirical
optimization landscape that is stated rather than derived. Deriving
\(\mu_{\min}\), or more generally characterizing the \(N\)-dependence of the
empirical constants for a particular architecture and initialization scheme, is
outside the scope of the present modular guarantee.

\begin{assumption}[Flow regularity, path consistency, and uniform stability]
\label{assump:flow_stability}

For every
\(
    \theta
    \in
    \Theta,
\)
the learned ODE
\[
    \frac{d}{dt}
    Y_t
    =
    u_\theta(Y_t,t)
\]
admits a unique solution on \([0,1]\) for every initial condition. The reference
ODE driven by the marginal velocity field \(u^\star\) also admits a unique
solution on \([0,1]\) for every initial condition. For every
\(
    \theta
    \in
    \Theta,
    \qquad
    t
    \in
    [0,1],
\)
let
\(
    L_\theta(t)
\)
be a spatial Lipschitz constant of
\(
    u_\theta(\cdot,t).
\)
That is,
\[
    \|u_\theta(x,t)-u_\theta(x',t)\|_2
    \le
    L_\theta(t)
    \|x-x'\|_2
\]
for every
\(
    x,x'
    \in
    \mathbb{R}^d.
\)
Assume that there exists a deterministic constant
\(
    \Lambda_{\mathrm{flow}}
    <
    \infty
\)
such that
\[
    \sup_{\theta\in\Theta}
    \int_0^1
    L_\theta(t)
    \,dt
    \le
    \Lambda_{\mathrm{flow}}.
\]
Define the uniform flow-stability constant
\[
    K_{\mathrm{flow}}
    :=
    \exp
    \left(
        \Lambda_{\mathrm{flow}}
    \right).
\]

Assume moreover that the marginal velocity field
\(
    u^\star
\) is path-consistent with the prescribed interpolation path. That is, if
\(Z_t\) denotes the unique solution of
\[
    \frac{d}{dt}
    Z_t
    =
    u^\star(Z_t,t),
    \qquad
    Z_0
    \sim
    \pi_0,
\]
then
\[
    \operatorname{Law}(Z_t)
    =
    \rho_t
    \qquad
    \text{for every }
    t
    \in
    [0,1].
\]
In particular,
\[
    \operatorname{Law}(Z_1)
    =
    \rho_1
    =
    \pi_1.
\]
\end{assumption}

Assumption~\ref{assump:flow_stability} is used to convert velocity-field
estimation error into terminal distributional error. The constant
\(
    K_{\mathrm{flow}}
\)
is a deterministic class-level quantity associated with the prescribed
bounded-parameter velocity class and its regularity regime, rather than with a
particular random realization of the stochastic-gradient algorithm. The main
theorem is stated for the exact learned ODE flow. If a numerical ODE solver is
used in implementation, an additional solver error can be added separately by
the triangle inequality, but we do not include this term in the main
sample-complexity statement.

\subsection{Neural Velocity Class}
\label{subsec:neural_vel_class}

The learned velocity field \(
    u_\theta:
    \mathbb{R}^d
    \times
    [0,1]
    \to
    \mathbb{R}^d
\) is parameterized by a fully connected feedforward network taking
\(
    (x,t)
    \in
    \mathbb{R}^{d+1}
\)
as input and returning a vector in \(\mathbb{R}^d\). Let the network have depth
\(D\), hidden width \(W\), and parameter count
\[
    P
    =
    W(d+1)
    +
    W
    +
    (D-2)(W^2+W)
    +
    dW
    +
    d.
\]
Thus,
\[
    P
    =
    O(DW^2+dW).
\]

The parameter set \(
    \Theta
    \subset
    \mathbb{R}^P
\)
is a nonempty compact convex subset of the finite-dimensional parameter space.
Compactness is used to obtain uniform bounds on the network output, its spatial
Lipschitz constants, and the empirical Hessian on a fixed finite dataset.
Convexity ensures that line segments between points in \(\Theta\) remain inside
the parameter region on which empirical smoothness is imposed. By
Lemma~\ref{lemma:automatic_containment}, projected SGD ensures that these
bounded-parameter properties apply to every optimization iterate and to the
final learned parameter \(\theta_T\).

Unlike the ReLU case, which is only piecewise smooth as a function of the
parameters, we work with a smooth activation function
\[
    \sigma:
    \mathbb{R}
    \to
    \mathbb{R}.
\]
Specifically, we assume
\[
    \sigma
    \in
    C^2(\mathbb{R}),
    \qquad
    \sup_{z\in\mathbb{R}}
    |\sigma'(z)|
    \le
    M_1,
\]
\[
    \sup_{z\in\mathbb{R}}
    |\sigma''(z)|
    \le
    M_2,
    \qquad
    |\sigma(0)|
    \le
    M_0.
\]
Softplus and \(\tanh\) are canonical examples satisfying these conditions. This
smooth-activation setup makes the empirical loss differentiable in the
parameters and is therefore compatible with finite-time projected-SGD analysis
under empirical smoothness. The empirical proximal-PL condition remains a
separate optimization assumption.

We use two standard consequences of this bounded-parameter smooth network class.
First, the class has uniform linear growth. There exist constants \(
    A_{\mathrm{NN}},
    B_{\mathrm{NN}}
    >
    0,
\) depending only on the architecture, the parameter bound, and the activation
constants, such that, for every

\(  \theta
    \in
    \Theta,
    x
    \in
    \mathbb{R}^d,
    t
    \in
    [0,1],
\)
we have
\[
    \|u_\theta(x,t)\|_2
    \le
    A_{\mathrm{NN}}
    +
    B_{\mathrm{NN}}
    \|x\|_2.
\]
This fact is proved in Lemma~\ref{lemma:nn_linear_growth}. It is used to control
both the loss on the endpoint-truncation event and the contribution of the
sub-exponential tail region. We do not require a dimension-free approximation
theorem for the neural velocity class.

Second, we use a metric-entropy uniform-convergence bound for the induced
truncated squared-loss class. For a sample
\[
    Z
    =
    (X_0,X_1,t),
\]
define
\[
    X_t
    =
    (1-t)X_0+tX_1,
    \qquad
    \dot X_t
    =
    X_1-X_0,
\]
and
\[
    \ell_\theta(Z)
    :=
    \|u_\theta(X_t,t)-\dot X_t\|_2^2.
\]
For
\(
    R
    >
    0,
\)
let
\[
    \mathcal{E}_R
    :=
    \left\{
        \|X_0\|_2
        \le
        R,
        \,
        \|X_1\|_2
        \le
        R
    \right\}
\]
and define the truncated squared-loss class
\[
    \mathcal{G}_{W,D}^{(R)}
    :=
    \left\{
        Z
        \mapsto
        \ell_\theta(Z)
        \mathbf{1}_{\mathcal{E}_R}
        :
        \theta
        \in
        \Theta
    \right\}.
\]
On \(\mathcal{E}_R\), the linear path satisfies
\[
    \|X_t\|_2
    \le
    R,
    \qquad
    \|\dot X_t\|_2
    =
    \|X_1-X_0\|_2
    \le
    2R.
\]
Together with the uniform linear-growth bound, this implies
\[
    0
    \le
    \ell_\theta(Z)
    \mathbf{1}_{\mathcal{E}_R}
    \le
    B_\ell(R),
\]
where
\[
    B_\ell(R)
    :=
    \left(
        A_{\mathrm{NN}}
        +
        (B_{\mathrm{NN}}+2)R
    \right)^2.
\]

The relevant statistical complexity object is the scalar truncated squared-loss
class \(
    \mathcal{G}_{W,D}^{(R)},
\)
not merely the raw vector-valued velocity class. For bounded-parameter smooth
networks, the map
\[
    \theta
    \mapsto
    \ell_\theta(Z)
    \mathbf{1}_{\mathcal{E}_R}
\]
is uniformly Lipschitz over \(\Theta\), uniformly for samples satisfying
\(\mathcal{E}_R\). Since
\(
    \Theta
    \subset
    \mathbb{R}^P
\)
is bounded, this yields the covering-number bound
\[
    \log
    \mathcal{N}
    \left(
        \varepsilon,
        \mathcal{G}_{W,D}^{(R)},
        \|\cdot\|_\infty
    \right)
    \le
    P
    \log
    \left(
        \frac{
            C_{\mathrm{sm}}(R)
        }{
            \varepsilon
        }
    \right),
    \qquad
    0
    <
    \varepsilon
    \le
    B_\ell(R),
\]
where
\[
    \max
    \left\{
        e,
        B_\ell(R)
    \right\}
    \le
    C_{\mathrm{sm}}(R)
    <
    \infty
\]
depends on the architecture, parameter bound, activation derivative bounds, and
truncation radius. The detailed argument is given in
Lemma~\ref{lemma:pdim_uniform_convergence}. Although the lemma label is
retained for notational continuity, the result used here is a metric-entropy
uniform-convergence bound rather than a ReLU pseudo-dimension bound.

Define the \(N\)-dependent entropy radius
\[
    \overline C_{\mathrm{sm}}(R,N)
    :=
    e
    \frac{
        N
        C_{\mathrm{sm}}(R)
    }{
        B_\ell(R)
    }.
\]
Since
\[
    C_{\mathrm{sm}}(R)
    \ge
    B_\ell(R),
\]
we have
\[
    \overline C_{\mathrm{sm}}(R,N)
    \ge
    e.
\]
Define the corresponding smooth-network complexity quantity by
\[
    \mathcal C_{\mathrm{sm}}(R,N)
    :=
    P
    \log
    \left(
        \overline C_{\mathrm{sm}}(R,N)
    \right).
\]
With this notation, the statistical term appearing in the main theorem has the
form
\[
    C_{\mathrm{stat}}
    B_\ell(R)
    \sqrt{
        \frac{
            \mathcal C_{\mathrm{sm}}(R,N)
            +
            \log(4/\delta)
        }{
            N
        }
    }.
\]

\subsection{Population and Empirical Objectives}


The population flow-matching loss is
\[
    L(\theta)
    :=
    \mathbb{E}
    \left[
        \ell_\theta(Z)
    \right].
\]
Equivalently,
\[
    L(\theta)
    =
    \int_0^1
    \mathbb{E}_{(X_0,X_1)\sim\Gamma}
    \left[
        \left\|
            u_\theta
            \left(
                (1-t)X_0+tX_1,
                t
            \right)
            -
            (X_1-X_0)
        \right\|_2^2
    \right]
    dt.
\]

Given an independent finite training dataset
\[
    \mathcal{D}_N
    :=
    \{Z_i\}_{i=1}^N,
    \qquad
    Z_i
    =
    (X_{0,i},X_{1,i},t_i),
\]
where
\[
    (X_{0,i},X_{1,i})
    \sim
    \Gamma,
    \qquad
    t_i
    \sim
    \operatorname{Uniform}[0,1],
\]
the empirical flow-matching loss is
\[
    \widehat L_N(\theta)
    :=
    \frac{1}{N}
    \sum_{i=1}^N
    \ell_\theta(Z_i).
\]
That is,
\[
    \widehat L_N(\theta)
    =
    \frac{1}{N}
    \sum_{i=1}^N
    \left\|
        u_\theta
        \left(
            (1-t_i)X_{0,i}+t_iX_{1,i},
            t_i
        \right)
        -
        (X_{1,i}-X_{0,i})
    \right\|_2^2.
\]

The parameter
\[
    \widehat\theta_N
    \in
    \arg\min_{\theta\in\Theta}
    \widehat L_N(\theta)
\]
is used only as a theoretical comparator and is not assumed to be available to
the algorithm. The learned parameter \(\theta_T\) is instead obtained after
\(T\) projected stochastic-gradient updates on \(\widehat L_N\). Thus,
\(N\) denotes the number of training samples used for statistical control, while
\(T\) denotes the number of stochastic-gradient updates used to approximately
minimize the fixed empirical objective.

\subsection{Projection Identity}

The regression target in the flow-matching loss is the pathwise velocity
\[
    \dot X_t
    =
    X_1-X_0,
\]
which depends on the coupled endpoint pair. The deployable target velocity is the
conditional mean
\[
    u^\star(x,t)
    :=
    \mathbb{E}
    \left[
        \dot X_t
        \,\middle|\,
        X_t=x,
        \,
        t
    \right].
\]

Let
\[
    L^\star
    :=
    \inf_u
    \mathbb{E}
    \left[
        \|u(X_t,t)-\dot X_t\|_2^2
    \right],
\]
where the infimum is over all measurable deployable velocity fields
\[
    u:
    \mathbb{R}^d
    \times
    [0,1]
    \to
    \mathbb{R}^d.
\]
By the \(L^2\) projection identity, \(u^\star\) minimizes this population risk
over all measurable deployable velocity fields. Moreover, for every \(
    \theta
    \in
    \Theta,
\) we have
\[
    L(\theta)
    -
    L^\star
    =
    \mathbb{E}
    \left[
        \|u_\theta(X_t,t)-u^\star(X_t,t)\|_2^2
    \right].
\]
Thus, the excess population flow-matching loss is exactly the integrated squared
error between the learned deployable velocity field and the target marginal
velocity field along the reference path.

\subsection{From Velocity Error to Distributional Error}

Let
\(
    \widehat\pi^\theta_1
\)
denote the terminal law generated by the exact learned ODE driven by
\(
    u_\theta:
\)
\[
    \frac{d}{dt}
    Y_t^\theta
    =
    u_\theta(Y_t^\theta,t),
    \qquad
    Y_0^\theta
    \sim
    \pi_0.
\]
Let
\(
    Z_t
\)
denote the reference flow driven by
\(
    u^\star:
\)
\[
    \frac{d}{dt}
    Z_t
    =
    u^\star(Z_t,t),
    \qquad
    Z_0
    \sim
    \pi_0.
\]
By the path-consistency clause in Assumption~\ref{assump:flow_stability},
\[
    \operatorname{Law}(Z_t)
    =
    \rho_t
    \qquad
    \text{for every }
    t
    \in
    [0,1].
\]
In particular,
\[
    \operatorname{Law}(Z_1)
    =
    \pi_1.
\]

For every \(
    \theta
    \in
    \Theta,
\) Assumption~\ref{assump:flow_stability} provides the deterministic uniform bound
\[
    \int_0^1
    L_\theta(t)
    \,dt
    \le
    \Lambda_{\mathrm{flow}},
\]
and defines
\[
    K_{\mathrm{flow}}
    :=
    \exp
    \left(
        \Lambda_{\mathrm{flow}}
    \right).
\]
The uniform flow-stability argument then gives
\[
    W_2
    \left(
        \widehat\pi^\theta_1,
        \pi_1
    \right)
    \le
    K_{\mathrm{flow}}
    \left(
        \int_0^1
        \mathbb{E}
        \left[
            \|u_\theta(Z_t,t)-u^\star(Z_t,t)\|_2^2
        \right]
        dt
    \right)^{1/2}.
\]
Since
\(
    Z_t
    \sim
    \rho_t,
\)
this can be written equivalently as
\[
    W_2
    \left(
        \widehat\pi^\theta_1,
        \pi_1
    \right)
    \le
    K_{\mathrm{flow}}
    \left(
        \mathbb{E}
        \left[
            \|u_\theta(X_t,t)-u^\star(X_t,t)\|_2^2
        \right]
    \right)^{1/2}.
\]
By the projection identity,
\[
    W_2
    \left(
        \widehat\pi^\theta_1,
        \pi_1
    \right)
    \le
    K_{\mathrm{flow}}
    \left(
        L(\theta)-L^\star
    \right)^{1/2},
    \qquad
    \forall
    \theta
    \in
    \Theta.
\]

The factor \(
    K_{\mathrm{flow}}
\) is a uniform stability amplification constant for the prescribed learned
velocity-field class. It is treated as an explicit problem-dependent regularity
quantity; we do not claim that it is dimension free or automatically bounded
uniformly over arbitrary neural architectures.

\subsection{Error Decomposition}

Let \(\theta_T\) denote the parameter obtained after \(T\) projected
stochastic-gradient updates on the finite empirical objective
\(
    \widehat L_N.
\)
By Lemma~\ref{lemma:automatic_containment},
\(
    \theta_T
    \in
    \Theta.
\)
Let
\(
    \widehat\theta_N
    \in
    \arg\min_{\theta\in\Theta}
    \widehat L_N(\theta)
\)
be an empirical risk minimizer, used only as a theoretical comparator. Also let
\(
    \theta^\star_\Theta
    \in
    \arg\min_{\theta\in\Theta}
    L(\theta)
\)
be a population minimizer within the prescribed neural velocity class. Then
\[
\begin{aligned}
    L(\theta_T)
    -
    L^\star
    &\le
    L(\theta^\star_\Theta)
    -
    L^\star
    +
    2
    \sup_{\theta\in\Theta}
    \left|
        L(\theta)
        -
        \widehat L_N(\theta)
    \right|
    \\
    &\quad
    +
    \widehat L_N(\theta_T)
    -
    \widehat L_N(\widehat\theta_N).
\end{aligned}
\]
Define
\[
    E_{\mathrm{approx}}
    :=
    L(\theta^\star_\Theta)
    -
    L^\star,
\]
\[
    E_{\mathrm{stat}}
    :=
    2
    \sup_{\theta\in\Theta}
    \left|
        L(\theta)
        -
        \widehat L_N(\theta)
    \right|,
\]
and
\[
    E_{\mathrm{opt}}
    :=
    \widehat L_N(\theta_T)
    -
    \widehat L_N(\widehat\theta_N).
\]
Then
\[
    L(\theta_T)
    -
    L^\star
    \le
    E_{\mathrm{approx}}
    +
    E_{\mathrm{stat}}
    +
    E_{\mathrm{opt}}.
\]

The approximation term is controlled by Assumption~\ref{assump:approx}, the
statistical term by uniform convergence over the finite dataset, and the
optimization term by the projected empirical proximal-PL stochastic-gradient
analysis.  While the algebraic decomposition above contains approximation, statistical, and optimization terms, under unbounded data the bound on the statistical term further separates into a truncated uniform-convergence contribution and a tail contribution. Combining this decomposition with uniform flow stability gives
\[
    W_2
    \left(
        \widehat\pi^{\theta_T}_1,
        \pi_1
    \right)
    \le
    K_{\mathrm{flow}}
    \left(
        E_{\mathrm{approx}}
        +
        E_{\mathrm{stat}}
        +
        E_{\mathrm{opt}}
    \right)^{1/2}.
\]

This decomposition follows a general pattern in learning theory: one separates
the limitation of the prescribed function class, the error incurred by replacing
a population objective with a finite empirical objective, and the error incurred
by only approximately optimizing that empirical objective. In large-scale
statistical learning, related approximation--estimation--optimization tradeoffs
are used to distinguish model-class limitations, finite-sample generalization,
and incomplete numerical optimization
\citep{bottou2008tradeoffs,daneshmand2016starting}.


\subsection{Statistical Control Under Sub-Exponential Data}

For
\(
    R
    >
    0,
\)
define the endpoint-truncation event
\[
    \mathcal{E}_R
    :=
    \left\{
        \|X_0\|_2
        \le
        R,
        \,
        \|X_1\|_2
        \le
        R
    \right\}.
\]
On \(\mathcal{E}_R\), the linear path satisfies
\[
    \|X_t\|_2
    \le
    R,
    \qquad
    \|X_1-X_0\|_2
    \le
    2R.
\]
The uniform linear-growth property gives
\[
    \ell_\theta(Z)
    \le
    B_\ell(R)
    :=
    \left(
        A_{\mathrm{NN}}
        +
        (B_{\mathrm{NN}}+2)R
    \right)^2
\]
on \(\mathcal{E}_R\). Define the tail contribution by
\[
    \epsilon_{\mathrm{tail}}(R)
    :=
    \sup_{\theta\in\Theta}
    \mathbb{E}
    \left[
        \ell_\theta(Z)
        \mathbf{1}_{\mathcal{E}_R^c}
    \right].
\]
The linear-growth property implies that there exists a constant \( C > 0
\) such that
\[
    \ell_\theta(Z)
    \le
    C
    \left(
        1
        +
        \|X_0\|_2^2
        +
        \|X_1\|_2^2
    \right)
\]
uniformly over
\(
    \theta
    \in
    \Theta.
\) Since
\[
    \mathcal{E}_R^c
    \subseteq
    \left\{
        \|X_0\|_2
        >
        R
    \right\}
    \cup
    \left\{
        \|X_1\|_2
        >
        R
    \right\},
\]
the endpoint-tail calculation in Lemma~\ref{lemma:tail_control_subexp} gives
\[
    \epsilon_{\mathrm{tail}}(R)
    \le
    C_{\mathrm{tail}}
    (1+R)^2
    \exp
    \left(
        -c_{\mathrm{tail}}R/K_X
    \right).
\]

The metric-entropy uniform-convergence result for the truncated squared-loss class
gives the statistical contribution
\[
\begin{aligned}
    E_{\mathrm{stat}}(N,R,\delta)
    &:=
    C_{\mathrm{stat}}
    B_\ell(R)
    \sqrt{
        \frac{
            \mathcal C_{\mathrm{sm}}(R,N)
            +
            \log(4/\delta)
        }{
            N
        }
    }
    \\
    &\quad
    +
    C_{\mathrm{tail}}
    (1+R)^2
    \exp
    \left(
        -c_{\mathrm{tail}}R/K_X
    \right).
\end{aligned}
\]
For every truncation radius satisfying
\[
    R
    \ge
    \frac{K_X}{c_1}
    \log
    \left(
        \frac{8Nc_0}{\delta}
    \right),
\]
with probability at least
\[
    1-\delta
\]
over the draw of \(\mathcal{D}_N\), we have
\[
    2
    \sup_{\theta\in\Theta}
    \left|
        L(\theta)
        -
        \widehat L_N(\theta)
    \right|
    \le
    E_{\mathrm{stat}}(N,R,\delta),
\]
where the leading factor \(2\) from the error decomposition has been absorbed
into the constants.

Choosing
\[
    R
    \asymp
    K_X
    \log
    \left(
        N/\delta
    \right)
\]
makes the tail contribution lower order up to logarithmic factors. Thus, the
finite-dataset statistical contribution scales as
\[
    \widetilde O
    \left(
        \sqrt{
            \frac{
                \mathcal C_{\mathrm{sm}}(R,N)
            }{
                N
            }
        }
    \right)
\]
at the squared-loss level, up to the truncation-dependent envelope and other
logarithmic factors.

\subsection{Optimization Control}

Conditioned on the finite dataset
\(
    \mathcal{D}_N,
\)
the projected stochastic-gradient updates are
\[
    \theta_{k+1}
    =
    \Pi_\Theta
    \left(
        \theta_k
        -
        \eta_k g_k
    \right),
\]
\[
    g_k
    =
    \nabla_\theta
    \ell_{\theta_k}(Z_{I_k}),
    \qquad
    k
    =
    0,1,\ldots,T-1,
\]
where
\[
    I_k
    \sim
    \operatorname{Uniform}
    \left\{
        1,\ldots,N
    \right\}.
\]
By Lemma~\ref{lemma:automatic_containment},
\(
    \theta_k
    \in
    \Theta,
    \qquad
    k
    =
    0,1,\ldots,T.
\)
Because \(\Theta\) is convex and \(\widehat L_N\) is \(\alpha_N\)-smooth on
\(\Theta\), the smoothness descent inequality applies along every update
segment.

For the projected recursion, define the stochastic gradient-mapping quantity
\[
    G_k
    :=
    \frac{1}{\eta_k}
    \left(
        \theta_k
        -
        \Pi_\Theta
        \left(
            \theta_k
            -
            \eta_k g_k
        \right)
    \right).
\]
Under the
interior operating regime in Assumption~\ref{assump:empirical_sgd}, the
projection is inactive along the relevant trajectory, so that
\(
    G_k
    =
    g_k
\) and the resulting one-step recursion agrees with the standard empirical
PL-SGD recursion:
\[
    e_{k+1}
    \le
    \left(
        1-\mu_N\eta_k
    \right)
    e_k
    +
    \frac{
        \alpha_N\sigma_N^2
    }{
        2
    }
    \eta_k^2,
\]
where
\[
    e_k
    :=
    \mathbb{E}_{\mathrm{SGD}}
    \left[
        \widehat L_N(\theta_k)
        -
        \widehat L_N(\widehat\theta_N)
        \,\middle|\,
        \mathcal{D}_N
    \right].
\]

We use a dataset-independent decreasing-stepsize choice based on the uniform
constants in Assumption~\ref{assump:uniform_empirical_constants}:
\[
    \alpha_0
    :=
    \frac{2}{\mu_{\min}},
    \qquad
    \gamma
    :=
    \alpha_0
    \alpha_{\max},
    \qquad
    \eta_k
    :=
    \frac{\alpha_0}{k+\gamma}.
\]
On the event \(\mathcal{G}_N\), we have
\[
    \alpha_0\mu_N
    \ge
    \alpha_0\mu_{\min}
    =
    2
    >
    1,
\]
\[
    \gamma
    =
    \alpha_0\alpha_{\max}
    \ge
    \alpha_0\mu_N,
\]
and
\[
    \eta_k
    \le
    \frac{\alpha_0}{\gamma}
    =
    \frac{1}{\alpha_{\max}}
    \le
    \frac{1}{\alpha_N}.
\]
Thus,
\[
    0
    \le
    1-\mu_N\eta_k
    \le
    1
\]
for every
\[
    k
    \ge
    0.
\]

Lemma~\ref{lemma:opt_error} shows that, conditional on
\(\mathcal{D}_N\), the empirical optimization error satisfies
\[
    \mathbb{E}_{\mathrm{SGD}}
    \left[
        \widehat L_N(\theta_T)
        -
        \widehat L_N(\widehat\theta_N)
        \,\middle|\,
        \mathcal{D}_N
    \right]
    \le
    \frac{
        C_{\mathrm{opt}}(\mathcal{D}_N)
    }{
        T
    },
\]
where, up to universal numerical constants,
\[
    C_{\mathrm{opt}}(\mathcal{D}_N)
    \le
    \max
    \left\{
        \gamma e_0,
        \frac{
            \alpha_N
            \sigma_N^2
            \alpha_0^2
        }{
            2
            \left(
                \alpha_0\mu_N-1
            \right)
        }
    \right\}.
\]
On the event \(\mathcal{G}_N\), this quantity is bounded by the deterministic
envelope
\[
    \overline C_{\mathrm{opt}}
    :=
    \max
    \left\{
        \gamma\overline e_0,
        \frac{
            \alpha_{\max}
            \sigma_{\max}^2
            \alpha_0^2
        }{
            2
            \left(
                \alpha_0\mu_{\min}-1
            \right)
        }
    \right\}.
\]
Since
\[
    \alpha_0
    =
    \frac{2}{\mu_{\min}},
    \qquad
    \gamma
    =
    \frac{
        2\alpha_{\max}
    }{
        \mu_{\min}
    },
\]
we may equivalently write
\[
    \overline C_{\mathrm{opt}}
    =
    \max
    \left\{
        \frac{
            2\alpha_{\max}
        }{
            \mu_{\min}
        }
        \overline e_0,
        \frac{
            2\alpha_{\max}
            \sigma_{\max}^2
        }{
            \mu_{\min}^2
        }
    \right\}.
\]
Therefore, on \(\mathcal{G}_N\),
\[
    \mathbb{E}_{\mathrm{SGD}}
    \left[
        \widehat L_N(\theta_T)
        -
        \widehat L_N(\widehat\theta_N)
        \,\middle|\,
        \mathcal{D}_N
    \right]
    \le
    \frac{
        \overline C_{\mathrm{opt}}
    }{
        T
    }.
\]

The smooth-activation setup makes differentiability and empirical smoothness
compatible with bounded parameter finite dataset training. The empirical
proximal-PL condition, the uniform-in- \(N\) regularity condition, and the
interior operating regime remain optimization assumptions; they are not implied
by smoothness alone. The main theorem uses the conditional expectation guarantee
above. A fully high-probability optimization result would require additional
martingale-tail assumptions and a separate concentration analysis.

\subsection{Summary of the Proof Strategy}

The preceding subsections identify the ingredients required for the finite-dataset
analysis. The approximation condition controls the best achievable population
error within the prescribed neural velocity class. The statistical argument controls
the discrepancy between the population objective and the fixed empirical objective
through endpoint truncation, sub-exponential tail control, and metric-entropy
uniform convergence. The projected empirical proximal-PL stochastic-gradient
analysis controls the error from running only a finite number of updates on that
fixed empirical objective.

More precisely, the error decomposition in Section~3.6 separates the excess
population flow-matching loss into an approximation contribution, a finite-dataset
statistical contribution, and an empirical optimization contribution. The
endpoint-tail term arises when bounding the statistical contribution under
unbounded source and target distributions. The optimization term is controlled on
the uniform regularity event \(
    \mathcal{G}_N
\) from Assumption~\ref{assump:uniform_empirical_constants}. The deterministic
uniform flow-stability constant from Assumption~\ref{assump:flow_stability} then
converts the resulting integrated velocity-field error into a Wasserstein-2 bound
for the terminal distribution generated by the exact learned ODE.

The formal finite-sample and finite-time guarantee combines the statistical event
from Lemma~\ref{lemma:stat_error} with the event
\(
    \mathcal{G}_N.
\) Accordingly, its conclusion holds with high probability over the sampled finite
dataset and in conditional expectation over the stochastic-gradient randomness.
The resulting theorem and sample-complexity consequence are stated in
Theorem~\ref{thm:main_flow_matching} in
Section~\ref{sec:theoretical_results}.

%% file: Sections/theory-clean.tex
\section{Theoretical Results}
\label{sec:theoretical_results}

In this section, we state the main finite-sample and finite-time guarantee for
flow matching under sub-exponential data and projected finite-dataset
stochastic-gradient training. The theorem combines the approximation condition,
the metric-entropy statistical bound for the smooth neural velocity class, the
empirical projected-optimization bound, and the path-consistent uniform
flow-stability argument described in the previous section.

Throughout this section,
\(
    \widehat\pi^{\theta_T}_1
\)
denotes the terminal law of the exact learned ODE flow driven by the learned
velocity field
\(
    u_{\theta_T}.
\)
The dataset size
\(
    N
\)
controls the statistical error incurred by replacing the population objective
with the empirical objective
\(
    \widehat L_N,
\)
whereas
\(
    T
\)
denotes the number of projected stochastic-gradient updates used to
approximately minimize that fixed empirical objective.

The learned velocity field
\[
    u_\theta:
    \mathbb{R}^d
    \times
    [0,1]
    \to
    \mathbb{R}^d
\]
is parameterized by the bounded-parameter smooth-activation network class
described in Section~\ref{subsec:neural_vel_class}. The network has depth
\(
    D,
\)
hidden width
\(
    W,
\)
and parameter count
\[
    P
    =
    W(d+1)
    +
    W
    +
    (D-2)(W^2+W)
    +
    dW
    +
    d.
\]
Thus,
\[
    P
    =
    O(DW^2+dW).
\]

For
\(
    R
    >
    0,
\)
define the endpoint-truncation event
\[
    \mathcal E_R
    :=
    \left\{
        \|X_0\|_2
        \le
        R,
        \,
        \|X_1\|_2
        \le
        R
    \right\}.
\]
On this event, the single-sample flow-matching loss is bounded by
\[
    B_\ell(R)
    :=
    \left(
        A_{\mathrm{NN}}
        +
        (B_{\mathrm{NN}}+2)R
    \right)^2,
\]
where
\(
    A_{\mathrm{NN}},
    B_{\mathrm{NN}}
\)
are the constants from the uniform linear-growth property of the
bounded-parameter smooth neural velocity class.

Recall that
\[
    \max
    \left\{
        e,
        B_\ell(R)
    \right\}
    \le
    C_{\mathrm{sm}}(R)
    <
    \infty
\]
denotes the raw metric-entropy radius of the truncated squared-loss class.
Define the corresponding \(N\)-dependent entropy radius by
\[
    \overline C_{\mathrm{sm}}(R,N)
    :=
    e
    \frac{
        N
        C_{\mathrm{sm}}(R)
    }{
        B_\ell(R)
    }.
\]
Since
\[
    C_{\mathrm{sm}}(R)
    \ge
    B_\ell(R),
\]
we have
\[
    \overline C_{\mathrm{sm}}(R,N)
    \ge
    e.
\]
The associated smooth-network complexity quantity is
\[
    \mathcal C_{\mathrm{sm}}(R,N)
    :=
    P
    \log
    \left(
        \overline C_{\mathrm{sm}}(R,N)
    \right).
\]
This is the complexity quantity appearing in the theorem and in
Lemma~\ref{lemma:pdim_uniform_convergence}. Under Assumption~\ref{assump:flow_stability}, the quantity
\(
    K_{\mathrm{flow}}
    =
    \exp
    \left(
        \Lambda_{\mathrm{flow}}
    \right)
\) is a deterministic uniform flow-stability constant satisfying
\[
    \sup_{\theta\in\Theta}
    \int_0^1
    L_\theta(t)
    \,dt
    \le
    \Lambda_{\mathrm{flow}}.
\]
By Lemma~\ref{lemma:automatic_containment}, projected SGD guarantees that every
optimization iterate, including the final learned parameter
\(
    \theta_T,
\)
belongs to
\(
    \Theta.
\)
Hence the same deterministic flow-stability constant applies to every iterate
and to the final learned velocity field.

\begin{theorem}[Finite-sample and finite-time guarantee for flow matching]
\label{thm:main_flow_matching}

Suppose Assumptions~\ref{assump:subexp}, \ref{assump:approx},
\ref{assump:empirical_pl}, \ref{assump:empirical_sgd},
\ref{assump:uniform_empirical_constants}, and
\ref{assump:flow_stability} hold. Suppose that the velocity field
\(
    u_\theta
\)
is parameterized by the bounded-parameter smooth-activation network class
described in Section~\ref{subsec:neural_vel_class}.

Let
\(
    \theta_T
\)
be the output of the projected finite-dataset stochastic-gradient recursion in
Assumption~\ref{assump:empirical_sgd}, with the decreasing stepsizes
\[
    \alpha_0
    :=
    \frac{2}{\mu_{\min}},
    \qquad
    \gamma
    :=
    \alpha_0\alpha_{\max},
    \qquad
    \eta_k
    :=
    \frac{\alpha_0}{k+\gamma}.
\]

For
\[
    \delta,
    \delta_{\mathrm{opt}}
    \in
    (0,1)
\]
and any truncation radius satisfying
\[
    R
    \ge
    \frac{K_X}{c_1}
    \log
    \left(
        \frac{8Nc_0}{\delta}
    \right),
\]
define
\[
\begin{aligned}
    \mathfrak E_{N,T}
    \left(
        R,
        \delta
    \right)
    &:=
    \epsilon_{\mathrm{approx}}
    +
    C_{\mathrm{stat}}
    B_\ell(R)
    \sqrt{
        \frac{
            \mathcal C_{\mathrm{sm}}(R,N)
            +
            \log(4/\delta)
        }{
            N
        }
    }
    \\
    &\quad
    +
    C_{\mathrm{tail}}
    (1+R)^2
    \exp
    \left(
        -c_{\mathrm{tail}}R/K_X
    \right)
    +
    \frac{
        \overline C_{\mathrm{opt}}
    }{
        T
    }.
\end{aligned}
\]

Then, with probability at least
\[
    1
    -
    \delta
    -
    \delta_{\mathrm{opt}}
\]
over the draw of the finite dataset
\[
    \mathcal{D}_N,
\]
the following conditional expected excess-risk bound holds:
\[
    \mathbb{E}_{\mathrm{SGD}}
    \left[
        L(\theta_T)
        -
        L^\star
        \,\middle|\,
        \mathcal{D}_N
    \right]
    \le
    \mathfrak E_{N,T}
    \left(
        R,
        \delta
    \right).
\]

Equivalently, by the projection identity,
\[
    \mathbb{E}_{\mathrm{SGD}}
    \left[
        \mathbb{E}
        \left[
            \|u_{\theta_T}(X_t,t)-u^\star(X_t,t)\|_2^2
        \right]
        \,\middle|\,
        \mathcal{D}_N
    \right]
    \le
    \mathfrak E_{N,T}
    \left(
        R,
        \delta
    \right).
\]

Consequently, the terminal distribution of the exact learned ODE flow satisfies
\[
    \mathbb{E}_{\mathrm{SGD}}
    \left[
        W_2^2
        \left(
            \widehat{\pi}^{\theta_T}_1,
            \pi_1
        \right)
        \,\middle|\,
        \mathcal{D}_N
    \right]
    \le
    K_{\mathrm{flow}}^2
    \mathfrak E_{N,T}
    \left(
        R,
        \delta
    \right).
\]
By Jensen's inequality,
\[
    \mathbb{E}_{\mathrm{SGD}}
    \left[
        W_2
        \left(
            \widehat{\pi}^{\theta_T}_1,
            \pi_1
        \right)
        \,\middle|\,
        \mathcal{D}_N
    \right]
    \le
    K_{\mathrm{flow}}
    \mathfrak E_{N,T}
    \left(
        R,
        \delta
    \right)^{1/2}.
\]
\end{theorem}

\begin{remark}[Exact learned flow versus numerical sampling]
\label{remark:exact_flow}

Theorem~\ref{thm:main_flow_matching} is stated for the exact learned ODE flow.
This matches the main theoretical focus of the paper: controlling the
finite-dataset statistical and finite-time optimization errors incurred in
learning the velocity field. If the learned ODE is solved numerically in
implementation, a separate solver error can be added by the triangle inequality.
We do not include this numerical term in the main sample-complexity statement.
\end{remark}

\begin{remark}[Sample-complexity consequence]
\label{remark:sample_complexity}

Choose
\[
    R
    =
    c_R
    K_X
    \log
    \left(
        N/\delta
    \right),
\]
where
\(
    c_R
    >
    0
\)
is sufficiently large that the truncation condition in
Theorem~\ref{thm:main_flow_matching} holds and the tail contribution
\(
    C_{\mathrm{tail}}
    (1+R)^2
    \exp
    \left(
        -c_{\mathrm{tail}}R/K_X
    \right)
\) is lower order up to logarithmic factors.

Suppose that the approximation and tail contributions satisfy
\[
    \epsilon_{\mathrm{approx}}
    +
    C_{\mathrm{tail}}
    (1+R)^2
    \exp
    \left(
        -c_{\mathrm{tail}}R/K_X
    \right)
    \le
    \frac{
        \epsilon^2
    }{
        2K_{\mathrm{flow}}^2
    }.
\]
For example, it is sufficient to require
\[
    \epsilon_{\mathrm{approx}}
    \le
    \frac{
        \epsilon^2
    }{
        4K_{\mathrm{flow}}^2
    }
\]
and to choose \(R\) sufficiently large that the tail term is at most
\(
    \frac{
        \epsilon^2
    }{
        4K_{\mathrm{flow}}^2
    }.
\) Then a sufficient dataset-size condition for
\[
    \mathbb{E}_{\mathrm{SGD}}
    \left[
        W_2
        \left(
            \widehat{\pi}^{\theta_T}_1,
            \pi_1
        \right)
        \,\middle|\,
        \mathcal{D}_N
    \right]
    \le
    O(\epsilon)
\]
on the event in Theorem~\ref{thm:main_flow_matching} is
\[
    N
    \ge
    C
    \frac{
        K_{\mathrm{flow}}^4
        B_\ell(R)^2
        \left(
            \mathcal C_{\mathrm{sm}}(R,N)
            +
            \log(1/\delta)
        \right)
    }{
        \epsilon^4
    },
\]
for a sufficiently large constant
\(
    C
    >
    0.
\) A sufficient optimization-budget condition is
\(
    T
    \ge
    C
    \frac{
        K_{\mathrm{flow}}^2
        \overline C_{\mathrm{opt}}
    }{
        \epsilon^2
    },
\)
for a sufficiently large constant
\(
    C
    >
    0
\).

Because the truncation radius R and the entropy quantity $C_{sm}(R,N)$ depend logarithmically on N, the preceding display is an implicit sufficient condition on the dataset size.

The optimization requirement is deterministic conditional on the displayed
uniform constants in Assumption~\ref{assump:uniform_empirical_constants}. In
particular,
\(
    \overline C_{\mathrm{opt}}
\)
does not depend on the realized dataset once
\(
    \mu_{\min},
    \alpha_{\max},
    \sigma_{\max}^2,
    \overline e_0
\) have been fixed. When the approximation, tail, flow-stability, smooth-network-complexity, and
uniform empirical-optimization quantities are treated as fixed or displayed
separately, the shorthand scaling is
\(
    N
    =
    \widetilde O
    \left(
        \epsilon^{-4}
    \right),
    T
    =
    \widetilde O
    \left(
        \epsilon^{-2}
    \right).
\)
\end{remark}

\begin{remark}[Status of the displayed constants]
\label{remark:displayed_constants}

The bound is stated modularly. The quantities
\(
    \epsilon_{\mathrm{approx}},
    K_{\mathrm{flow}}
    =
    \exp
    \left(
        \Lambda_{\mathrm{flow}}
    \right),
\)
and the uniform empirical constants
\(
    \mu_{\min},
    \alpha_{\max},
    \sigma_{\max}^2,
    \overline e_0,
\)
which enter through
\(
    \overline C_{\mathrm{opt}},
\)
are displayed problem-dependent constants and are not derived in this work.

In particular,
\(
    \epsilon_{\mathrm{approx}}
\)
is not claimed to vanish for an arbitrary fixed architecture. The factor
\(
    K_{\mathrm{flow}}
\)
is a deterministic class-level amplification factor in the terminal
Wasserstein bound; it is not claimed to be dimension free or uniformly bounded
over arbitrary neural architectures or parameter balls. Finally, the shorthand
optimization scaling
\(
    T
    =
    \widetilde O
    \left(
        \epsilon^{-2}
    \right)
\)
is conditional on the uniform lower bound
\(
    \mu_N
    \ge
    \mu_{\min}
\) and the other uniform empirical-constant bounds in
Assumption~\ref{assump:uniform_empirical_constants}.

The contribution of the theorem is an explicit reduction of terminal
Wasserstein-2 error to these displayed approximation, statistical,
optimization, and flow-stability quantities.
\end{remark}

\begin{remark}[High-probability optimization variant]
\label{remark:hp_optimization}

Theorem~\ref{thm:main_flow_matching} uses the optimization guarantee implied by
the bounded conditional-variance assumption in
Assumption~\ref{assump:empirical_sgd}. Accordingly, its conclusion holds with
high probability over the finite dataset and in conditional expectation over the
stochastic-gradient randomness.

A fully high-probability optimization guarantee would require additional explicit
assumptions on the stochastic gradient noise, together with a separate
martingale-concentration analysis of the resulting recursion. For example, one
could impose a conditional Bernstein-type or conditionally sub-Gaussian
martingale-difference assumption. We do not require such a strengthening here
and therefore do not state a high-probability optimization rate in the main
result.
\end{remark}

The proof of Theorem~\ref{thm:main_flow_matching} follows by combining the next
three lemmas.

\begin{lemma}[Approximation error]
\label{lemma:approx_error}

Under Assumption~\ref{assump:approx},
\[
    L(\theta^\star_\Theta)
    -
    L^\star
    \le
    \epsilon_{\mathrm{approx}}.
\]
\end{lemma}

\begin{proof}
This is exactly Assumption~\ref{assump:approx}. By the projection identity,
\[
    L(\theta^\star_\Theta)
    -
    L^\star
    =
    \mathbb{E}
    \left[
        \|u_{\theta^\star_\Theta}(X_t,t)-u^\star(X_t,t)\|_2^2
    \right].
\]
Therefore,
\[
    L(\theta^\star_\Theta)
    -
    L^\star
    \le
    \epsilon_{\mathrm{approx}}.
\]
\end{proof}

\begin{lemma}[Statistical error under sub-exponential data]
\label{lemma:stat_error}

Suppose Assumption~\ref{assump:subexp} holds, and suppose the bounded-parameter
smooth neural velocity class has the uniform linear-growth and metric-entropy
properties stated in Section~\ref{subsec:neural_vel_class}. For
\(
    R
    >
    0,
\)
define
\[
    \mathcal{E}_R
    :=
    \left\{
        \|X_0\|_2
        \le
        R,
        \,
        \|X_1\|_2
        \le
        R
    \right\},
\]
and
\[
    B_\ell(R)
    :=
    \left(
        A_{\mathrm{NN}}
        +
        (B_{\mathrm{NN}}+2)R
    \right)^2.
\]
If
\[
    R
    \ge
    \frac{K_X}{c_1}
    \log
    \left(
        \frac{8Nc_0}{\delta}
    \right),
\]
then, with probability at least
\(
    1-\delta
\)
over the draw of
\(
    \mathcal{D}_N,
\)
we have
\[
\begin{aligned}
    2
    \sup_{\theta\in\Theta}
    \left|
        L(\theta)
        -
        \widehat L_N(\theta)
    \right|
    &\le
    C_{\mathrm{stat}}
    B_\ell(R)
    \sqrt{
        \frac{
            \mathcal C_{\mathrm{sm}}(R,N)
            +
            \log(4/\delta)
        }{
            N
        }
    }
    \\
    &\quad
    +
    C_{\mathrm{tail}}
    (1+R)^2
    \exp
    \left(
        -c_{\mathrm{tail}}R/K_X
    \right).
\end{aligned}
\]

Choosing
\(
    R
    =
    c_R
    K_X
    \log
    \left(
        N/\delta
    \right)
\)
with sufficiently large
\(
    c_R
    >
    0
\) makes the tail contribution lower order up to logarithmic factors. Thus,
\[
    2
    \sup_{\theta\in\Theta}
    \left|
        L(\theta)
        -
        \widehat L_N(\theta)
    \right|
    =
    \widetilde O
    \left(
        B_\ell(R)
        \sqrt{
            \frac{
                \mathcal C_{\mathrm{sm}}(R,N)
                +
                \log(1/\delta)
            }{
                N
            }
        }
    \right),
\]
up to logarithmic factors arising from the truncation radius and
smooth-network entropy radius.
\end{lemma}

The proof of Lemma~\ref{lemma:stat_error} is given in
Appendix~\ref{app:proof_stat_error}. The proof uses endpoint truncation,
sub-exponential tail control, and the metric-entropy/Rademacher
uniform-convergence bound for the smooth-network truncated squared-loss class.

\begin{lemma}[Empirical optimization error]
\label{lemma:opt_error}

Let
\[
    \widehat\theta_N
    \in
    \arg\min_{\theta\in\Theta}
    \widehat L_N(\theta)
\]
be an empirical risk minimizer. Let
\(
    \theta_T
\)
be obtained after \(T\) projected stochastic-gradient updates on the fixed
finite empirical objective:
\[
    \theta_{k+1}
    =
    \Pi_\Theta
    \left(
        \theta_k
        -
        \eta_k g_k
    \right),
\]
\[
    g_k
    =
    \nabla_\theta
    \ell_{\theta_k}(Z_{I_k}),
    \qquad
    I_k
    \sim
    \operatorname{Uniform}
    \left\{
        1,\ldots,N
    \right\}.
\]

Suppose Assumptions~\ref{assump:empirical_pl},
\ref{assump:empirical_sgd}, and
\ref{assump:uniform_empirical_constants} hold. In particular, the projected
recursion guarantees
\(
    \theta_k
    \in
    \Theta,
    \qquad
    k
    =
    0,1,\ldots,T,
\)
by Lemma~\ref{lemma:automatic_containment}, and the clean recursion below uses
the interior operating regime in Assumption~\ref{assump:empirical_sgd}. On the event
\(
    \mathcal{G}_N,
\) choose
\[
    \alpha_0
    :=
    \frac{2}{\mu_{\min}},
    \qquad
    \gamma
    :=
    \alpha_0\alpha_{\max},
    \qquad
    \eta_k
    :=
    \frac{\alpha_0}{k+\gamma}.
\]
Then
\(
    \alpha_0\mu_N
    >
    1,
    \qquad
    \gamma
    \ge
    \alpha_0\mu_N,
    \qquad
    \eta_k
    \le
    \frac{1}{\alpha_N}
    \quad
    \text{for every }
    k
    \ge
    0.
\)

Define
\(
    e_0
    :=
    \widehat L_N(\theta_0)
    -
    \widehat L_N(\widehat\theta_N).
\)
Then, conditional on
\(
    \mathcal{D}_N,
\)
the empirical optimization error satisfies
\[
    \mathbb{E}_{\mathrm{SGD}}
    \left[
        \widehat L_N(\theta_T)
        -
        \widehat L_N(\widehat\theta_N)
        \,\middle|\,
        \mathcal{D}_N
    \right]
    \le
    \frac{
        C_{\mathrm{opt}}(\mathcal{D}_N)
    }{
        T
    },
\]
where the bound follows from 

\[
e_{T} \le \frac{C_{opt}(D_{N})}{T+\gamma} \le \frac{C_{opt}(D_{N})}{T}, T \ge 1 
\]

Moreover, on \(
    \mathcal{G}_N,
\)
\(
    C_{\mathrm{opt}}(\mathcal{D}_N)
    \le
    \overline C_{\mathrm{opt}},
\) and hence
\[
    \mathbb{E}_{\mathrm{SGD}}
    \left[
        \widehat L_N(\theta_T)
        -
        \widehat L_N(\widehat\theta_N)
        \,\middle|\,
        \mathcal{D}_N
    \right]
    \le
    \frac{
        \overline C_{\mathrm{opt}}
    }{
        T
    }.
\]
\end{lemma}

The proof of Lemma~\ref{lemma:opt_error} is given in
Appendix~\ref{app:proof_opt_error}. The proof uses the stochastic gradient
mapping associated with the projected update and the empirical proximal-PL
condition. Under the interior operating regime, the projection is inactive along
the relevant stochastic and mean-gradient steps, and the resulting recursion
reduces to
\[
    e_{k+1}
    \le
    \left(
        1-\mu_N\eta_k
    \right)
    e_k
    +
    \frac{
        \alpha_N\sigma_N^2
    }{
        2
    }
    \eta_k^2,
\]
where
\[
    e_k
    :=
    \mathbb{E}_{\mathrm{SGD}}
    \left[
        \widehat L_N(\theta_k)
        -
        \widehat L_N(\widehat\theta_N)
        \,\middle|\,
        \mathcal{D}_N
    \right].
\]
This is the same form as the unconstrained empirical PL-SGD recursion, but its
validity here follows from the projected update, automatic containment, and the
interior operating regime.

\begin{proof}[Proof sketch of Theorem~\ref{thm:main_flow_matching}]
Let
\(
    \mathcal{H}_{N,R,\delta}
\)
denote the event supplied by Lemma~\ref{lemma:stat_error}. By that lemma,
\(
    \mathbb{P}
    \left(
        \mathcal{H}_{N,R,\delta}
    \right)
    \ge
    1-\delta.
\)
By Assumption~\ref{assump:uniform_empirical_constants},
\(
    \mathbb{P}
    \left(
        \mathcal{G}_N
    \right)
    \ge
    1-\delta_{\mathrm{opt}}.
\) Hence, by the union bound,
\(
    \mathbb{P}
    \left(
        \mathcal{H}_{N,R,\delta}
        \cap
        \mathcal{G}_N
    \right)
    \ge
    1
    -
    \delta
    -
    \delta_{\mathrm{opt}}.
\)

Fix a dataset realization in
\(
    \mathcal{H}_{N,R,\delta}
    \cap
    \mathcal{G}_N.
\)
The error decomposition gives
\[
\begin{aligned}
    L(\theta_T)
    -
    L^\star
    &\le
    L(\theta^\star_\Theta)
    -
    L^\star
    +
    2
    \sup_{\theta\in\Theta}
    \left|
        L(\theta)
        -
        \widehat L_N(\theta)
    \right|
    \\
    &\quad
    +
    \widehat L_N(\theta_T)
    -
    \widehat L_N(\widehat\theta_N).
\end{aligned}
\]
The first term is bounded by Lemma~\ref{lemma:approx_error}, the second by
Lemma~\ref{lemma:stat_error}, and the third in conditional expectation by
Lemma~\ref{lemma:opt_error}. Taking conditional expectation with respect to
the stochastic-gradient randomness yields
\[
    \mathbb{E}_{\mathrm{SGD}}
    \left[
        L(\theta_T)
        -
        L^\star
        \,\middle|\,
        \mathcal{D}_N
    \right]
    \le
    \mathfrak E_{N,T}
    \left(
        R,
        \delta
    \right).
\]

By Lemma~\ref{lemma:automatic_containment},
\(
    \theta_T
    \in
    \Theta.
\)
The deterministic uniform stability inequality from
Lemma~\ref{lemma:flow_stability} may therefore be applied pointwise to
\(
    \theta_T.
\) Taking conditional expectation and using Jensen's inequality gives the stated
Wasserstein guarantees. The full proof is given in
Appendix~\ref{app:theorem_proof}.
\end{proof}

\begin{remark}[Pure population-oracle version]
\label{remark:population_oracle}

For comparison, suppose instead that the optimization algorithm has access to a
population stochastic-gradient oracle and that the corresponding population
objective satisfies suitable population proximal-PL, smoothness, and
stochastic-gradient conditions. The finite-dataset statistical term is then
absent. Under those distinct assumptions, the corresponding flow-stability
argument gives
\[
    \mathbb{E}
    \left[
        W_2
        \left(
            \widehat{\pi}^{\theta_T}_1,
            \pi_1
        \right)
    \right]
    \le
    K_{\mathrm{flow}}
    \left(
        \epsilon_{\mathrm{approx}}
        +
        \frac{
            C_{\mathrm{opt,pop}}
        }{
            T
        }
    \right)^{1/2},
\]
where
\(
    C_{\mathrm{opt,pop}}
\)
is the optimization constant associated with the distinct population-oracle
assumptions. This comparison highlights that the
\(
    \widetilde O
    \left(
        \epsilon^{-4}
    \right)
\)
dataset-size scaling in the finite-dataset result arises from the
\(
    \widetilde O
    \left(
        N^{-1/2}
    \right)
\)
statistical contribution at the squared-loss level.
\end{remark}

\begin{remark}[Dependence on dimension and architecture]
\label{remark:dimension_architecture}

The theorem should be interpreted for a prescribed bounded-parameter smooth
neural velocity class. The ambient dimension enters through the parameter count
\(
    P
    =
    O(DW^2+dW),
\)
the smooth-network metric-entropy complexity
\(
    \mathcal C_{\mathrm{sm}}(R,N),
\)
the truncation envelope
\(
    B_\ell(R),
\)
the tail and truncation constants, the flow-stability factor
\(
    K_{\mathrm{flow}},
\)
the approximation error
\(
    \epsilon_{\mathrm{approx}},
\)
and the uniform empirical optimization quantity
\(
    \overline C_{\mathrm{opt}}.
\)
\end{remark}

%% file: Sections/conclusion.tex
\section{Conclusion}

We developed a finite-sample and finite-time analysis of flow matching when the velocity field is learned by
stochastic-gradient training on a fixed empirical objective. Rather than assuming that the learning algorithm
can access an exact empirical risk minimizer, we analyze the parameter obtained after a finite number of
stochastic-gradient updates on a finite stored dataset.

At the population-loss level, the analysis uses a three-term algebraic decomposition. The excess
flow-matching loss is bounded by an approximation contribution associated with the prescribed
bounded-parameter smooth neural velocity class, a finite-dataset statistical contribution arising from
replacing the population objective with its empirical counterpart, and an empirical optimization contribution
arising from approximate minimization of the fixed empirical objective. The empirical minimizer appears only
as a theoretical comparator and is not assumed to be available to the learning algorithm.

Because the source and target distributions are allowed to be unbounded, the statistical contribution is
controlled through endpoint-magnitude truncation. More specifically, its bound consists of a truncated
metric-entropy or Rademacher uniform-convergence term together with an explicit endpoint-tail remainder under
the sub exponential endpoint assumption. Consequently, the final theorem displays approximation,
finite-dataset statistical, endpoint tail, and empirical optimization contributions separately, while the tail
term arises as part of the control of the statistical component.

Combining these ingredients with the path-consistent uniform flow-stability condition yields a
Wasserstein-2 guarantee for the terminal law of the exact learned ODE flow. The guarantee holds with high
probability over the sampled finite dataset and in conditional expectation over the stochastic-gradient
randomness. It makes explicit that the dataset size
$ N$
controls the finite-dataset statistical contribution, whereas the optimization budget
$T$ controls the finite-time empirical optimization contribution.

For a prescribed bounded-parameter smooth neural velocity class, and after displaying  the
approximation, tail, flow stability, smooth-network-complexity, and empirical-optimization quantities, the
result yields the shorthand requirements
\[
    N
    =
    \widetilde O
    \left(
        \epsilon^{-4}
    \right),
    \qquad
    T
    =
    \widetilde O
    \left(
        \epsilon^{-2}
    \right)
\]
for obtaining Wasserstein-2 error of order
$O(\epsilon).$ These rates should be interpreted only under the stated assumptions. In particular, the ambient dimension and
network architecture enter through the parameterization of the neural class, the metric-entropy quantity, the
truncation envelope, the approximation error, the flow-stability factor, and the empirical optimization
constant. We do not claim dimension-free approximation or dimension-free flow-stability guarantees.

Several directions remain open. First, deriving explicit approximation results for marginal velocity fields
under structural assumptions on the target distribution would clarify the dependence on dimension,
regularity, and network size. Second, replacing the empirical proximal Polyak-\L{}ojasiewicz condition, the interior operating regime, and the uniform-in-\(N\) empirical-constant
regularity condition with weaker or more directly verifiable optimization
requirements would broaden the scope of the result. Third, additional martingale-tail assumptions could support a fully
high-probability optimization analysis jointly over the dataset and stochastic-gradient randomness. Fourth,
incorporating numerical ODE solver error into the main theorem would yield an end-to-end guarantee for
practical sampling. Finally, extending the analysis beyond linear or rectified interpolation paths to more
general conditional flow-matching, optimal-transport, and stochastic-interpolant constructions remains an
important direction for future work.

%% file: Sections/lemma_2.tex
\section{Proof of Lemma~\ref{lemma:stat_error}}
\label{app:proof_stat_error}

\begin{proof}
Let
\[
    Z
    =
    (X_0,X_1,t),
    \qquad
    (X_0,X_1)
    \sim
    \Gamma,
    \qquad
    t
    \sim
    \operatorname{Uniform}[0,1].
\]
For the linear interpolation path,
\[
    X_t
    =
    (1-t)X_0+tX_1,
    \qquad
    \dot X_t
    =
    X_1-X_0.
\]
For every
\(
    \theta
    \in
    \Theta,
\)
define the single-sample flow-matching loss
\[
    g_\theta(Z)
    :=
    \ell_\theta(Z)
    =
    \left\|
        u_\theta
        \left(
            (1-t)X_0+tX_1,
            t
        \right)
        -
        (X_1-X_0)
    \right\|_2^2.
\]
Thus,
\[
    L(\theta)
    =
    \mathbb{E}
    \left[
        g_\theta(Z)
    \right],
    \qquad
    \widehat L_N(\theta)
    =
    \frac{1}{N}
    \sum_{i=1}^N
    g_\theta(Z_i),
\]
where
\[
    \mathcal{D}_N
    =
    \{Z_i\}_{i=1}^N
\]
consists of independent copies of \(Z\). We seek a high-probability bound on
\[
    2
    \sup_{\theta\in\Theta}
    \left|
        L(\theta)-\widehat L_N(\theta)
    \right|.
\]

The source and target distributions are not assumed to be compactly supported, so the loss need not be
uniformly bounded on the full sample space. We therefore use endpoint truncation.

For
\(
    R
    >
    0,
\)
define the endpoint-truncation event
\[
    \mathcal{E}_R
    :=
    \left\{
        \|X_0\|_2
        \le
        R,
        \,
        \|X_1\|_2
        \le
        R
    \right\}.
\]
For the \(i\)-th sample, write
\[
    \mathcal{E}_{R,i}
    :=
    \left\{
        \|X_{0,i}\|_2
        \le
        R,
        \,
        \|X_{1,i}\|_2
        \le
        R
    \right\}.
\]
On
\(
    \mathcal{E}_R,
\)
the linear path satisfies
\[
\begin{aligned}
    \|X_t\|_2
    &=
    \|(1-t)X_0+tX_1\|_2
    \\
    &\le
    (1-t)\|X_0\|_2+t\|X_1\|_2
    \\
    &\le
    R,
\end{aligned}
\]
and
\[
    \|X_1-X_0\|_2
    \le
    \|X_1\|_2+\|X_0\|_2
    \le
    2R.
\]

By the uniform linear-growth property of the bounded-parameter smooth neural velocity class from
Lemma~\ref{lemma:nn_linear_growth}, there exist constants
\[
    A_{\mathrm{NN}},
    B_{\mathrm{NN}}
    >
    0
\]
such that
\[
    \|u_\theta(x,t)\|_2
    \le
    A_{\mathrm{NN}}
    +
    B_{\mathrm{NN}}
    \|x\|_2
\]
for every
\[
    \theta
    \in
    \Theta,
    \qquad
    x
    \in
    \mathbb{R}^d,
    \qquad
    t
    \in
    [0,1].
\]
Hence, on $\mathcal{E}_R,$
\[
    \|u_\theta(X_t,t)\|_2
    \le
    A_{\mathrm{NN}}
    +
    B_{\mathrm{NN}}R.
\]
Therefore,
\[
\begin{aligned}
    g_\theta(Z)
    &=
    \left\|
        u_\theta(X_t,t)-(X_1-X_0)
    \right\|_2^2 \le
    \left(
        A_{\mathrm{NN}}
        +
        (B_{\mathrm{NN}}+2)R
    \right)^2.
\end{aligned}
\]
Define the truncated-loss envelope
\[
    B_\ell(R)
    :=
    \left(
        A_{\mathrm{NN}}
        +
        (B_{\mathrm{NN}}+2)R
    \right)^2.
\]

Define the truncated loss
\[
    g_{\theta,R}(Z)
    :=
    g_\theta(Z)
    \mathbf{1}_{\mathcal{E}_R},
\]
and the truncated squared-loss class
\[
    \mathcal{G}_{W,D}^{(R)}
    :=
    \left\{
        Z
        \mapsto
        g_{\theta,R}(Z)
        :
        \theta
        \in
        \Theta
    \right\}.
\]
Every function in
$ \mathcal{G}_{W,D}^{(R)}$
takes values in
$ [0,B_\ell(R)].$
Recall that $ C_{\mathrm{sm}}(R)$ is the raw metric-entropy radius of this class. Let us define define
\[
    \mathcal C_{\mathrm{sm}}(R,N)
    :=
    P
    \log
    \left(
        \overline C_{\mathrm{sm}}(R,N)
    \right).
\]

By the metric-entropy/Rademacher uniform-convergence result in
Lemma~\ref{lemma:pdim_uniform_convergence}, there exists a constant
\(
    C_{\mathrm{uc}}
    >
    0
\)
such that, for every
\[
    \delta_{\mathrm{uc}}
    \in
    (0,1),
\]
with probability at least
\[
    1-\delta_{\mathrm{uc}},
\]
\[
\begin{aligned}
    \sup_{\theta\in\Theta}
    \left|
        \mathbb{E}
        \left[
            g_{\theta,R}(Z)
        \right]
        -
        \frac{1}{N}
        \sum_{i=1}^N
        g_{\theta,R}(Z_i)
    \right|
    &\le
    C_{\mathrm{uc}}
    B_\ell(R)
    \sqrt{
        \frac{
            \mathcal C_{\mathrm{sm}}(R,N)
            +
            \log(1/\delta_{\mathrm{uc}})
        }{
            N
        }
    }.
\end{aligned}
\]

Set $\delta_{\mathrm{uc}} := \frac{3\delta}{4}.$
Since
\[
    \log
    \left(
        \frac{1}{\delta_{\mathrm{uc}}}
    \right)
    =
    \log
    \left(
        \frac{4}{3\delta}
    \right)
    \le
    \log
    \left(
        \frac{4}{\delta}
    \right),
\]
it follows that, with probability at least
$1-\frac{3\delta}{4},$
\[
\begin{aligned}
    \sup_{\theta\in\Theta}
    \left|
        \mathbb{E}
        \left[
            g_{\theta,R}(Z)
        \right]
        -
        \frac{1}{N}
        \sum_{i=1}^N
        g_{\theta,R}(Z_i)
    \right|
    &\le
    C_{\mathrm{uc}}
    B_\ell(R)
    \sqrt{
        \frac{
            \mathcal C_{\mathrm{sm}}(R,N)
            +
            \log(4/\delta)
        }{
            N
        }
    }.
\end{aligned}
\]

It remains to control the contribution outside the endpoint-truncation event. Define
\[
    \epsilon_{\mathrm{tail}}(R)
    :=
    \sup_{\theta\in\Theta}
    \mathbb{E}
    \left[
        g_\theta(Z)
        \mathbf{1}_{\mathcal{E}_R^c}
    \right].
\]
By the linear-growth property,
\[
    \|u_\theta(X_t,t)\|_2
    \le
    A_{\mathrm{NN}}
    +
    B_{\mathrm{NN}}
    \|X_t\|_2.
\]
Since
\[
    X_t
    =
    (1-t)X_0+tX_1,
\]
we have
\[
    \|X_t\|_2
    \le
    \|X_0\|_2+\|X_1\|_2,
\]
and
\[
    \|X_1-X_0\|_2
    \le
    \|X_0\|_2+\|X_1\|_2.
\]
Thus, after increasing a constant if necessary, there exists
\(
    C_{\mathrm{env}}
    >
    0
\)
such that
\[
    g_\theta(Z)
    \le
    C_{\mathrm{env}}
    \left(
        1
        +
        \|X_0\|_2^2
        +
        \|X_1\|_2^2
    \right)
\]
uniformly over
\(
    \theta
    \in
    \Theta.
\)
Moreover,
\[
    \mathcal{E}_R^c
    \subseteq
    \left\{
        \|X_0\|_2
        >
        R
    \right\}
    \cup
    \left\{
        \|X_1\|_2
        >
        R
    \right\}.
\]
Using Assumption~\ref{assump:subexp} and the tail-integration identity for nonnegative random variables,
there exist constants
\(
    C_{\mathrm{tail}}',
    c_{\mathrm{tail}}
    >
    0
\)
such that
\[
    \epsilon_{\mathrm{tail}}(R)
    \le
    C_{\mathrm{tail}}'
    (1+R)^2
    \exp
    \left(
        -c_{\mathrm{tail}}R/K_X
    \right).
\]

We next connect the truncated empirical loss to the original empirical loss. Define the all-sample
truncation event
\[
    \mathcal{A}_R
    :=
    \bigcap_{i=1}^N
    \mathcal{E}_{R,i}.
\]
On
\[
    \mathcal{A}_R,
\]
\[
    g_{\theta,R}(Z_i)
    =
    g_\theta(Z_i),
    \qquad
    i
    =
    1,\ldots,N,
    \qquad
    \forall
    \theta
    \in
    \Theta.
\]
Consequently, on $\mathcal{A}_R,$
\[
    \widehat L_N(\theta)
    =
    \frac{1}{N}
    \sum_{i=1}^N
    g_{\theta,R}(Z_i),
    \qquad
    \forall
    \theta
    \in
    \Theta.
\]

By Assumption~\ref{assump:subexp} and the union bound,
\[
\begin{aligned}
    \mathbb{P}
    \left(
        \mathcal{A}_R^c
    \right)
    &\le
    \sum_{i=1}^N
    \mathbb{P}
    \left(
        \mathcal{E}_{R,i}^c
    \right)
    \\
    &\le
    N
    \left[
        \mathbb{P}
        \left(
            \|X_0\|_2>R
        \right)
        +
        \mathbb{P}
        \left(
            \|X_1\|_2>R
        \right)
    \right]
    \\
    &\le
    2Nc_0
    \exp
    \left(
        -c_1R/K_X
    \right).
\end{aligned}
\]
Therefore, if
\[
    R
    \ge
    \frac{K_X}{c_1}
    \log
    \left(
        \frac{8Nc_0}{\delta}
    \right),
\]
then
\[
    \mathbb{P}
    \left(
        \mathcal{A}_R^c
    \right)
    \le
    \frac{\delta}{4}.
\]

On the intersection of the truncated uniform-convergence event and
\(
    \mathcal{A}_R,
\)
for every
\(
    \theta
    \in
    \Theta,
\)
\[
\begin{aligned}
    \left|
        L(\theta)-\widehat L_N(\theta)
    \right|
    &=
    \left|
        \mathbb{E}
        \left[
            g_\theta(Z)
        \right]
        -
        \frac{1}{N}
        \sum_{i=1}^N
        g_\theta(Z_i)
    \right|
    \\
    &\le
    \left|
        \mathbb{E}
        \left[
            g_{\theta,R}(Z)
        \right]
        -
        \frac{1}{N}
        \sum_{i=1}^N
        g_{\theta,R}(Z_i)
    \right|
    \\
    &\quad
    +
    \mathbb{E}
    \left[
        g_\theta(Z)
        \mathbf{1}_{\mathcal{E}_R^c}
    \right].
\end{aligned}
\]
Taking the supremum over
\(
    \theta
    \in
    \Theta,
\)
we obtain
\[
\begin{aligned}
    \sup_{\theta\in\Theta}
    \left|
        L(\theta)-\widehat L_N(\theta)
    \right|
    &\le
    C_{\mathrm{uc}}
    B_\ell(R)
    \sqrt{
        \frac{
            \mathcal C_{\mathrm{sm}}(R,N)
            +
            \log(4/\delta)
        }{
            N
        }
    }
    \\
    &\quad
    +
    C_{\mathrm{tail}}'
    (1+R)^2
    \exp
    \left(
        -c_{\mathrm{tail}}R/K_X
    \right).
\end{aligned}
\]
Multiplying by $2$
and absorbing the factor into the constants, there exist constants
\[
    C_{\mathrm{stat}},
    C_{\mathrm{tail}}
    >
    0
\]
such that
\[
\begin{aligned}
    2
    \sup_{\theta\in\Theta}
    \left|
        L(\theta)-\widehat L_N(\theta)
    \right|
    &\le
    C_{\mathrm{stat}}
    B_\ell(R)
    \sqrt{
        \frac{
            \mathcal C_{\mathrm{sm}}(R,N)
            +
            \log(4/\delta)
        }{
            N
        }
    }
    \\
    &\quad
    +
    C_{\mathrm{tail}}
    (1+R)^2
    \exp
    \left(
        -c_{\mathrm{tail}}R/K_X
    \right).
\end{aligned}
\]

The truncated uniform-convergence event fails in the set  with probability at most
$\frac{3\delta}{4},$
and $\mathcal{A}_R$
fails with probability at most $\frac{\delta}{4}.$
Hence, by the union bound, the preceding display holds with probability at least $1-\delta.$
This proves the first claim.

Finally, choosing
\[
    R
    \asymp
    K_X
    \log
    \left(
        N/\delta
    \right)
\]
satisfies the all-sample truncation requirement and makes
\[
    C_{\mathrm{tail}}
    (1+R)^2
    \exp
    \left(
        -c_{\mathrm{tail}}R/K_X
    \right)
\]
lower order up to logarithmic factors. Hence,
\[
\begin{aligned}
    2
    \sup_{\theta\in\Theta}
    \left|
        L(\theta)-\widehat L_N(\theta)
    \right|
    =
    \widetilde O
    \left(
        B_\ell(R)
        \sqrt{
            \frac{
                \mathcal C_{\mathrm{sm}}(R,N)
                +
                \log(1/\delta)
            }{
                N
            }
        }
    \right),
\end{aligned}
\]
where the suppressed factors include the truncation-dependent envelope, the smooth-network entropy
factors, and logarithmic terms. This completes the proof.
\end{proof}

%% file: Sections/lemma_3.tex
\section{Proof of Lemma~\ref{lemma:opt_error}}
\label{app:proof_opt_error}

\begin{proof}

Let
\[
    \widehat\theta_N
    \in
    \arg\min_{\theta\in\Theta}
    \widehat L_N(\theta),
    \qquad
    \widehat L_N^\star
    :=
    \widehat L_N(\widehat\theta_N).
\]
We analyze projected stochastic-gradient descent on the fixed finite empirical
objective
\[
    \widehat L_N(\theta)
    =
    \frac{1}{N}
    \sum_{i=1}^N
    \ell_\theta(Z_i),
\]
where
\[
    \mathcal{D}_N
    =
    \{Z_i\}_{i=1}^N
\]
is held fixed throughout the optimization analysis. The empirical minimizer
\(
    \widehat\theta_N
\)
is used only as a theoretical comparator; the algorithm is not assumed to have
access to it.

Conditioned on
\(
    \mathcal{D}_N,
\)
the projected stochastic-gradient recursion is
\[
    \theta_{k+1}
    =
    \Pi_\Theta
    \left(
        \theta_k
        -
        \eta_k g_k
    \right),
    \qquad
    k
    =
    0,1,\ldots,T-1,
\]
where
\[
    g_k
    =
    \nabla_\theta
    \ell_{\theta_k}(Z_{I_k}),
    \qquad
    I_k
    \sim
    \operatorname{Uniform}
    \left\{
        1,\ldots,N
    \right\}.
\]
By Lemma~\ref{lemma:automatic_containment},
\[
    \theta_k
    \in
    \Theta,
    \qquad
    k
    =
    0,1,\ldots,T.
\]
Moreover, by the interior-operating-regime clause in
Assumption~\ref{assump:empirical_sgd}, the projection is inactive along the
analyzed stochastic trajectory. Thus, almost surely conditional on
\(
    \mathcal{D}_N,
\)
\[
    \theta_{k+1}
    =
    \theta_k
    -
    \eta_k g_k.
\]
Since
\(
    \theta_k,
    \theta_{k+1}
    \in
    \Theta
\)
and \(\Theta\) is convex, the line segment joining
\(
    \theta_k
    \qquad
    \text{and}
    \qquad
    \theta_{k+1}
\)
lies in \(\Theta\). Hence the \(\alpha_N\)-smoothness inequality for
\(
    \widehat L_N
\)
applies along this segment.

Let
\[
    \{\mathcal{F}_k\}_{k\geq 0}
\]
denote the filtration generated by the optimization process up to iteration
\(k\), conditionally on
\(
    \mathcal{D}_N.
\)
By Assumption~\ref{assump:empirical_sgd},
\[
    \mathbb{E}
    \left[
        g_k
        \,\middle|\,
        \mathcal{F}_k,\mathcal{D}_N
    \right]
    =
    \nabla\widehat L_N(\theta_k),
\]
and
\[
    \mathbb{E}
    \left[
        \left\|
            g_k
            -
            \nabla\widehat L_N(\theta_k)
        \right\|_2^2
        \,\middle|\,
        \mathcal{F}_k,\mathcal{D}_N
    \right]
    \le
    \sigma_N^2.
\]
Thus, the stochastic gradient is conditionally unbiased for the empirical
gradient, rather than for the population gradient.

For convenience, define the random empirical suboptimality
\[
    \Delta_k
    :=
    \widehat L_N(\theta_k)
    -
    \widehat L_N^\star.
\]
Because
\(
    \theta_k
    \in
    \Theta
\)
and
\[
    \widehat\theta_N
    \in
    \arg\min_{\theta\in\Theta}
    \widehat L_N(\theta),
\]
we have
\(
    \Delta_k
    \ge
    0.
\)

We first record the consequence of the empirical proximal-PL condition needed
below. For every
\(
    \theta
    \in
    \Theta,
\)
the definition of the constrained gradient-mapping quantity gives
\[
\begin{aligned}
    \mathcal{M}_N(\theta)
    &=
    -2\alpha_N
    \min_{\theta'\in\Theta}
    \left[
        \left\langle
            \nabla\widehat L_N(\theta),
            \theta'-\theta
        \right\rangle
        +
        \frac{\alpha_N}{2}
        \|\theta'-\theta\|_2^2
    \right]
    \\
    &\le
    -2\alpha_N
    \min_{\theta'\in\mathbb{R}^P}
    \left[
        \left\langle
            \nabla\widehat L_N(\theta),
            \theta'-\theta
        \right\rangle
        +
        \frac{\alpha_N}{2}
        \|\theta'-\theta\|_2^2
    \right]
    \\
    &=
    \left\|
        \nabla\widehat L_N(\theta)
    \right\|_2^2.
\end{aligned}
\]
Therefore, Assumption~\ref{assump:empirical_pl} implies
\[
    \frac{1}{2}
    \left\|
        \nabla\widehat L_N(\theta)
    \right\|_2^2
    \ge
    \frac{1}{2}
    \mathcal{M}_N(\theta)
    \ge
    \mu_N
    \left(
        \widehat L_N(\theta)
        -
        \widehat L_N^\star
    \right).
\]
In particular,
\[
    \frac{1}{2}
    \left\|
        \nabla\widehat L_N(\theta_k)
    \right\|_2^2
    \ge
    \mu_N
    \Delta_k.
\]

By \(\alpha_N\)-smoothness of
\[
    \widehat L_N,
\]
we have
\[
\begin{aligned}
    \widehat L_N(\theta_{k+1})
    &\le
    \widehat L_N(\theta_k)
    +
    \left\langle
        \nabla\widehat L_N(\theta_k),
        \theta_{k+1}-\theta_k
    \right\rangle
    +
    \frac{\alpha_N}{2}
    \left\|
        \theta_{k+1}-\theta_k
    \right\|_2^2
    \\
    &=
    \widehat L_N(\theta_k)
    -
    \eta_k
    \left\langle
        \nabla\widehat L_N(\theta_k),
        g_k
    \right\rangle
    +
    \frac{\alpha_N\eta_k^2}{2}
    \|g_k\|_2^2.
\end{aligned}
\]
Taking conditional expectation given
\[
    \mathcal{F}_k
    \qquad
    \text{and}
    \qquad
    \mathcal{D}_N,
\]
conditional unbiasedness gives
\[
    \mathbb{E}
    \left[
        \left\langle
            \nabla\widehat L_N(\theta_k),
            g_k
        \right\rangle
        \,\middle|\,
        \mathcal{F}_k,\mathcal{D}_N
    \right]
    =
    \left\|
        \nabla\widehat L_N(\theta_k)
    \right\|_2^2.
\]
Moreover,
\[
\begin{aligned}
    \mathbb{E}
    \left[
        \|g_k\|_2^2
        \,\middle|\,
        \mathcal{F}_k,\mathcal{D}_N
    \right]
    &=
    \left\|
        \nabla\widehat L_N(\theta_k)
    \right\|_2^2
    \\
    &\quad
    +
    \mathbb{E}
    \left[
        \left\|
            g_k
            -
            \nabla\widehat L_N(\theta_k)
        \right\|_2^2
        \,\middle|\,
        \mathcal{F}_k,\mathcal{D}_N
    \right]
    \\
    &\le
    \left\|
        \nabla\widehat L_N(\theta_k)
    \right\|_2^2
    +
    \sigma_N^2.
\end{aligned}
\]
Consequently,
\[
\begin{aligned}
    \mathbb{E}
    \left[
        \Delta_{k+1}
        \,\middle|\,
        \mathcal{F}_k,\mathcal{D}_N
    \right]
    &\le
    \Delta_k
    -
    \eta_k
    \left\|
        \nabla\widehat L_N(\theta_k)
    \right\|_2^2
    \\
    &\quad
    +
    \frac{\alpha_N\eta_k^2}{2}
    \left(
        \left\|
            \nabla\widehat L_N(\theta_k)
        \right\|_2^2
        +
        \sigma_N^2
    \right).
\end{aligned}
\]
Since
\[
    \eta_k
    \le
    \frac{1}{\alpha_N},
\]
we have
\[
    -\eta_k
    +
    \frac{\alpha_N\eta_k^2}{2}
    \le
    -\frac{\eta_k}{2}.
\]
Therefore,
\[
\begin{aligned}
    \mathbb{E}
    \left[
        \Delta_{k+1}
        \,\middle|\,
        \mathcal{F}_k,\mathcal{D}_N
    \right]
    &\le
    \Delta_k
    -
    \frac{\eta_k}{2}
    \left\|
        \nabla\widehat L_N(\theta_k)
    \right\|_2^2
    +
    \frac{\alpha_N\sigma_N^2}{2}
    \eta_k^2
    \\
    &\le
    \left(
        1-\mu_N\eta_k
    \right)
    \Delta_k
    +
    \frac{\alpha_N\sigma_N^2}{2}
    \eta_k^2.
\end{aligned}
\]

Define
\[
    e_k
    :=
    \mathbb{E}_{\mathrm{SGD}}
    \left[
        \Delta_k
        \,\middle|\,
        \mathcal{D}_N
    \right].
\]
Taking expectation over the optimization randomness conditional on
\(
    \mathcal{D}_N,
\)
we obtain
\[
    e_{k+1}
    \le
    \left(
        1-\mu_N\eta_k
    \right)
    e_k
    +
    \frac{\alpha_N\sigma_N^2}{2}
    \eta_k^2.
\]

We first establish the bound for arbitrary positive parameters
\(
    \alpha_0,
    \qquad
    \gamma
\)
satisfying
\[
    \alpha_0\mu_N
    >
    1,
    \qquad
    \gamma
    \ge
    \alpha_0\mu_N,
    \qquad
    \eta_k
    =
    \frac{\alpha_0}{k+\gamma}
    \le
    \frac{1}{\alpha_N}.
\]
Substitute
\[
    \eta_k
    =
    \frac{\alpha_0}{k+\gamma},
    \qquad
    p
    :=
    \alpha_0\mu_N,
    \qquad
    b
    :=
    \frac{
        \alpha_N
        \sigma_N^2
        \alpha_0^2
    }{
        2
    }.
\]
Then
\[
    e_{k+1}
    \le
    \left(
        1-\frac{p}{k+\gamma}
    \right)
    e_k
    +
    \frac{b}{(k+\gamma)^2}.
\]
The step-size conditions imply
\[
    p
    >
    1,
    \qquad
    \gamma
    \ge
    p.
\]
Hence, for every
\[
    k
    \ge
    0,
\]
\[
    0
    \le
    1-\frac{p}{k+\gamma}
    \le
    1.
\]

Define
\[
    e_0
    :=
    \widehat L_N(\theta_0)
    -
    \widehat L_N(\widehat\theta_N),
\]
and
\[
    C_0
    :=
    \max
    \left\{
        \gamma e_0,
        \frac{b}{p-1}
    \right\}.
\]
We claim that
\[
    e_k
    \le
    \frac{C_0}{k+\gamma}
    \qquad
    \text{for every }
    k
    \ge
    0.
\]
For
\[
    k
    =
    0,
\]
the claim follows from
\[
    C_0
    \ge
    \gamma e_0.
\]

Suppose the claim holds for some
\[
    k
    \ge
    0,
\]
and write
\[
    s
    :=
    k+\gamma.
\]
Since
\[
    1-\frac{p}{s}
    \ge
    0,
\]
the recursion and the induction hypothesis imply
\[
\begin{aligned}
    e_{k+1}
    &\le
    \left(
        1-\frac{p}{s}
    \right)
    \frac{C_0}{s}
    +
    \frac{b}{s^2}
    \\
    &=
    \frac{C_0}{s}
    -
    \frac{pC_0-b}{s^2}.
\end{aligned}
\]
Because
\[
    C_0
    \ge
    \frac{b}{p-1},
\]
we have
\[
    C_0(p-1)
    \ge
    b,
\]
and hence
\[
    pC_0-b
    \ge
    C_0.
\]
It follows that
\[
    e_{k+1}
    \le
    \frac{C_0}{s}
    -
    \frac{C_0}{s^2}
    =
    \frac{C_0(s-1)}{s^2}.
\]
Since
\[
    s
    \ge
    \gamma
    \ge
    p
    >
    1,
\]
we have
\[
    \frac{s-1}{s^2}
    \le
    \frac{1}{s+1}.
\]
Thus,
\[
    e_{k+1}
    \le
    \frac{C_0}{s+1}
    =
    \frac{C_0}{k+1+\gamma}.
\]
This completes the induction.

Consequently,
\[
    e_T
    \le
    \frac{C_0}{T+\gamma}
    \le
    \frac{C_0}{T}.
\]
Therefore,
\[
    \mathbb{E}_{\mathrm{SGD}}
    \left[
        \widehat L_N(\theta_T)
        -
        \widehat L_N(\widehat\theta_N)
        \,\middle|\,
        \mathcal{D}_N
    \right]
    \le
    \frac{
        C_{\mathrm{opt}}(\mathcal{D}_N)
    }{
        T
    },
\]
where we may take
\[
    C_{\mathrm{opt}}(\mathcal{D}_N)
    :=
    \max
    \left\{
        \gamma e_0,
        \frac{
            \alpha_N
            \sigma_N^2
            \alpha_0^2
        }{
            2
            \left(
                \alpha_0\mu_N-1
            \right)
        }
    \right\}.
\]

It remains to derive the deterministic uniform bound used in the main theorem.
On the event
\(
    \mathcal{G}_N
\)

from Assumption~\ref{assump:uniform_empirical_constants}, choose
\[
    \alpha_0
    :=
    \frac{2}{\mu_{\min}},
    \qquad
    \gamma
    :=
    \alpha_0\alpha_{\max},
    \qquad
    \eta_k
    :=
    \frac{\alpha_0}{k+\gamma}.
\]
Then
\[
    \alpha_0\mu_N
    \ge
    \alpha_0\mu_{\min}
    =
    2
    >
    1,
\]
\[
    \gamma
    =
    \alpha_0\alpha_{\max}
    \ge
    \alpha_0\mu_N,
\]
and
\[
    \eta_k
    \le
    \frac{\alpha_0}{\gamma}
    =
    \frac{1}{\alpha_{\max}}
    \le
    \frac{1}{\alpha_N}.
\]
Moreover,
\[
    \gamma e_0
    \le
    \gamma\overline e_0,
\]
and
\[
    \frac{
        \alpha_N
        \sigma_N^2
        \alpha_0^2
    }{
        2
        \left(
            \alpha_0\mu_N-1
        \right)
    }
    \le
    \frac{
        \alpha_{\max}
        \sigma_{\max}^2
        \alpha_0^2
    }{
        2
        \left(
            \alpha_0\mu_{\min}-1
        \right)
    }.
\]
Hence, on
\[
    \mathcal{G}_N,
\]
\[
    C_{\mathrm{opt}}(\mathcal{D}_N)
    \le
    \overline C_{\mathrm{opt}},
\]
where
\[
    \overline C_{\mathrm{opt}}
    :=
    \max
    \left\{
        \gamma\overline e_0,
        \frac{
            \alpha_{\max}
            \sigma_{\max}^2
            \alpha_0^2
        }{
            2
            \left(
                \alpha_0\mu_{\min}-1
            \right)
        }
    \right\}.
\]
Therefore, on
\[
    \mathcal{G}_N,
\]
\[
    \mathbb{E}_{\mathrm{SGD}}
    \left[
        \widehat L_N(\theta_T)
        -
        \widehat L_N(\widehat\theta_N)
        \,\middle|\,
        \mathcal{D}_N
    \right]
    \le
    \frac{
        \overline C_{\mathrm{opt}}
    }{
        T
    }.
\]

A high-probability analogue would require an additional explicit conditional tail
assumption on the stochastic-gradient noise and a separate martingale-concentration
analysis. Under the bounded conditional-variance assumption used here, the
conditional expectation guarantee above is the asserted conclusion.
\end{proof}

%% file: Sections/theoretical_result.tex
\section{Additional Details for Theorem~\ref{thm:main_flow_matching}}
\label{app:theorem_proof}

This section provides the detailed proof of
Theorem~\ref{thm:main_flow_matching}. The theorem and its supporting lemmas are
stated in the main text, so we do not restate them here. The proof combines the
projection identity, the finite-dataset statistical bound from
Lemma~\ref{lemma:stat_error}, the projected empirical optimization bound from
Lemma~\ref{lemma:opt_error}, and the deterministic uniform flow-stability bound
from Lemma~\ref{lemma:flow_stability}.

\begin{proof}[Proof of Theorem~\ref{thm:main_flow_matching}]
Let
\[
    \widehat\theta_N
    \in
    \arg\min_{\theta\in\Theta}
    \widehat L_N(\theta)
\]
be an empirical risk minimizer, and let
\[
    \theta^\star_\Theta
    \in
    \arg\min_{\theta\in\Theta}
    L(\theta)
\]
be a population minimizer within the prescribed neural velocity class. Both
parameters are theoretical comparators: the algorithm is only required to
produce the finite-time projected stochastic-gradient iterate
\(
    \theta_T.
\)
By Lemma~\ref{lemma:automatic_containment},
\(
    \theta_T
    \in
    \Theta.
\)

For every realization of the stochastic-gradient randomness, we decompose the
excess population loss as
\[
\begin{aligned}
    L(\theta_T)
    -
    L^\star
    &=
    L(\theta_T)
    -
    \widehat L_N(\theta_T)
    +
    \widehat L_N(\theta_T)
    -
    \widehat L_N(\widehat\theta_N)
    \\
    &\quad
    +
    \widehat L_N(\widehat\theta_N)
    -
    \widehat L_N(\theta^\star_\Theta)
    \\
    &\quad
    +
    \widehat L_N(\theta^\star_\Theta)
    -
    L(\theta^\star_\Theta)
    \\
    &\quad
    +
    L(\theta^\star_\Theta)
    -
    L^\star.
\end{aligned}
\]



Since
\[
    \widehat\theta_N
    \in
    \arg\min_{\theta\in\Theta}
    \widehat L_N(\theta),
\]
we have
\[
    \widehat L_N(\widehat\theta_N)
    -
    \widehat L_N(\theta^\star_\Theta)
    \le
    0.
\]
Therefore,
\[
\begin{aligned}
    L(\theta_T)
    -
    L^\star
    &\le
    L(\theta^\star_\Theta)
    -
    L^\star
    \\
    &\quad
    +
    \left|
        L(\theta_T)
        -
        \widehat L_N(\theta_T)
    \right|
    \\
    &\quad
    +
    \left|
        \widehat L_N(\theta^\star_\Theta)
        -
        L(\theta^\star_\Theta)
    \right|
    \\
    &\quad
    +
    \widehat L_N(\theta_T)
    -
    \widehat L_N(\widehat\theta_N)
    \\
    &\le
    L(\theta^\star_\Theta)
    -
    L^\star
    \\
    &\quad
    +
    2
    \sup_{\theta\in\Theta}
    \left|
        L(\theta)
        -
        \widehat L_N(\theta)
    \right|
    \\
    &\quad
    +
    \widehat L_N(\theta_T)
    -
    \widehat L_N(\widehat\theta_N).
\end{aligned}
\]

Define
\[
    E_{\mathrm{approx}}
    :=
    L(\theta^\star_\Theta)
    -
    L^\star,
\]
\[
    E_{\mathrm{stat}}
    :=
    2
    \sup_{\theta\in\Theta}
    \left|
        L(\theta)
        -
        \widehat L_N(\theta)
    \right|,
\]
and
\[
    E_{\mathrm{opt}}
    :=
    \widehat L_N(\theta_T)
    -
    \widehat L_N(\widehat\theta_N).
\]
Then
\[
    L(\theta_T)
    -
    L^\star
    \le
    E_{\mathrm{approx}}
    +
    E_{\mathrm{stat}}
    +
    E_{\mathrm{opt}}.
\]
By Lemma~\ref{lemma:approx_error},
\[
    E_{\mathrm{approx}}
    \le
    \epsilon_{\mathrm{approx}}.
\]
Fix
\(
    \delta,
    \delta_{\mathrm{opt}}
    \in
    (0,1)
\)
and a truncation radius satisfying
\[
    R
    \ge
    \frac{K_X}{c_1}
    \log
    \left(
        \frac{8Nc_0}{\delta}
    \right).
\]
Let
\[
    \mathcal{H}_{N,R,\delta}
\]
denote the statistical event from Lemma~\ref{lemma:stat_error}. On this event,
\[
\begin{aligned}
    E_{\mathrm{stat}}
    &\le
    C_{\mathrm{stat}}
    B_\ell(R)
    \sqrt{
        \frac{
            \mathcal C_{\mathrm{sm}}(R,N)
            +
            \log(4/\delta)
        }{
            N
        }
    }
    \\
    &\quad
    +
    C_{\mathrm{tail}}
    (1+R)^2
    \exp
    \left(
        -c_{\mathrm{tail}}R/K_X
    \right),
\end{aligned}
\]
and
\[
    \mathbb{P}
    \left(
        \mathcal{H}_{N,R,\delta}
    \right)
    \ge
    1-\delta.
\]

By Assumption~\ref{assump:uniform_empirical_constants}, there is an event
\(
    \mathcal{G}_N
\)
such that
\[
    \mathbb{P}
    \left(
        \mathcal{G}_N
    \right)
    \ge
    1-\delta_{\mathrm{opt}},
\]
and, on \(\mathcal{G}_N\), the uniform empirical-constant bounds hold. In
particular, Lemma~\ref{lemma:opt_error} gives
\[
    \mathbb{E}_{\mathrm{SGD}}
    \left[
        E_{\mathrm{opt}}
        \,\middle|\,
        \mathcal{D}_N
    \right]
    =
    \mathbb{E}_{\mathrm{SGD}}
    \left[
        \widehat L_N(\theta_T)
        -
        \widehat L_N(\widehat\theta_N)
        \,\middle|\,
        \mathcal{D}_N
    \right]
    \le
    \frac{
        \overline C_{\mathrm{opt}}
    }{
        T
    }.
\]

By the union bound,
\[
\begin{aligned}
    \mathbb{P}
    \left(
        \mathcal{H}_{N,R,\delta}
        \cap
        \mathcal{G}_N
    \right)
    &\ge
    1
    -
    \delta
    -
    \delta_{\mathrm{opt}}.
\end{aligned}
\]
Fix a dataset realization in
\[
    \mathcal{H}_{N,R,\delta}
    \cap
    \mathcal{G}_N.
\]
Taking conditional expectation over the stochastic-gradient randomness in the
error decomposition gives
\[
\begin{aligned}
    \mathbb{E}_{\mathrm{SGD}}
    \left[
        L(\theta_T)
        -
        L^\star
        \,\middle|\,
        \mathcal{D}_N
    \right]
    &\le
    \epsilon_{\mathrm{approx}}
    +
    C_{\mathrm{stat}}
    B_\ell(R)
    \sqrt{
        \frac{
            \mathcal C_{\mathrm{sm}}(R,N)
            +
            \log(4/\delta)
        }{
            N
        }
    }
    \\
    &\quad
    +
    C_{\mathrm{tail}}
    (1+R)^2
    \exp
    \left(
        -c_{\mathrm{tail}}R/K_X
    \right)
    +
    \frac{
        \overline C_{\mathrm{opt}}
    }{
        T
    }.
\end{aligned}
\]
Equivalently,
\[
    \mathbb{E}_{\mathrm{SGD}}
    \left[
        L(\theta_T)
        -
        L^\star
        \,\middle|\,
        \mathcal{D}_N
    \right]
    \le
    \mathfrak E_{N,T}
    \left(
        R,
        \delta
    \right).
\]

By the projection identity,
\[
    L(\theta_T)
    -
    L^\star
    =
    \mathbb{E}
    \left[
        \left\|
            u_{\theta_T}(X_t,t)
            -
            u^\star(X_t,t)
        \right\|_2^2
    \right].
\]
Hence, on
\[
    \mathcal{H}_{N,R,\delta}
    \cap
    \mathcal{G}_N,
\]
\[
\begin{aligned}
    \mathbb{E}_{\mathrm{SGD}}
    \left[
        \mathbb{E}
        \left[
            \left\|
                u_{\theta_T}(X_t,t)
                -
                u^\star(X_t,t)
            \right\|_2^2
        \right]
        \,\middle|\,
        \mathcal{D}_N
    \right]
    &\le
    \mathfrak E_{N,T}
    \left(
        R,
        \delta
    \right).
\end{aligned}
\]

It remains to convert velocity-field error into terminal distributional error.
By Assumption~\ref{assump:flow_stability}, the flow-stability constant
\[
    K_{\mathrm{flow}}
\]
is deterministic and uniform over
\(
    \Theta.
\)
In particular, Lemma~\ref{lemma:flow_stability} gives, for every
\(
    \theta
    \in
    \Theta,
\)
\[
    W_2^2
    \left(
        \widehat\pi^\theta_1,
        \pi_1
    \right)
    \le
    K_{\mathrm{flow}}^2
    \left(
        L(\theta)
        -
        L^\star
    \right).
\]
Applying this inequality to
\[
    \theta
    =
    \theta_T
    \in
    \Theta
\]
yields
\[
    W_2^2
    \left(
        \widehat\pi^{\theta_T}_1,
        \pi_1
    \right)
    \le
    K_{\mathrm{flow}}^2
    \left(
        L(\theta_T)
        -
        L^\star
    \right).
\]
Since
\(
    K_{\mathrm{flow}}
\)
is deterministic and independent of the stochastic-gradient randomness, taking
conditional expectation gives
\[
\begin{aligned}
    \mathbb{E}_{\mathrm{SGD}}
    \left[
        W_2^2
        \left(
            \widehat\pi^{\theta_T}_1,
            \pi_1
        \right)
        \,\middle|\,
        \mathcal{D}_N
    \right]
    &\le
    K_{\mathrm{flow}}^2
    \mathbb{E}_{\mathrm{SGD}}
    \left[
        L(\theta_T)
        -
        L^\star
        \,\middle|\,
        \mathcal{D}_N
    \right]
    \\
    &\le
    K_{\mathrm{flow}}^2
    \mathfrak E_{N,T}
    \left(
        R,
        \delta
    \right).
\end{aligned}
\]
Finally, Jensen's inequality gives
\[
\begin{aligned}
    \mathbb{E}_{\mathrm{SGD}}
    \left[
        W_2
        \left(
            \widehat\pi^{\theta_T}_1,
            \pi_1
        \right)
        \,\middle|\,
        \mathcal{D}_N
    \right]
    &\le
    \left(
        \mathbb{E}_{\mathrm{SGD}}
        \left[
            W_2^2
            \left(
                \widehat\pi^{\theta_T}_1,
                \pi_1
            \right)
            \,\middle|\,
            \mathcal{D}_N
        \right]
    \right)^{1/2}
    \\
    &\le
    K_{\mathrm{flow}}
    \mathfrak E_{N,T}
    \left(
        R,
        \delta
    \right)^{1/2}.
\end{aligned}
\]
Since
\[
    \mathbb{P}
    \left(
        \mathcal{H}_{N,R,\delta}
        \cap
        \mathcal{G}_N
    \right)
    \ge
    1
    -
    \delta
    -
    \delta_{\mathrm{opt}},
\]
this proves all conclusions of Theorem~\ref{thm:main_flow_matching}.
\end{proof}

\paragraph{Sample-complexity consequence.}

Choose
\[
    R
    =
    c_R
    K_X
    \log
    \left(
        N/\delta
    \right),
\]
where
\[
    c_R
    >
    0
\]
is sufficiently large that the truncation condition in
Theorem~\ref{thm:main_flow_matching} is satisfied. Then the tail term
\[
    C_{\mathrm{tail}}
    (1+R)^2
    \exp
    \left(
        -c_{\mathrm{tail}}R/K_X
    \right)
\]
is lower order up to logarithmic factors.

Suppose that the approximation and tail terms satisfy
\[
    \epsilon_{\mathrm{approx}}
    +
    C_{\mathrm{tail}}
    (1+R)^2
    \exp
    \left(
        -c_{\mathrm{tail}}R/K_X
    \right)
    \le
    \frac{
        \epsilon^2
    }{
        2K_{\mathrm{flow}}^2
    }.
\]
It is then sufficient to impose
\[
    C_{\mathrm{stat}}
    B_\ell(R)
    \sqrt{
        \frac{
            \mathcal C_{\mathrm{sm}}(R,N)
            +
            \log(4/\delta)
        }{
            N
        }
    }
    \le
    \frac{
        \epsilon^2
    }{
        4K_{\mathrm{flow}}^2
    },
\]
and
\[
    \frac{
        \overline C_{\mathrm{opt}}
    }{
        T
    }
    \le
    \frac{
        \epsilon^2
    }{
        4K_{\mathrm{flow}}^2
    }.
\]
In particular, there exists a sufficiently large constant
\(
    C
    >
    0
\)
such that it is sufficient to choose
\[
    N
    \ge
    C
    \frac{
        K_{\mathrm{flow}}^4
        B_\ell(R)^2
        \left(
            \mathcal C_{\mathrm{sm}}(R,N)
            +
            \log(1/\delta)
        \right)
    }{
        \epsilon^4
    },
\]
and
\[
    T
    \ge
    C
    \frac{
        K_{\mathrm{flow}}^2
        \overline C_{\mathrm{opt}}
    }{
        \epsilon^2
    }.
\]

The quantity
\(
    \overline C_{\mathrm{opt}}
\)
is deterministic once the constants in
Assumption~\ref{assump:uniform_empirical_constants} have been fixed. Thus, the
displayed optimization-budget requirement is no longer conditional on
uncontrolled dataset-specific constants. Rather, it holds on the event
\(
    \mathcal{G}_N,
\)
which is already included in the theorem's probability statement.

When the approximation, tail, flow-stability, smooth-network-complexity, and
uniform empirical-optimization quantities are treated as fixed or displayed
separately, the shorthand scaling is
\[
    N
    =
    \widetilde O
    \left(
        \epsilon^{-4}
    \right),
    \qquad
    T
    =
    \widetilde O
    \left(
        \epsilon^{-2}
    \right).
\]
The latter shorthand is conditional on the uniform empirical-constant
regularity in Assumption~\ref{assump:uniform_empirical_constants} and on the
interior operating regime in Assumption~\ref{assump:empirical_sgd}.

\paragraph{Pure population-oracle comparison.}

For comparison, suppose instead that the algorithm has access to a population
stochastic-gradient oracle and that the population objective satisfies
corresponding population proximal-PL, smoothness, and stochastic-gradient
conditions. The finite-dataset statistical term is then absent. Under these
distinct assumptions, the corresponding flow-stability argument gives
\[
    \mathbb{E}
    \left[
        W_2
        \left(
            \widehat\pi^{\theta_T}_1,
            \pi_1
        \right)
    \right]
    \le
    K_{\mathrm{flow}}
    \left(
        \epsilon_{\mathrm{approx}}
        +
        \frac{
            C_{\mathrm{opt,pop}}
        }{
            T
        }
    \right)^{1/2},
\]
where
\(
    C_{\mathrm{opt,pop}}
\) is the optimization constant associated with the distinct population-oracle
assumptions. This comparison highlights that the finite-dataset
\(
    \widetilde O
    \left(
        \epsilon^{-4}
    \right)
\)
dataset-size scaling arises from the
\(
    \widetilde O
    \left(
        N^{-1/2}
    \right)
\) statistical contribution at the squared-loss level.

%% file: Sections/supp_lemmas.tex
\section{Supplementary Lemmas}
\label{app:supplementary_lemmas}

In this appendix, we collect auxiliary facts used in the proof of the main theorem. These lemmas are stated
for the revised flow-matching formulation based on the linear, or rectified, path
\[
    X_t
    =
    (1-t)X_0+tX_1,
    \qquad
    (X_0,X_1)
    \sim
    \Gamma,
    \qquad
    X_0
    \sim
    \pi_0,
    \qquad
    X_1
    \sim
    \pi_1.
\]
They provide the pathwise bounds, projection identity, smooth-network properties, metric-entropy control,
tail bound, flow-stability argument, and numerical-discretization observation used throughout the paper.

\begin{lemma}[Boundedness of the linear interpolation path]
\label{lemma:linear_path_boundedness}

Suppose that the source and target variables satisfy
\[
    \|X_0\|_2
    \le
    R,
    \qquad
    \|X_1\|_2
    \le
    R.
\]
Then, for every
\[
    t
    \in
    [0,1],
\]
\[
    \|X_t\|_2
    \le
    R,
    \qquad
    \|\dot X_t\|_2
    =
    \|X_1-X_0\|_2
    \le
    2R.
\]
\end{lemma}

\begin{proof}
For the linear path,
\[
    X_t
    =
    (1-t)X_0+tX_1.
\]
By convexity of the Euclidean norm,
\[
\begin{aligned}
    \|X_t\|_2
    &\le
    (1-t)\|X_0\|_2+t\|X_1\|_2
    \\
    &\le
    (1-t)R+tR
    \\
    &=
    R.
\end{aligned}
\]
Moreover,
\[
    \dot X_t
    =
    X_1-X_0,
\]
so the triangle inequality gives
\[
\begin{aligned}
    \|\dot X_t\|_2
    &=
    \|X_1-X_0\|_2
    \\
    &\le
    \|X_1\|_2+\|X_0\|_2
    \\
    &\le
    2R.
\end{aligned}
\]
\end{proof}

\begin{lemma}[\(L^2\) projection identity for flow matching]
\label{lemma:l2_projection_identity}

Let
\[
    L(u)
    :=
    \mathbb{E}
    \left[
        \left\|
            u(X_t,t)-\dot X_t
        \right\|_2^2
    \right],
\]
where the expectation is over
\(
    (X_0,X_1)
    \sim
    \Gamma
\)
and
\(
    t
    \sim
    \operatorname{Uniform}[0,1].
\)
Define the marginal velocity field by
\[
    u^\star(x,t)
    :=
    \mathbb{E}
    \left[
        \dot X_t
        \,\middle|\,
        X_t=x,
        \,
        t
    \right].
\]
Then
$ u^\star$
minimizes
$ L(u)$
over all measurable deployable velocity fields
\[
    u:
    \mathbb{R}^d
    \times
    [0,1]
    \to
    \mathbb{R}^d.
\]
Moreover, for every candidate field
\[
    u,
\]
\[
    L(u)-L(u^\star)
    =
    \mathbb{E}
    \left[
        \left\|
            u(X_t,t)-u^\star(X_t,t)
        \right\|_2^2
    \right].
\]
In particular, for the neural velocity field
\[
    u_\theta,
\]
\[
    L(\theta)-L^\star
    =
    \mathbb{E}
    \left[
        \left\|
            u_\theta(X_t,t)-u^\star(X_t,t)
        \right\|_2^2
    \right].
\]
\end{lemma}

\begin{proof}
For every measurable deployable field
\[
    u,
\]
write
\[
    u(X_t,t)-\dot X_t
    =
    \left(
        u(X_t,t)-u^\star(X_t,t)
    \right)
    +
    \left(
        u^\star(X_t,t)-\dot X_t
    \right).
\]
Squaring both sides and taking expectation gives
\[
\begin{aligned}
    L(u)
    &=
    \mathbb{E}
    \left[
        \left\|
            u(X_t,t)-u^\star(X_t,t)
        \right\|_2^2
    \right]
    \\
    &\quad
    +
    \mathbb{E}
    \left[
        \left\|
            u^\star(X_t,t)-\dot X_t
        \right\|_2^2
    \right]
    \\
    &\quad
    +
    2
    \mathbb{E}
    \left[
        \left\langle
            u(X_t,t)-u^\star(X_t,t),
            u^\star(X_t,t)-\dot X_t
        \right\rangle
    \right].
\end{aligned}
\]
The cross term is zero. Indeed, conditioning on
\[
    X_t
    \qquad
    \text{and}
    \qquad
    t,
\]
we obtain
\[
\begin{aligned}
    &
    \mathbb{E}
    \left[
        \left\langle
            u(X_t,t)-u^\star(X_t,t),
            u^\star(X_t,t)-\dot X_t
        \right\rangle
        \,\middle|\,
        X_t,t
    \right]
    \\
    &\qquad
    =
    \left\langle
        u(X_t,t)-u^\star(X_t,t),
        \mathbb{E}
        \left[
            u^\star(X_t,t)-\dot X_t
            \,\middle|\,
            X_t,t
        \right]
    \right\rangle
    \\
    &\qquad
    =
    0.
\end{aligned}
\]
Therefore,
\[
    L(u)
    =
    \mathbb{E}
    \left[
        \left\|
            u(X_t,t)-u^\star(X_t,t)
        \right\|_2^2
    \right]
    +
    L(u^\star).
\]
This proves both the optimality of
$ u^\star$
and the excess-risk identity.
\end{proof}

\begin{lemma}[Uniform linear growth and parameter regularity of bounded smooth networks]
\label{lemma:nn_linear_growth}

Let
\[
    u_\theta:
    \mathbb{R}^d
    \times
    [0,1]
    \to
    \mathbb{R}^d
\]
be a fully connected feedforward network with depth
\(
    D,
\)
hidden width
\(
    W,
\)
and parameters
\(
    \theta
    \in
    \Theta.
\)
Suppose that
\(
    \Theta
    \subset
    \mathbb{R}^P
\)
is nonempty, compact, and convex. Suppose that the activation function
\(
    \sigma:
    \mathbb{R}
    \to
    \mathbb{R}
\)
satisfies
\(
    \sigma
    \in
    C^2(\mathbb{R}),
    \qquad
    \sup_{z\in\mathbb{R}}
    |\sigma'(z)|
    \le
    M_1,
    \qquad
    \sup_{z\in\mathbb{R}}
    |\sigma''(z)|
    \le
    M_2,
    \qquad
    |\sigma(0)|
    \le
    M_0.
\)
Then there exist finite constants
\[
    A_{\mathrm{NN}},
    B_{\mathrm{NN}}
    >
    0,
\]
depending only on the architecture, parameter constraint, and activation constants, such that
\[
    \|u_\theta(x,t)\|_2
    \le
    A_{\mathrm{NN}}
    +
    B_{\mathrm{NN}}
    \|x\|_2
\]
for every
\[
    \theta
    \in
    \Theta,
    \qquad
    x
    \in
    \mathbb{R}^d,
    \qquad
    t
    \in
    [0,1].
\]

Moreover, for every fixed finite dataset
\[
    \mathcal{D}_N
    =
    \{Z_i\}_{i=1}^N,
\]
the empirical flow-matching loss
\[
    \widehat L_N(\theta)
    =
    \frac{1}{N}
    \sum_{i=1}^N
    \ell_\theta(Z_i)
\]
is twice continuously differentiable in
\(
    \theta.
\)
There exists a finite constant
\(
    \alpha_N
    <
    \infty,
\)
depending on the fixed dataset, architecture, parameter constraint, and activation constants, such that
\[
    \left\|
        \nabla\widehat L_N(\theta)
        -
        \nabla\widehat L_N(\theta')
    \right\|_2
    \le
    \alpha_N
    \left\|
        \theta-\theta'
    \right\|_2
\]
for every
\(
    \theta,
    \theta'
    \in
    \Theta.
\)
Thus, for smooth activations, empirical smoothness is compatible with bounded-parameter finite-dataset
training on the compact convex region
\(
    \Theta.
\)
The projected recursion in Assumption~\ref{assump:empirical_sgd} ensures that all optimization iterates remain
in \(\Theta\) by Lemma~\ref{lemma:automatic_containment}. The empirical proximal
Polyak--\L{}ojasiewicz condition, the interior operating regime, and the
uniform-in-\(N\) empirical-constant condition remain separate optimization assumptions.
\end{lemma}

\begin{proof}
Let
\[
    h_0
    =
    (x,t)
    \in
    \mathbb{R}^{d+1}
\]
denote the network input. Since
\[
    t
    \in
    [0,1],
\]
\[
    \|h_0\|_2
    \le
    \|x\|_2+1.
\]
Write the hidden-layer recursion as
\[
    h_{\ell+1}
    =
    \sigma(A_\ell h_\ell+b_\ell),
    \qquad
    \ell
    =
    0,\ldots,D-1,
\]
where \(\sigma\) is applied coordinatewise. Since \(\sigma\)
has bounded first derivative,
\[
    |\sigma(z)|
    \le
    |\sigma(0)|+M_1|z|.
\]
Therefore, for a vector
\(
    v,
\)
\[
    \|\sigma(v)\|_2
    \le
    M_1\|v\|_2
    +
    M_0\sqrt{\dim(v)}.
\]
Because \(\Theta\)
is compact and the network has finitely many layers and parameters, there exist finite constants
\[
    M_A,
    M_b
    >
    0
\]
such that, for every layer and every
\(
    \theta
    \in
    \Theta,
\)
\[
    \|A_\ell\|_{\mathrm{op}}
    \le
    M_A,
    \qquad
    \|b_\ell\|_2
    \le
    M_b.
\]
Hence, after adjusting constants to absorb the finite layer widths, there exist
\[
    c_1,
    c_2
    >
    0
\]
such that
\[
    \|h_{\ell+1}\|_2
    \le
    c_1\|h_\ell\|_2+c_2.
\]
Iterating this recursion over the finitely many layers gives finite constants
\[
    a_D,
    b_D
    >
    0
\]
such that
\[
    \|h_D\|_2
    \le
    a_D\|h_0\|_2+b_D.
\]
The final affine output layer is controlled by the same compact-parameter argument. Thus, after enlarging
the constants if necessary,
\[
    \|u_\theta(x,t)\|_2
    \le
    a_D(\|x\|_2+1)+b_D
    =
    (a_D+b_D)+a_D\|x\|_2.
\]
The linear-growth conclusion follows by taking
\[
    A_{\mathrm{NN}}
    :=
    a_D+b_D,
    \qquad
    B_{\mathrm{NN}}
    :=
    a_D.
\]

For the parameter-regularity claim, fix
\[
    \mathcal{D}_N
    =
    \{Z_i\}_{i=1}^N.
\]
For every sample
\[
    Z_i
    =
    (X_{0,i},X_{1,i},t_i),
\]
the corresponding input
\[
    \left(
        (1-t_i)X_{0,i}+t_iX_{1,i},
        t_i
    \right)
\]
and target velocity
\[
    X_{1,i}-X_{0,i}
\]
are fixed finite vectors. Because the activation is
\(
    C^2,
\)
the map
\[
    \theta
    \mapsto
    u_\theta
    \left(
        (1-t_i)X_{0,i}+t_iX_{1,i},
        t_i
    \right)
\]
is twice continuously differentiable. Consequently,
\[
    \theta
    \mapsto
    \ell_\theta(Z_i)
\]
is twice continuously differentiable, and hence so is
\(
    \widehat L_N(\theta).
\)

Since
\(
    \Theta
\)
is compact and
\[
    \nabla^2\widehat L_N(\theta)
\]
is continuous in \(\theta\), the quantity
\[
    \alpha_N
    :=
    \sup_{\theta\in\Theta}
    \left\|
        \nabla^2\widehat L_N(\theta)
    \right\|_{\mathrm{op}}
\]
is finite. Since \(\Theta\) is convex, the line segment joining any two points
\[
    \theta,
    \theta'
    \in
    \Theta
\]
remains in \(\Theta\). Applying the fundamental theorem of calculus to the gradient along this segment yields
\[
\begin{aligned}
    \left\|
        \nabla\widehat L_N(\theta)
        -
        \nabla\widehat L_N(\theta')
    \right\|_2
    &\le
    \sup_{\xi\in\Theta}
    \left\|
        \nabla^2\widehat L_N(\xi)
    \right\|_{\mathrm{op}}
    \left\|
        \theta-\theta'
    \right\|_2
    \\
    &\le
    \alpha_N
    \left\|
        \theta-\theta'
    \right\|_2.
\end{aligned}
\]
This proves the empirical smoothness compatibility claim.
\end{proof}

\begin{lemma}[Boundedness of the truncated flow-matching loss]
\label{lemma:bounded_truncated_loss}

Suppose the endpoint-truncation event
\[
    \mathcal{E}_R
    :=
    \left\{
        \|X_0\|_2
        \le
        R,
        \,
        \|X_1\|_2
        \le
        R
    \right\}
\]
holds. Then, for every
\(
    \theta
    \in
    \Theta,
\)
\[
    \ell_\theta(Z)
    =
    \|u_\theta(X_t,t)-(X_1-X_0)\|_2^2
    \le
    B_\ell(R),
\]
where
\[
    B_\ell(R)
    :=
    \left(
        A_{\mathrm{NN}}
        +
        (B_{\mathrm{NN}}+2)R
    \right)^2.
\]
\end{lemma}

\begin{proof}
On $ \mathcal{E}_R,$ Lemma~\ref{lemma:linear_path_boundedness} gives
\[
    \|X_t\|_2
    \le
    R,
    \qquad
    \|X_1-X_0\|_2
    \le
    2R.
\]
By the smooth-network linear-growth bound in Lemma~\ref{lemma:nn_linear_growth},
\[
    \|u_\theta(X_t,t)\|_2
    \le
    A_{\mathrm{NN}}
    +
    B_{\mathrm{NN}}
    \|X_t\|_2
    \le
    A_{\mathrm{NN}}
    +
    B_{\mathrm{NN}}R.
\]
Therefore,
\[
\begin{aligned}
    \ell_\theta(Z)
    &=
    \|u_\theta(X_t,t)-(X_1-X_0)\|_2^2
    \\
    &\le
    \left(
        \|u_\theta(X_t,t)\|_2
        +
        \|X_1-X_0\|_2
    \right)^2
    \\
    &\le
    \left(
        A_{\mathrm{NN}}
        +
        (B_{\mathrm{NN}}+2)R
    \right)^2
    \\
    &=
    B_\ell(R).
\end{aligned}
\]
\end{proof}

\begin{lemma}[Metric-entropy uniform-convergence bound for the truncated loss class]
\label{lemma:pdim_uniform_convergence}

Let
\(
    \mathcal{G}
\)
be a class of real-valued functions satisfying
$
    0
    \le
    g(z)
    \le
    B
    \qquad
    \text{for every }
    g
    \in
    \mathcal{G},$
where
$B > 0.$
Let
$Z_1,\ldots,Z_N$
be independent samples, and suppose that, for some
$V \ge 1$
and some entropy radius
$ C_{\mathrm{ent}}
    \ge
    \max
    \left\{
        e,
        B
    \right\},$
the class satisfies
\[
    \log
    \mathcal{N}
    \left(
        \varepsilon,
        \mathcal{G},
        \|\cdot\|_\infty
    \right)
    \le
    V
    \log
    \left(
        \frac{
            C_{\mathrm{ent}}
        }{
            \varepsilon
        }
    \right),
    \qquad
    0
    <
    \varepsilon
    \le
    B.
\]
Define
\[
    \overline C_{\mathrm{ent}}(B,N)
    :=
    e
    \frac{
        N
        C_{\mathrm{ent}}
    }{
        B
    }.
\]
Then
\[
    \overline C_{\mathrm{ent}}(B,N)
    \ge
    e,
\]
and, with probability at least $1-\delta,$
\[
\begin{aligned}
    \sup_{g\in\mathcal{G}}
    \left|
        \mathbb{E}
        \left[
            g(Z)
        \right]
        -
        \frac{1}{N}
        \sum_{i=1}^N
        g(Z_i)
    \right|
    &\le
    C
    B
    \sqrt{
        \frac{
            V
            \log
            \left(
                \overline C_{\mathrm{ent}}(B,N)
            \right)
            +
            \log(1/\delta)
        }{
            N
        }
    },
\end{aligned}
\]
where \(
    C
    >
    0
\) is a universal constant.

In particular, consider the truncated flow-matching squared-loss class
\[
    \mathcal{G}_{W,D}^{(R)}
    :=
    \left\{
        Z
        \mapsto
        \ell_\theta(Z)
        \mathbf{1}_{\mathcal{E}_R}
        :
        \theta
        \in
        \Theta
    \right\},
\]
where
\[
    \ell_\theta(Z)
    =
    \left\|
        u_\theta
        \left(
            (1-t)X_0+tX_1,
            t
        \right)
        -
        (X_1-X_0)
    \right\|_2^2
\]
and
\[
    \mathcal{E}_R
    :=
    \left\{
        \|X_0\|_2
        \le
        R,
        \,
        \|X_1\|_2
        \le
        R
    \right\}.
\]
For the \(i\)-th sample, write
\[
    \mathcal{E}_{R,i}
    :=
    \left\{
        \|X_{0,i}\|_2
        \le
        R,
        \,
        \|X_{1,i}\|_2
        \le
        R
    \right\}.
\]

For the bounded-parameter smooth-activation neural velocity class in
Lemma~\ref{lemma:nn_linear_growth}, there exists a finite constant
\[
    L_{\mathrm{loss}}(R)
    >
    0
\]
such that
\[
    \sup_{Z:\,\mathcal{E}_R}
    \left|
        \ell_\theta(Z)-\ell_{\theta'}(Z)
    \right|
    \le
    L_{\mathrm{loss}}(R)
    \left\|
        \theta-\theta'
    \right\|_2
\]
for every
\(
    \theta,
    \theta'
    \in
    \Theta.
\)
Consequently, there exists a finite raw entropy radius
\[
    C_{\mathrm{sm}}(R)
    \ge
    \max
    \left\{
        e,
        B_\ell(R)
    \right\},
\]
depending only on the architecture, parameter constraint, activation derivative bounds, and truncation
radius, such that
\[
    \log
    \mathcal{N}
    \left(
        \varepsilon,
        \mathcal{G}_{W,D}^{(R)},
        \|\cdot\|_\infty
    \right)
    \le
    P
    \log
    \left(
        \frac{
            C_{\mathrm{sm}}(R)
        }{
            \varepsilon
        }
    \right),
    \qquad
    0
    <
    \varepsilon
    \le
    B_\ell(R).
\]

Define
\[
    \overline C_{\mathrm{sm}}(R,N)
    :=
    e
    \frac{
        N
        C_{\mathrm{sm}}(R)
    }{
        B_\ell(R)
    },
\]
and
\[
    \mathcal C_{\mathrm{sm}}(R,N)
    :=
    P
    \log
    \left(
        \overline C_{\mathrm{sm}}(R,N)
    \right).
\]
Then, with probability at least
\[
    1-\delta,
\]
\[
\begin{aligned}
    \sup_{\theta\in\Theta}
    \left|
        \mathbb{E}
        \left[
            \ell_\theta(Z)
            \mathbf{1}_{\mathcal{E}_R}
        \right]
        -
        \frac{1}{N}
        \sum_{i=1}^N
        \ell_\theta(Z_i)
        \mathbf{1}_{\mathcal{E}_{R,i}}
    \right|
    &\le
    C
    B_\ell(R)
    \sqrt{
        \frac{
            \mathcal C_{\mathrm{sm}}(R,N)
            +
            \log(1/\delta)
        }{
            N
        }
    }.
\end{aligned}
\]
\end{lemma}

\begin{proof}
We first prove the generic statement. Let
\[
    \varepsilon_N
    :=
    \frac{B}{N}.
\]
Since
\[
    0
    <
    \varepsilon_N
    \le
    B,
\]
the assumed covering-number bound implies that there exists an
\(
    \varepsilon_N
\) net
\[
    \mathcal{N}_{\varepsilon_N}
    \subseteq
    \mathcal{G}
\]
with cardinality satisfying
\[
\begin{aligned}
    \log
    \left|
        \mathcal{N}_{\varepsilon_N}
    \right|
    &\le
    V
    \log
    \left(
        \frac{
            C_{\mathrm{ent}}
        }{
            \varepsilon_N
        }
    \right)
    \\
    &=
    V
    \log
    \left(
        \frac{
            N
            C_{\mathrm{ent}}
        }{
            B
        }
    \right).
\end{aligned}
\]

For every
\(
    h
    \in
    \mathcal{N}_{\varepsilon_N},
\) Hoeffding's inequality gives
\[
    \mathbb{P}
    \left(
        \left|
            \mathbb{E}
            \left[
                h(Z)
            \right]
            -
            \frac{1}{N}
            \sum_{i=1}^N
            h(Z_i)
        \right|
        >
        B
        \sqrt{
            \frac{
                \log
                \left(
                    2
                    \left|
                        \mathcal{N}_{\varepsilon_N}
                    \right|
                    /
                    \delta
                \right)
            }{
                2N
            }
        }
    \right)
    \le
    \frac{
        \delta
    }{
        \left|
            \mathcal{N}_{\varepsilon_N}
        \right|
    }.
\]
Applying the union bound over
\(
    h
    \in
    \mathcal{N}_{\varepsilon_N}
\)
shows that, with probability at least
\(
    1-\delta,
\)
\[
\begin{aligned}
    \sup_{h\in\mathcal{N}_{\varepsilon_N}}
    \left|
        \mathbb{E}
        \left[
            h(Z)
        \right]
        -
        \frac{1}{N}
        \sum_{i=1}^N
        h(Z_i)
    \right|
    &\le
    B
    \sqrt{
        \frac{
            \log
            \left(
                2
                \left|
                    \mathcal{N}_{\varepsilon_N}
                \right|
                /
                \delta
            \right)
        }{
            2N
        }
    }.
\end{aligned}
\]

Because
\[
    V
    \ge
    1,
\]
and
\[
    \overline C_{\mathrm{ent}}(B,N)
    =
    e
    \frac{
        N
        C_{\mathrm{ent}}
    }{
        B
    },
\]
we have
\[
\begin{aligned}
    \log
    \left(
        2
        \left|
            \mathcal{N}_{\varepsilon_N}
        \right|
        /
        \delta
    \right)
    &\le
    V
    \log
    \left(
        \overline C_{\mathrm{ent}}(B,N)
    \right)
    +
    \log(1/\delta).
\end{aligned}
\]

Now fix an arbitrary
\(
    g
    \in
    \mathcal{G},
\)
and let
\(
    h_g
    \in
    \mathcal{N}_{\varepsilon_N}
\)
satisfy
\[
    \|g-h_g\|_\infty
    \le
    \varepsilon_N.
\]
Then
\[
\begin{aligned}
    \left|
        \mathbb{E}
        \left[
            g(Z)
        \right]
        -
        \frac{1}{N}
        \sum_{i=1}^N
        g(Z_i)
    \right|
    &\le
    \left|
        \mathbb{E}
        \left[
            h_g(Z)
        \right]
        -
        \frac{1}{N}
        \sum_{i=1}^N
        h_g(Z_i)
    \right|
    \\
    &\quad
    +
    2
    \varepsilon_N.
\end{aligned}
\]
Since
\[
    2
    \varepsilon_N
    =
    \frac{
        2B
    }{
        N
    },
\]
this approximation term is bounded by a universal constant multiple of
\[
    B
    \sqrt{
        \frac{
            V
            \log
            \left(
                \overline C_{\mathrm{ent}}(B,N)
            \right)
            +
            \log(1/\delta)
        }{
            N
        }
    }.
\]
Taking the supremum over
\(
    g
    \in
    \mathcal{G}
\)
proves the generic claim.

We next verify the entropy condition for the truncated smooth-network loss class. On
\(
    \mathcal{E}_R,
\)
Lemma~\ref{lemma:linear_path_boundedness} gives
\[
    \|X_t\|_2
    \le
    R,
    \qquad
    \|X_1-X_0\|_2
    \le
    2R.
\]
By Lemma~\ref{lemma:bounded_truncated_loss},
\[
    0
    \le
    \ell_\theta(Z)
    \mathbf{1}_{\mathcal{E}_R}
    \le
    B_\ell(R).
\]

Because the activation is smooth and the parameter region
$\Theta$
is compact, the map
$ \theta
    \mapsto
    \ell_\theta(Z)$
is continuously differentiable jointly over
$\Theta$
and the compact truncated input--target domain. Hence there exists a finite constant
$ L_{\mathrm{loss}}(R) > 0$
such that
\[
    \sup_{Z:\,\mathcal{E}_R}
    \left|
        \ell_\theta(Z)-\ell_{\theta'}(Z)
    \right|
    \le
    L_{\mathrm{loss}}(R)
    \left\|
        \theta-\theta'
    \right\|_2
\]
for every
$ \theta,
    \theta'
    \in
    \Theta.$
Since the indicator
$\mathbf{1}_{\mathcal{E}_R}$
does not depend on
$ \theta,$
the same inequality holds for the truncated losses.

Because
$\Theta$
is compact in
$\mathbb{R}^P,$
there exists a finite constant
$ C_\Theta > 0$
such that, for every
$ 0 < \eta \le C_\Theta,$
the parameter region has an
$ \eta$
net with cardinality at most
$ \left(
        \frac{
            C_\Theta
        }{
            \eta
        }
    \right)^P.$
Consequently, for every
$ 0 < \varepsilon \le C_\Theta L_{\mathrm{loss}}(R),$
\[
\begin{aligned}
    \log
    \mathcal{N}
    \left(
        \varepsilon,
        \mathcal{G}_{W,D}^{(R)},
        \|\cdot\|_\infty
    \right)
    &\le
    P
    \log
    \left(
        \frac{
            C_\Theta
            L_{\mathrm{loss}}(R)
        }{
            \varepsilon
        }
    \right).
\end{aligned}
\]
For larger
\(
    \varepsilon,
\)
a one-element cover suffices after increasing
\(
    C_\Theta
\)
if necessary. Therefore, defining
\[
    C_{\mathrm{sm}}(R)
    :=
    \max
    \left\{
        e,
        B_\ell(R),
        C_\Theta
        L_{\mathrm{loss}}(R)
    \right\},
\]
we obtain
\[
    \log
    \mathcal{N}
    \left(
        \varepsilon,
        \mathcal{G}_{W,D}^{(R)},
        \|\cdot\|_\infty
    \right)
    \le
    P
    \log
    \left(
        \frac{
            C_{\mathrm{sm}}(R)
        }{
            \varepsilon
        }
    \right),
    \qquad
    0
    <
    \varepsilon
    \le
    B_\ell(R).
\]

Finally, apply the generic result with
\[
    V
    =
    P,
    \qquad
    B
    =
    B_\ell(R),
    \qquad
    C_{\mathrm{ent}}
    =
    C_{\mathrm{sm}}(R).
\]
Then
\[
    \overline C_{\mathrm{ent}}(B,N)
    =
    e
    \frac{
        N
        C_{\mathrm{sm}}(R)
    }{
        B_\ell(R)
    }
    =
    \overline C_{\mathrm{sm}}(R,N),
\]
and the claimed uniform-convergence bound follows.
\end{proof}

\begin{lemma}[Tail control under sub-exponential endpoints]
\label{lemma:tail_control_subexp}

Suppose Assumption~\ref{assump:subexp} holds and the neural velocity class satisfies the uniform
linear-growth bound from Lemma~\ref{lemma:nn_linear_growth}. Define
\[
    \epsilon_{\mathrm{tail}}(R)
    :=
    \sup_{\theta\in\Theta}
    \mathbb{E}
    \left[
        \ell_\theta(Z)
        \mathbf{1}_{\mathcal{E}_R^c}
    \right],
\]
where
\[
    \mathcal{E}_R
    :=
    \left\{
        \|X_0\|_2
        \le
        R,
        \,
        \|X_1\|_2
        \le
        R
    \right\}.
\]
Then there exist constants
\[
    C_{\mathrm{tail}},
    c_{\mathrm{tail}}
    >
    0
\]
such that
\[
    \epsilon_{\mathrm{tail}}(R)
    \le
    C_{\mathrm{tail}}
    (1+R)^2
    \exp
    \left(
        -c_{\mathrm{tail}}R/K_X
    \right).
\]
\end{lemma}

\begin{proof}
By the uniform linear-growth property from Lemma~\ref{lemma:nn_linear_growth},
\[
    \|u_\theta(X_t,t)\|_2
    \le
    A_{\mathrm{NN}}
    +
    B_{\mathrm{NN}}
    \|X_t\|_2.
\]
For the linear path,
\[
    X_t
    =
    (1-t)X_0+tX_1,
\]
so
\[
    \|X_t\|_2
    \le
    \|X_0\|_2+\|X_1\|_2.
\]
Also,
\[
    \|X_1-X_0\|_2
    \le
    \|X_0\|_2+\|X_1\|_2.
\]
Therefore, after increasing constants if necessary, there exists
\(
    C_{\mathrm{env}}
    >
    0
\)
such that
\[
    \ell_\theta(Z)
    \le
    C_{\mathrm{env}}
    \left(
        1
        +
        \|X_0\|_2^2
        +
        \|X_1\|_2^2
    \right)
\]
uniformly over
\(
    \theta
    \in
    \Theta.
\)
Moreover,
\[
    \mathcal{E}_R^c
    \subseteq
    \left\{
        \|X_0\|_2
        >
        R
    \right\}
    \cup
    \left\{
        \|X_1\|_2
        >
        R
    \right\}.
\]
Thus,
\[
\begin{aligned}
    \epsilon_{\mathrm{tail}}(R)
    &\le
    C_{\mathrm{env}}
    \sum_{j\in\{0,1\}}
    \mathbb{E}
    \left[
        \left(
            1+\|X_j\|_2^2
        \right)
        \mathbf{1}_{\{\|X_j\|_2>R\}}
    \right].
\end{aligned}
\]

Let
\[
    Y_j
    :=
    \|X_j\|_2,
    \qquad
    j
    \in
    \{0,1\}.
\]
By Assumption~\ref{assump:subexp},
\[
    \mathbb{P}
    \left(
        Y_j>r
    \right)
    \le
    c_0
    \exp
    \left(
        -c_1r/K_X
    \right)
\]
for every
\[
    r
    \ge
    0.
\]
Moreover,
\[
    \mathbb{E}
    \left[
        \mathbf{1}_{\{Y_j>R\}}
    \right]
    =
    \mathbb{P}
    \left(
        Y_j>R
    \right)
    \le
    c_0
    \exp
    \left(
        -c_1R/K_X
    \right).
\]
For the quadratic contribution, the tail-integration identity gives
\[
\begin{aligned}
    \mathbb{E}
    \left[
        Y_j^2
        \mathbf{1}_{\{Y_j>R\}}
    \right]
    &=
    R^2
    \mathbb{P}
    \left(
        Y_j>R
    \right)
    \\
    &\quad
    +
    \int_R^\infty
    2r
    \mathbb{P}
    \left(
        Y_j>r
    \right)
    \,dr.
\end{aligned}
\]
Using the sub-exponential tail bound and setting
\[
    a
    :=
    c_1/K_X,
\]
we obtain
\[
\begin{aligned}
    \mathbb{E}
    \left[
        Y_j^2
        \mathbf{1}_{\{Y_j>R\}}
    \right]
    &\le
    c_0R^2
    \exp
    \left(
        -aR
    \right)
    \\
    &\quad
    +
    2c_0
    \int_R^\infty
    r
    \exp
    \left(
        -ar
    \right)
    \,dr
    \\
    &=
    c_0R^2
    \exp
    \left(
        -aR
    \right)
    \\
    &\quad
    +
    2c_0
    \exp
    \left(
        -aR
    \right)
    \left(
        \frac{R}{a}
        +
        \frac{1}{a^2}
    \right).
\end{aligned}
\]
After absorbing problem-dependent constants into
$C_{\mathrm{tail}}$ and, if necessary, decreasing the exponential rate to $c_{\mathrm{tail}},$
this gives
\[
    \mathbb{E}
    \left[
        \left(
            1+Y_j^2
        \right)
        \mathbf{1}_{\{Y_j>R\}}
    \right]
    \le
    C_{\mathrm{tail}}
    (1+R)^2
    \exp
    \left(
        -c_{\mathrm{tail}}R/K_X
    \right).
\]
Applying this bound for
\[
    j
    =
    0
    \qquad
    \text{and}
    \qquad
    j
    =
    1,
\]
and enlarging $C_{\mathrm{tail}}$
if necessary, proves the claim.
\end{proof}

\begin{lemma}[Uniform flow stability by Gronwall]
\label{lemma:flow_stability}

Let $u^\star$
be the target marginal velocity field, and let
$\{u_\theta:\theta\in\Theta\}$
be a class of learned velocity fields. Fix
$\theta
    \in
    \Theta.$
Let
$Z_t$
and
$Y_t^\theta$
solve
\[
    \frac{d}{dt}Z_t
    =
    u^\star(Z_t,t),
    \qquad
    \frac{d}{dt}Y_t^\theta
    =
    u_\theta(Y_t^\theta,t),
    \qquad
    Y_0^\theta
    =
    Z_0
    \sim
    \pi_0.
\]
Suppose Assumption~\ref{assump:flow_stability} holds. Then, by its path-consistency clause,
\[
    \operatorname{Law}(Z_t)
    =
    \rho_t
    \qquad
    \text{for every }t\in[0,1],
\]
and, in particular,
\[
    \operatorname{Law}(Z_1)
    =
    \pi_1.
\]
Moreover,
\[
    W_2
    \left(
        \widehat\pi^\theta_1,
        \pi_1
    \right)
    \le
    K_{\mathrm{flow}}
    \left(
        \int_0^1
        \mathbb{E}
        \left[
            \left\|
                u_\theta(Z_t,t)-u^\star(Z_t,t)
            \right\|_2^2
        \right]
        dt
    \right)^{1/2},
\]
where
\[
    K_{\mathrm{flow}}
    =
    \exp
    \left(
        \Lambda_{\mathrm{flow}}
    \right)
\]
is the deterministic uniform flow-stability constant from
Assumption~\ref{assump:flow_stability}. Equivalently,
\[
    W_2
    \left(
        \widehat\pi^\theta_1,
        \pi_1
    \right)
    \le
    K_{\mathrm{flow}}
    \left(
        \int_0^1
        \mathbb{E}_{X_t\sim\rho_t}
        \left[
            \left\|
                u_\theta(X_t,t)-u^\star(X_t,t)
            \right\|_2^2
        \right]
        dt
    \right)^{1/2}.
\]
\end{lemma}

\begin{proof}
Fix
\[
    \theta
    \in
    \Theta,
\]
and couple the learned and reference flows through the same initial condition
\[
    Y_0^\theta
    =
    Z_0
    \sim
    \pi_0.
\]
By the path-consistency clause in Assumption~\ref{assump:flow_stability},
\[
    \operatorname{Law}(Z_1)
    =
    \pi_1.
\]
Consequently, the joint construction of
\[
    (Y_1^\theta,Z_1)
\]
is a coupling of
\[
    \widehat\pi^\theta_1
    \qquad
    \text{and}
    \qquad
    \pi_1.
\]
Therefore,
\[
    W_2^2
    \left(
        \widehat\pi^\theta_1,
        \pi_1
    \right)
    \le
    \mathbb{E}
    \left[
        \left\|
            Y_1^\theta-Z_1
        \right\|_2^2
    \right].
\]

Define
\[
    \Delta_t
    :=
    Y_t^\theta-Z_t.
\]
Then
\[
\begin{aligned}
    \frac{d}{dt}\Delta_t
    &=
    u_\theta(Y_t^\theta,t)-u^\star(Z_t,t)
    \\
    &=
    \left(
        u_\theta(Y_t^\theta,t)-u_\theta(Z_t,t)
    \right)
    +
    \left(
        u_\theta(Z_t,t)-u^\star(Z_t,t)
    \right).
\end{aligned}
\]
For almost every $t \in [0,1],$
the spatial Lipschitz property of
$u_\theta$ gives
\[
    \frac{d}{dt}
    \|\Delta_t\|_2
    \le
    L_\theta(t)
    \|\Delta_t\|_2
    +
    \left\|
        u_\theta(Z_t,t)-u^\star(Z_t,t)
    \right\|_2.
\]
Since
\[
    \Delta_0
    =
    0,
\]
Gronwall's inequality yields
\[
\begin{aligned}
    \|\Delta_1\|_2
    &\le
    \int_0^1
    \exp
    \left(
        \int_s^1
        L_\theta(r)
        \,dr
    \right)
    \left\|
        u_\theta(Z_s,s)-u^\star(Z_s,s)
    \right\|_2
    \,ds
    \\
    &\le
    K_{\mathrm{flow}}
    \int_0^1
    \left\|
        u_\theta(Z_s,s)-u^\star(Z_s,s)
    \right\|_2
    \,ds,
\end{aligned}
\]
where the second inequality follows from
\[
    \int_0^1
    L_\theta(r)
    \,dr
    \le
    \Lambda_{\mathrm{flow}}.
\]
By the Cauchy--Schwarz inequality,
\[
    \|\Delta_1\|_2^2
    \le
    K_{\mathrm{flow}}^2
    \int_0^1
    \left\|
        u_\theta(Z_s,s)-u^\star(Z_s,s)
    \right\|_2^2
    \,ds.
\]
Taking expectation and combining the result with the coupling inequality for $W_2$
gives
\[
\begin{aligned}
    W_2^2
    \left(
        \widehat\pi^\theta_1,
        \pi_1
    \right)
    &\le
    K_{\mathrm{flow}}^2
    \int_0^1
    \mathbb{E}
    \left[
        \left\|
            u_\theta(Z_s,s)-u^\star(Z_s,s)
        \right\|_2^2
    \right]
    \,ds.
\end{aligned}
\]
Taking square roots proves the first stated inequality. The equivalent formulation follows from
\[
    \operatorname{Law}(Z_t)
    =
    \rho_t
    \qquad
    \text{for every }t\in[0,1].
\]
\end{proof}

\begin{lemma}[Numerical ODE discretization]
\label{lemma:ode_discretization_placeholder}

Suppose the learned velocity field
\[
    u_\theta
\]
satisfies the regularity assumptions required by a chosen ODE solver. Let
$\widehat\pi^\theta_1$
denote the terminal law of the exact learned ODE, and let
$\widehat\pi^\theta_{1,K}$
denote the terminal law obtained by a numerical solver with
$K$
steps. If the solver has distributional discretization error
\[
    W_2
    \left(
        \widehat\pi^\theta_{1,K},
        \widehat\pi^\theta_1
    \right)
    \le
    \epsilon_{\mathrm{disc}}(K),
\]
then
\[
    W_2
    \left(
        \widehat\pi^\theta_{1,K},
        \pi_1
    \right)
    \le
    W_2
    \left(
        \widehat\pi^\theta_1,
        \pi_1
    \right)
    +
    \epsilon_{\mathrm{disc}}(K).
\]
\end{lemma}

\begin{proof}
This follows immediately from the triangle inequality for
\[
    W_2:
\]
\[
\begin{aligned}
    W_2
    \left(
        \widehat\pi^\theta_{1,K},
        \pi_1
    \right)
    &\le
    W_2
    \left(
        \widehat\pi^\theta_{1,K},
        \widehat\pi^\theta_1
    \right)
    +
    W_2
    \left(
        \widehat\pi^\theta_1,
        \pi_1
    \right).
\end{aligned}
\]
Substituting the assumed discretization bound proves the result.
\end{proof}

%% file: simulation_results.tex
\section{Simulation Results}
\label{sec:simulation-results}

This section reports mechanism-oriented simulations for the finite-dataset
flow-matching analysis. The experiments are designed to illustrate the distinct roles of
the stored dataset size \(N\), which controls the statistical discrepancy between the
population objective \(L\) and the fixed empirical objective \(\widehat L_N\), and the
optimization budget \(T\), which controls the extent to which stochastic-gradient
training minimizes that empirical objective.

The simulations are not proofs of the theorem and should not be interpreted as empirical
verification of the formal sample-complexity exponent or the high-probability statement
over datasets. Rather, they provide controlled diagnostics of whether the mechanisms
separated in the theoretical decomposition are visible in a two-dimensional example. All
reported numerical values are means over five independent seeds, with standard errors
reported after the \(\pm\) symbol unless otherwise stated. The raw seed-level
measurements are included in the supplementary material.

All experiments use the standard Gaussian source distribution
\[
    \pi_0 = \mathcal{N}(0,I_2)
\]
and an eight-component Gaussian-mixture target distribution \(\pi_1\). For each
stored-dataset experiment and each seed, a finite dataset
\[
    \mathcal{D}_N
    =
    \left\{(X_{0,i},X_{1,i},t_i)\right\}_{i=1}^N
\]
is sampled once and reused during stochastic-gradient training. The interpolation path is
\[
    X_t = (1-t)X_0+tX_1,
    \qquad
    \dot X_t = X_1-X_0.
\]
The learned velocity field is a multilayer perceptron \(u_\theta(x,t)\). Training uses
projected stochastic gradient descent with projection onto a global Euclidean parameter
ball of radius \(25\). The observed parameter norms are approximately \(11\), well below
the projection radius, so the projection does not appear to be active in these runs.
Velocity loss is evaluated on independent validation samples unless otherwise stated.
Terminal quality is measured using sliced Wasserstein distance, empirical \(W_2\), MMD,
and scatter plots of samples generated by numerical integration of the learned ODE.

\subsection{Endpoint Sanity Check}

The endpoint sanity check verifies the probability-path convention used in the theory. If
the variable used as the terminal endpoint has law \(\pi_0\), then the interpolation does
not terminate at the target distribution. In contrast, when the terminal endpoint is
sampled as \(X_1 \sim \pi_1\), the path reaches the target endpoint by construction.

\begin{figure}[htbp]
    \centering
    \includegraphics[width=\linewidth]{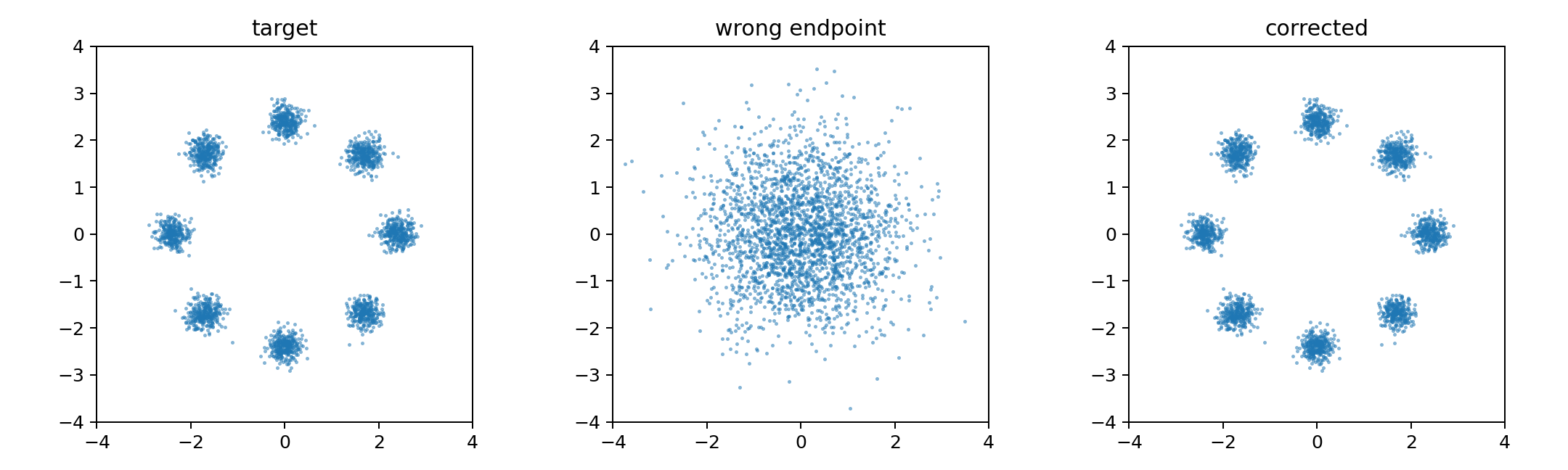}
    \caption{Endpoint sanity check. The incorrect endpoint construction remains close to
    the Gaussian source, whereas the corrected construction terminates at the target
    mixture by construction.}
    \textit{Three scatter plots compare target samples, an incorrect endpoint
    construction, and the corrected endpoint construction. The target and corrected
    endpoint panels show eight compact clusters arranged in a ring. The incorrect
    endpoint panel shows samples from a diffuse Gaussian source distribution instead of
    the target mixture.}
    \label{fig:offline-endpoint-sanity}
\end{figure}

The quantitative diagnostics agree with the visual comparison in
Figure~\ref{fig:offline-endpoint-sanity}. The incorrect endpoint construction has sliced
Wasserstein distance \(0.819 \pm 0.005\), MMD \(0.270 \pm 0.004\), and empirical \(W_2\)
distance \(1.317 \pm 0.013\) from the target samples. The corrected construction has
zero sample discrepancy in this diagnostic because it directly uses the target endpoint
samples themselves. Thus, this experiment is retained as an implementation and notation
sanity check, rather than as evidence for the learning or sample-complexity claims of
the theorem.

\subsection{Dataset-Size Scaling}

The dataset-size experiment varies \(N\) while training each fixed empirical objective
for \(T=4000\) projected-SGD updates. This diagnostic examines the finite-dataset
component of the analysis: as the stored dataset changes, the empirical objective may
provide a more or less accurate approximation to the population flow-matching objective.

\begin{figure}[htbp]
    \centering
    \includegraphics[width=\linewidth]{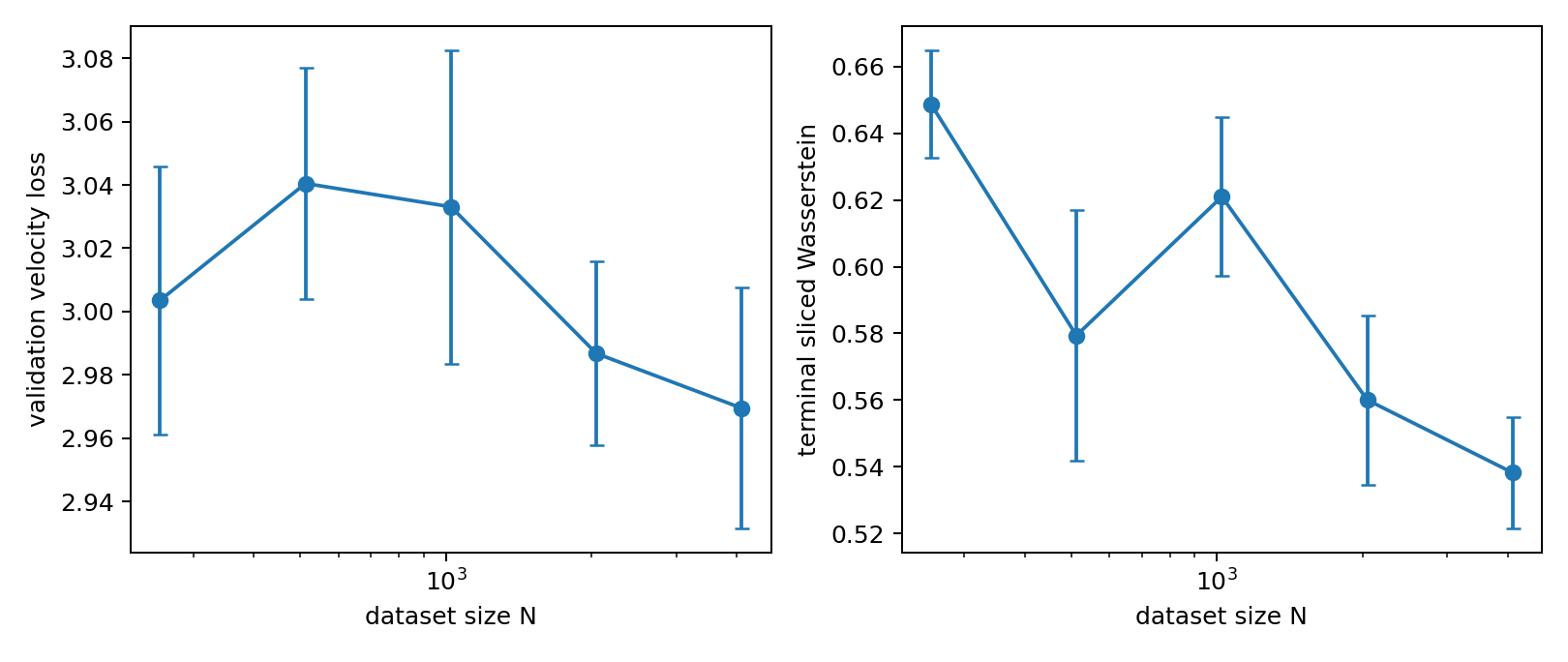}
    \caption{Dataset-size diagnostic. Points show seed means and error bars show standard
    errors over five seeds.}
    \textit{Two side-by-side line plots show validation velocity loss and terminal
    sliced Wasserstein distance as functions of stored dataset size. Both horizontal axes
    use a logarithmic scale. Validation loss remains close to 3 across all dataset sizes,
    with a small improvement at the largest sizes. Terminal sliced Wasserstein distance
    decreases overall from the smallest to the largest dataset size, but not
    monotonically. Error bars show standard errors across five seeds.}
    \label{fig:offline-dataset-size}
\end{figure}

The mean validation velocity losses are \(3.004 \pm 0.042\), \(3.040 \pm 0.037\),
\(3.033 \pm 0.050\), \(2.987 \pm 0.029\), and \(2.969 \pm 0.038\) for
\(N=256,512,1024,2048,\) and \(4096\), respectively. Terminal sliced Wasserstein
distance decreases from \(0.649 \pm 0.016\) at \(N=256\) to \(0.538 \pm 0.017\) at
\(N=4096\), although the trend is not monotone at intermediate values. Thus the
experiment shows a modest finite-dataset effect under this optimizer and training
budget. This is consistent with the theorem's decomposition, which separates the
finite-dataset contribution from optimization and approximation effects, but it does not
by itself identify a sample-complexity exponent.

\subsection{Optimization-Budget Scaling}

The optimization-budget experiment fixes \(N=2048\) and varies the number of empirical
projected-SGD updates \(T\). Since the exact empirical minimizer \(\widehat\theta_N\) is
unavailable, the smallest empirical loss observed along each seed's training trajectory
is used only as a visualization proxy for
\[
    \widehat L_N(\theta_T)-\widehat L_N(\widehat\theta_N).
\]

\begin{figure}[htbp]
    \centering
    \includegraphics[width=\linewidth]{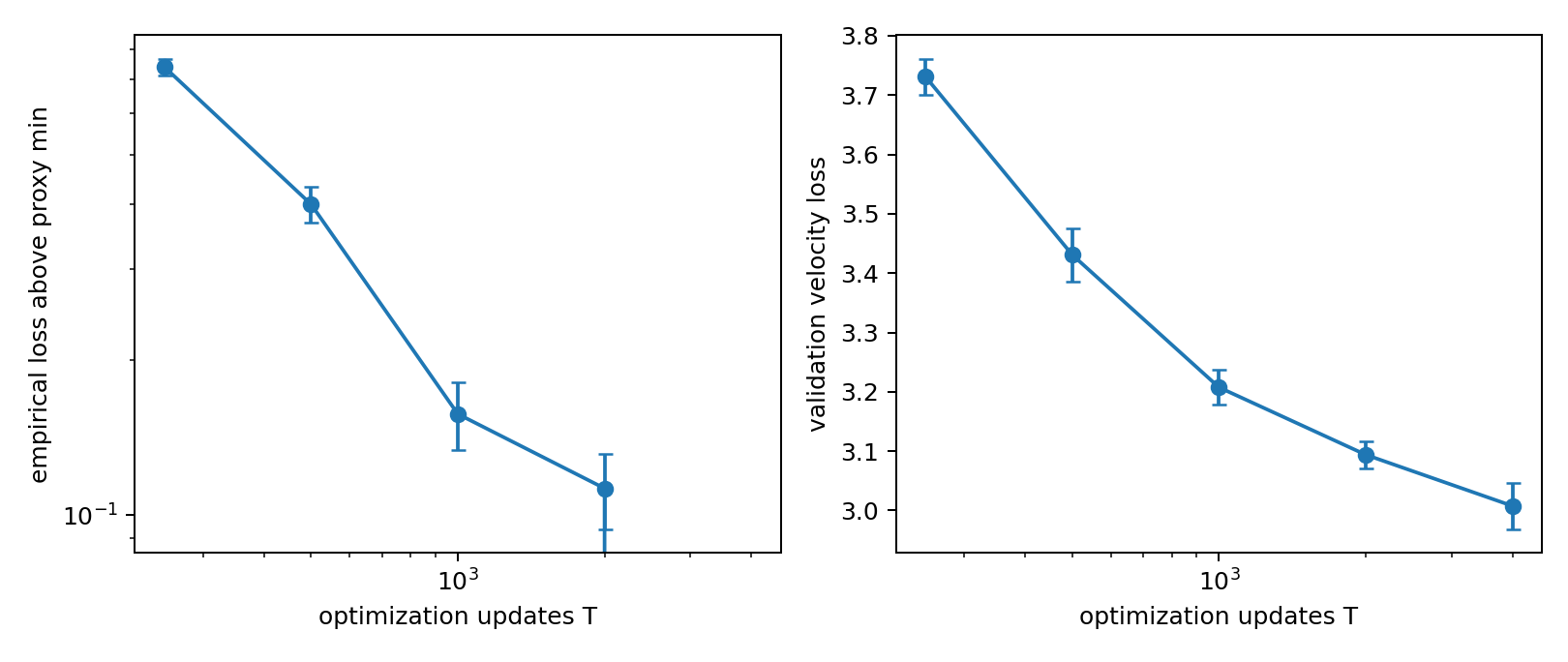}
    \caption{Optimization-budget diagnostic. Points show seed means and error bars show
    standard errors over five seeds.}
    \textit{Two side-by-side line plots show how performance changes as the number
    of projected-SGD updates increases. The left plot shows empirical loss above a proxy
    minimum on logarithmic horizontal and vertical axes; it decreases sharply toward zero
    as the update budget increases. The right plot shows validation velocity loss on a
    logarithmic horizontal axis; it decreases more gradually and begins to flatten at
    larger budgets.}
    \label{fig:offline-optimization-budget}
\end{figure}

The empirical optimization proxy decreases from \(0.740 \pm 0.028\) at \(T=250\) to
\(0\) at \(T=4000\), where the proxy minimum is attained by construction. Validation
velocity loss decreases from \(3.731 \pm 0.030\) at \(T=250\) to
\(3.008 \pm 0.039\) at \(T=4000\). Thus additional optimization visibly improves both
the empirical proxy and held-out loss, but held-out validation improves more slowly than
the empirical proxy. This illustrates that reducing empirical optimization error alone
need not eliminate finite-dataset statistical error, approximation error, or
endpoint-tail effects.

The plotted proxy should not be interpreted as the exact empirical suboptimality
appearing in Lemma~\ref{lemma:opt_error}. It is included only as a qualitative diagnostic
for finite-time optimization of the stored empirical objective.

\subsection{Joint Dataset-Size and Optimization-Budget Diagnostic}

The joint diagnostic varies \(N\) and \(T\) simultaneously. The theorem separates a
finite-dataset statistical contribution from a finite-time empirical optimization
contribution. Accordingly, this diagnostic examines whether the stored dataset size and
the training budget can have visibly different effects on held-out velocity loss.

\begin{figure}[htbp]
    \centering
    \includegraphics[width=0.72\linewidth]{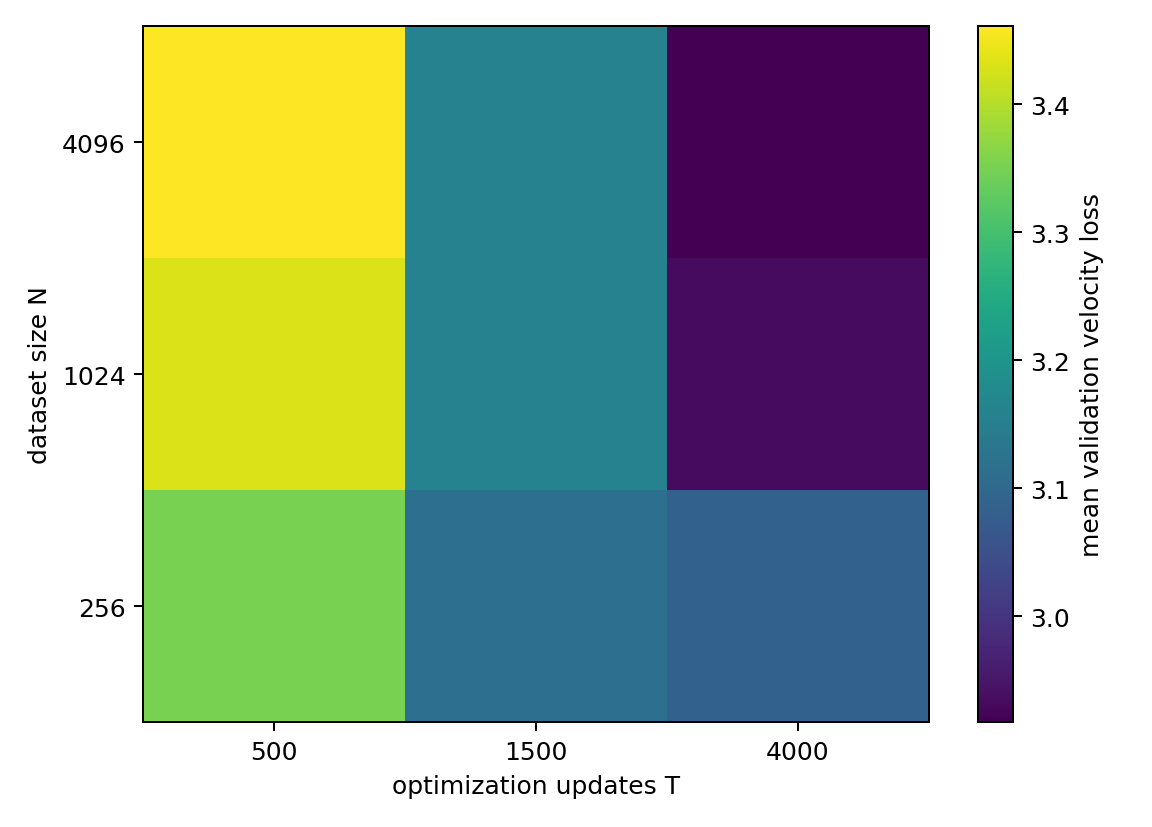}
    \caption{Joint \(N\)-versus-\(T\) diagnostic for mean held-out velocity loss over
    five seeds.}
    \textit{A heatmap shows validation velocity loss for three dataset sizes and
    three optimization budgets. Dataset size is on the vertical axis and optimization
    updates are on the horizontal axis. Brighter colors indicate larger validation loss
    and darker colors indicate smaller validation loss. For each dataset size, the cells
    become darker as the optimization budget increases from 500 to 4000 updates.}
    \label{fig:offline-joint-budget}
\end{figure}

The heatmap in Figure~\ref{fig:offline-joint-budget} shows a consistent improvement as
the projected-SGD budget increases. At \(N=256\), validation loss decreases from
\(3.352 \pm 0.083\) at \(T=500\) to \(3.086 \pm 0.019\) at \(T=4000\). At \(N=1024\),
it decreases from \(3.433 \pm 0.047\) to \(2.934 \pm 0.061\), and at \(N=4096\), it
decreases from \(3.461 \pm 0.037\) to \(2.917 \pm 0.042\). Across this limited grid,
increasing \(T\) has a clearer effect than increasing \(N\). The diagnostic therefore
supports the qualitative distinction between optimization budget and stored sample size,
but it does not imply that either effect is monotone in general or that the asymptotic
theorem's rates can be read from this small grid.

\subsection{Finite-Dataset Training Versus One-Pass Training}

The main theorem analyzes stochastic-gradient training on a fixed empirical objective
\(\widehat L_N\). As an ablation, we compare this finite-dataset protocol with a
one-pass population-style protocol in which each update uses freshly generated
flow-matching samples. These procedures optimize different objectives, so this
comparison is outside the formal scope of the finite-dataset theorem.

\begin{figure}[htbp]
    \centering
    \includegraphics[width=0.72\linewidth]{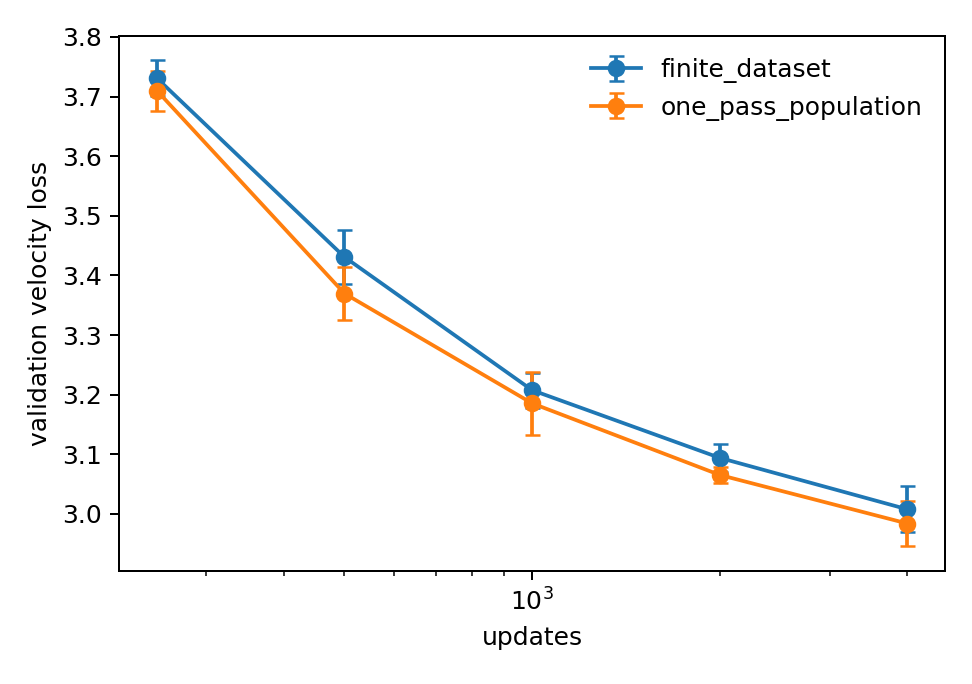}
    \caption{Finite-dataset training versus one-pass training. Points show seed means and
    error bars show standard errors over five seeds.}
    \textit{A line plot compares validation velocity loss for finite-dataset
    training and one-pass population training as the number of updates increases. Both
    curves decrease with more updates. The one-pass population curve is slightly below
    the finite-dataset curve at most budgets, and the two curves are close at the largest
    budget.}
    \label{fig:offline-vs-online}
\end{figure}

Both protocols improve with additional optimization. At \(T=4000\), the finite-dataset
validation loss is \(3.008 \pm 0.039\), while the one-pass validation loss is
\(2.984 \pm 0.038\). This does not contradict the theorem: the one-pass method receives
fresh samples at each update and therefore does not repeatedly optimize a single fixed
empirical objective. The comparison illustrates why the finite-dataset analysis
separately accounts for statistical error from the stored dataset and optimization error
from approximately minimizing its empirical objective.

\subsection{Coupling Choice in the Linear Path}

The coupling experiment compares independent coupling and minibatch optimal-transport
coupling for the same linear interpolation path. The relevant velocity-loss comparison
evaluates each trained model on held-out validation data drawn from the same coupling
used during training. For context, we also report independent-coupling validation loss
for the OT-trained model, but this cross-coupling quantity is not the objective optimized
by that model.

\begin{figure}[htbp]
    \centering
    \includegraphics[width=\linewidth]{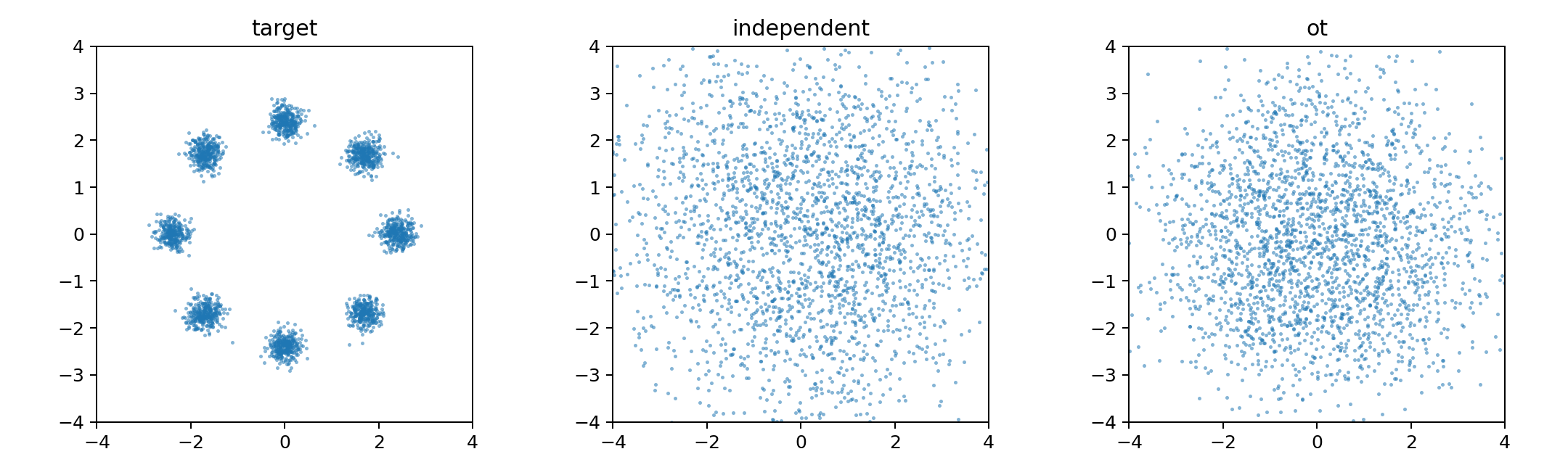}
    \caption{Coupling comparison for the linear path. The scatter panels show one
    representative seed; the table reports five-seed summaries.}
    \textit{Three side-by-side scatter plots compare target samples, generated
    samples from independent coupling, and generated samples from minibatch
    optimal-transport coupling. The target distribution has eight compact clusters
    arranged in a ring. The independent-coupling samples form a broad diffuse cloud. The
    optimal-transport-coupling samples are still diffuse, but are more concentrated near
    the target support.}
    \label{fig:offline-coupling}
\end{figure}

\begin{table}[t]
    \centering
    \begin{tabular}{lrrrr}
        \toprule
        Coupling & Same-coupling loss & Independent loss & Sliced \(W\) & Empirical \(W_2\) \\
        \midrule
        Independent & \(3.060 \pm 0.040\) & \(3.029 \pm 0.020\) & \(0.573 \pm 0.014\) & \(1.156 \pm 0.025\) \\
        Minibatch OT & \(0.360 \pm 0.006\) & \(3.248 \pm 0.015\) & \(0.449 \pm 0.007\) & \(0.978 \pm 0.011\) \\
        \bottomrule
    \end{tabular}
    \caption{Coupling comparison metrics. The same-coupling validation loss is the
    appropriate objective-specific velocity-regression quantity. The independent-coupling
    loss for the OT model is reported only as a cross-objective diagnostic.}
    \label{tab:offline-coupling}
\end{table}

The same-coupling validation loss is substantially smaller for minibatch OT coupling
than for independent coupling, and the terminal sliced Wasserstein and empirical \(W_2\)
metrics are also improved on average. However, the two couplings define different
regression problems, so their loss values should not be interpreted as a common-objective
comparison. The high independent-coupling loss of the OT-trained model confirms that
changing the coupling changes the learning problem itself. This supports treating
\(\Gamma\) as part of the probability-path specification rather than as a cosmetic
implementation detail.

\subsection{Assumption Diagnostics}

These plots are not proofs of the empirical proximal Polyak-\L{}ojasiewicz condition,
the interior operating regime required for inactive projection, the uniform-in-\(N\)
empirical-constant condition, or a conditional stochastic-gradient variance bound. They
are included only to assess whether the assumed optimization behavior is qualitatively
plausible for the finite empirical objective in this controlled setting.

\begin{figure}[htbp]
    \centering
    \includegraphics[width=\linewidth]{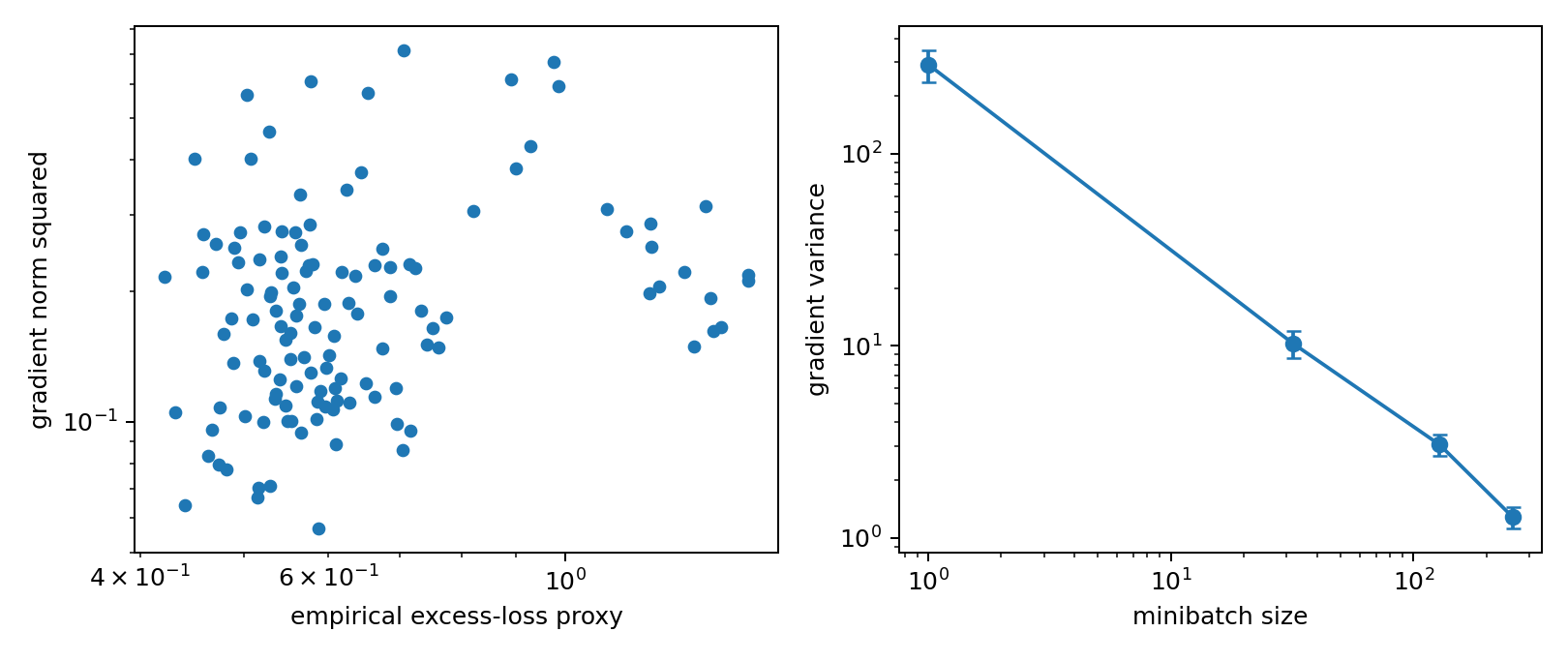}
    \caption{Empirical optimization diagnostics. The PL-style plot pools seed-level
    trajectory points, while the variance plot reports seed means and standard errors.}
    \textit{Two diagnostic plots are shown side by side. The left plot is a
    log-log scatter plot of empirical excess-loss proxy against squared gradient norm,
    pooling trajectory points from five seeds. The points form a noisy cloud rather than
    a clean line. The right plot is a log-log line plot of gradient variance versus
    minibatch size. Gradient variance decreases sharply as minibatch size increases, and
    error bars show standard errors across five seeds.}
    \label{fig:offline-assumption-diagnostics}
\end{figure}

The PL-style diagnostic is noisy and should not be interpreted as empirical verification
of gradient dominance. The longer-run empirical-loss baseline is not known to equal
\[
    \widehat L_N(\widehat\theta_N),
\]
which is why the empirical PL condition remains an explicit assumption in the theory.
The final parameter norms reported in the experiment logs are approximately \(11\), well
below the projection radius \(25\). This observation is consistent with, but does not by
itself verify, the interior operating regime assumed in the analysis.

The gradient-variance diagnostic is more direct. The estimated gradient variance
decreases from \(292.0 \pm 55.2\) for single-sample gradients to \(1.28 \pm 0.17\) for
minibatches of size \(256\). This observed reduction is consistent with the
finite-dataset stochastic-gradient oracle formulation and with the appearance of a
conditional variance quantity in the empirical optimization analysis. It does not, by
itself, establish the uniform conditional variance assumption required by
Lemma~\ref{lemma:opt_error}.